\documentclass[10pt,twocolumn,letterpaper]{article}

\usepackage[pagenumbers]{cvpr}              % To produce the CAMERA-READY version
\usepackage[dvipsnames, table]{xcolor}
\usepackage{graphicx}
\usepackage{amsmath}
\usepackage{amssymb}
\usepackage{booktabs}
\usepackage{subcaption}
\usepackage{pifont}
\usepackage{tocloft}
\usepackage{titletoc}
\usepackage{mathtools}
\usepackage[linesnumbered, commentsnumbered, ruled, vlined]{algorithm2e}
\usepackage{float}
\usepackage{setspace}
\usepackage{tcolorbox}

\AtEndDocument{
  \addtocontents{toc}{\protect\par\noindent\rule{\linewidth}{1pt}}%
}
\newcommand{\myparagraph}[1]{\vspace{0pt}\noindent{\bf #1}}
\newcommand{\vectorname}[1]{{\mathrm{\mathbf{#1}}}}

\newcommand{\highlight}[1]{\colorbox{Goldenrod}{#1}}
\newcommand{\hl}{\highlight}
\newcommand{\mainsec}[1]{\textcolor{magenta}{#1}}

\usepackage{listings}

\definecolor{codegreen}{rgb}{0,0.6,0}
\definecolor{codegray}{rgb}{0.5,0.5,0.5}
\definecolor{codepurple}{rgb}{0.58,0,0.82}
\definecolor{backcolour}{rgb}{0.95,0.95,0.92}

\lstdefinestyle{mystyle}{
    backgroundcolor=\color{backcolour},   
    commentstyle=\color{codegreen},
    keywordstyle=\color{magenta},
    numberstyle=\tiny\color{codegray},
    stringstyle=\color{codepurple},
    basicstyle=\ttfamily\footnotesize,
    breakatwhitespace=false,         
    breaklines=true,                 
    captionpos=b,                    
    keepspaces=true,                 
    numbers=left,                    
    numbersep=5pt,                  
    showspaces=false,                
    showstringspaces=false,
    showtabs=false,                  
    tabsize=2
}

\newcommand{\nocontentsline}[3]{}
\newcommand{\tocless}[2]{\bgroup\let\addcontentsline=\nocontentsline#1{#2}\egroup}
\usepackage[colorlinks,citecolor=green, pagebackref]{hyperref}

\graphicspath{{images/}}
\DeclareGraphicsExtensions{.pdf,.png,.jpeg}

\definecolor{cvprblue}{rgb}{0.21,0.49,0.74}
\newcommand{\supp}[1]{\textcolor{red}{Supp.}}

\def\paperID{7599} % *** Enter the Paper ID here
\def\confName{CVPR}
\def\confYear{2024}
\title{CLIPDraw++: Text-to-Sketch Synthesis with Simple Primitives}

\author{Nityanand Mathur$^{1\:*}$, Shyam Marjit$^{2\:*}$, Abhra Chaudhuri$^{3,4}$, Anjan Dutta$^{4}$\\
{$^{1}$ Smallest AI,  
$^{2}$ Indian Institute of Science,  
$^{3}$ University of Exeter,
$^{4}$ University of Surrey}\\
{\tt\footnotesize $^{1}$nityanand@smallest.ai, $^{2}$marjitshyam@gmail.com, $^{3}$ac1151@exeter.ac.uk, $^{4}$anjan.dutta@surrey.ac.uk}
}

\begin{document}
% \maketitle
\twocolumn[{%
\renewcommand\twocolumn[1][]{#1}%
\maketitle

% \twocolumn[{%
% \renewcommand\twocolumn[1][]{#1}%
% \def\thefootnote{*}\footnote{These authors contributed equally to this work}\def\thefootnote{\arabic{footnote}}
\begin{center}
\vspace{-4mm}
Project Page: \href{https://clipdrawx.github.io/}{https://clipdrawx.github.io/}
\end{center}

\begin{center}
    \centering
    \captionsetup{type=figure}
    \includegraphics[width=\linewidth]{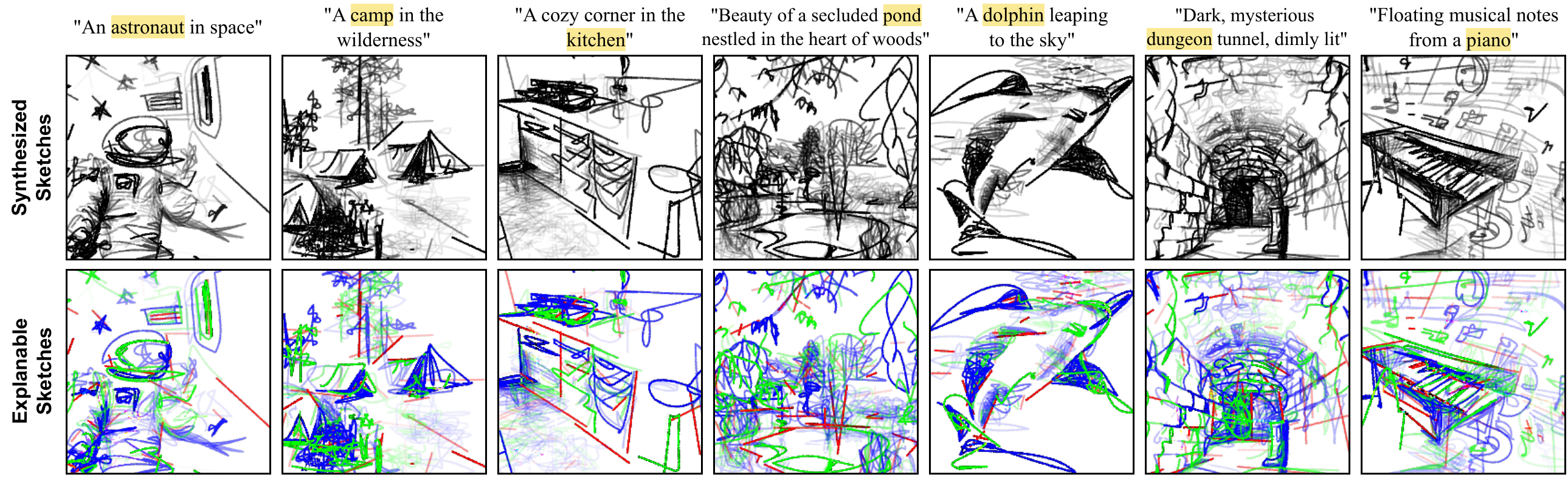}
    \captionof{figure}{
    Our method, CLIPDraw++ synthesizes vector sketches conditioned on an input text prompt using simple primitive shapes like \textcolor{blue}{circles}, \textcolor{red}{straight lines}, and \textcolor{green}{semi-circles}, with focus on the \colorbox{Goldenrod}{highlighted} words. More results are in the \supp.}
    \label{fig:teaser_diagram}
\end{center}
}]
\def\thefootnote{*}\footnotetext{Equal contribution.}

% \def\thefootnote{*}\footnote{Equal contribution.}\def\thefootnote{\arabic{footnote}}
% \def\thefootnote{}\footnote{Preprint. Under review.}\def\thefootnote{\arabic{footnote}}

% \footnote{}{}

% Abstract
\vspace{-25pt}
\begin{abstract}

With the goal of understanding the visual concepts that CLIP associates with text prompts, we show that the latent space of CLIP can be visualized solely in terms of linear transformations on simple geometric primitives like straight lines and circles.
Although existing approaches achieve this by sketch-synthesis-through-optimization, they do so on the space of higher order B\'ezier curves, which exhibit a wastefully large set of structures that they can evolve into, as most of them are non-essential for generating meaningful sketches. We present CLIPDraw++, an algorithm that provides significantly better visualizations for CLIP text embeddings, using only simple primitive shapes like straight lines and circles.
This constrains the set of possible outputs to linear transformations on these primitives, thereby exhibiting an inherently simpler mathematical form. The synthesis process of CLIPDraw++ can be tracked end-to-end, with each visual concept being expressed exclusively in terms of primitives. 
% Project Page: \href{https://clipdrawx.github.io/}{https://clipdrawx.github.io/}
%Implementation will be released upon acceptance. 

\end{abstract}

% Introduction
\vspace{-10pt}
\tocless{
\section{Introduction}
\label{sec:intro}
}
Simplified representations like sketches and verbal descriptions are potent mediums for communicating ideas, focusing on the core essence of the subject. While language conveys abstract meanings, sketches capture visual specifics. 
For instance, a designer might sketch a client's ideas for clarity during reviews of design plans, and automating this process could cut labour costs. Understanding the importance of this text-to-sketch generation task, several research initiatives have explored text-conditioned sketch generation, 
utilizing the CLIP model \cite{Radford2021CLIP} 
and the transformative diffusion models \cite{Rombach2022LatentDiffusion}. CLIPDraw \cite{Frans2022CLIPDraw} creates drawings from text using pretrained CLIP text-image encoders, while VectorFusion \cite{Jain2023VectorFusion} adapts text-guided models for vector graphics, without relying on extensive datasets.
Such models are thus finding a growing role in art and design \cite{Gao2022AIArt,Jain2023VectorFusion}, often surpassing human performance in many tasks \cite{Jumper2021Protein,Silver2016AlphaGo}. However, the complexity of the underlying algorithms tend to grow with with their performance, making them less transparent, and hence, less controllable.

To this end, we aim to cast the problem of sketch synthesis exclusively in terms of mathematically tractable primitives. The atoms of our algorithm consist of basic geometric shapes like straight lines, circles, and semi-circles, with a synthesis process that can be completely summarized as a linear transformation over these basic shapes.
The benefit of such a construction is two-fold -- (1) the number of parameters that need to be optimized is dramatically reduced, and (2) each step of the synthesis process can be clearly tracked and expressed via a closed form linear expression.
Our approach is based on synthesis-through-optimization, that visualizes using vector strokes, concepts encoded in the representation space of CLIP \cite{Radford2021CLIP} corresponding to natural language text prompts. Such vector strokes, as emphasized by CLIPDraw \cite{Frans2022CLIPDraw}, enable clearer breakdowns and simpler component attribution than pixel images. While advanced models like diffusion can produce high-quality pixel sketches \cite{Rombach2022LatentDiffusion}, their pixel complexity can obscure underlying logic. However, using vectorized strokes, especially when limited to primitive shapes, enhances clarity and a better understanding of the steps that a model takes to synthesize a sketch, while significantly lowering the number of parameters required to achieve state-of-the-art results.

In this paper, we present CLIPDraw++, an algorithm that can synthesize high-quality vector sketches based on text descriptions. It not only synthesizes sketches but also tracks the evolution of each vector stroke from the initial geometric primitives, like straight lines, circles, and half-circles, to the final output. Leveraging the cross-attention maps from a pretrained text-to-image diffusion model \cite{Rombach2022LatentDiffusion}, we define an initial sketch canvas as a set of primitive shapes, with their parameters tuned using a differentiable rasterizer \cite{Li2022Diffvg}.
In order to ensure minimality and a simple mathematical form, we have constrained the primitives to adhere to specific initial shapes with more rigid geometries than Be\'zier curves.
Our CLIPDraw++, built upon CLIPDraw \cite{Frans2022CLIPDraw}, differentiates itself by using predefined primitive shapes for strokes, emphasizing simplicity, while CLIPDraw initializes strokes with complex, arbitrarily-shaped Bézier curves, that exhibit a wastefully large set of shapes which they can transform into, but are not necessary for producing meaningful sketches. Our method linearly transforms these simple primitives into corresponding sketch strokes, unlike the more abstract and harder-to-understand sketches produced by CLIPDraw. Our CLIPDraw++ synthesizes superior sketches in a more efficient manner, both in terms of performance and memory usage, when compared to the existing methods. This improvement is attributed to the use of fewer primitives, each with fewer control points strategically distributed across the canvas. Unlike prevalent generative models \cite{Murdock2021BigGAN}, including diffusion models \cite{Rombach2022LatentDiffusion} that demand extensive parameter training, our CLIPDraw++ synthesizes sketches via optimization and operates without any specific training. Instead, a pretrained CLIP model is used as a metric for maximizing similarity between the
input text prompt and the synthesized sketch.

In summary, we make the following contributions -- \textbf{(1)} Pose the problem of sketch synthesis via optimization in terms of a well understood mathematical framework of learning linear transformations on simple geometric primitives; \textbf{(2)} Propose a sketch canvas initialization approach using primitive shapes. By leveraging the cross-attention map of a pre-trained diffusion model, we strategically distribute these primitives across the canvas based on necessity. Furthermore, by initializing these primitives with a low opacity, the model accentuates only those primitives pertinent to the text prompt; \textbf{(3)} Primitive-level dropout as an innovative technique to ``regularize'' our optimization. By doing so, we effectively diminish over-optimization, cut down on noisy strokes, and elevate the overall quality of the synthesized sketches; \textbf{(4)} Extensive qualitative and quantitative experiments demonstrate the usefulness of our novel components in delivering performance surpassing existing methods, while exhibiting a significantly simpler and parameter-efficient synthesis scheme.

\tocless{
\section{Related Works} 
\label{sec: related-works}
}

\myparagraph{Sketch Generation:} Free-hand sketches communicate abstract ideas leveraging the minimalism of human visual perception. They aim for abstract representations based on both structural \cite{Chan2022LineDrawings} and semantic \cite{Vinker2022CLIPasso} interpretations.
% This differs markedly from techniques that rely solely on extracting edge maps from images \cite{Canny1986Edge}, as free-hand sketching strives to showcase abstract depictions rooted in both structural \cite{Chan2022LineDrawings} and semantic interpretations \cite{Vinker2022CLIPasso}. 
% As such, digital sketching methods aspiring to replicate the human approach to drawing encompass a broad spectrum of sketch representations. These can range 
Digital sketching methods aiming to mimic human drawing span a wide range of representations, from those founded on the input image's edge map \cite{Li2017SketchSynth,Li2019PhotoSketching,Liu2020UnsupSketch,Chen2020DeepFaceDrawing,Tong2020SketchGenVectorFlow} to ones that venture into a higher degree of abstraction \cite{Ha2018NeuralSketch,Bhunia2022DoodleFormer,Frans2022CLIPDraw}, typically represented in vector format. 
In the realm of vector sketch creation, initiatives such as CLIPasso \cite{Vinker2022CLIPasso} and CLIPascene \cite{Vinker2022CLIPascene} are dependent on an image input; meanwhile, approaches like CLIPDraw \cite{Frans2022CLIPDraw}, VectorFusion \cite{Jain2023VectorFusion} and DiffSketcher \cite{xing2023diffsketcher} engage with text-based conditional inputs, standing distinct from other unconditional methodologies. 
Specifically, CLIPDraw \cite{Frans2022CLIPDraw} employs optimization-based sketch synthesis, whereas both VectorFusion \cite{Jain2023VectorFusion} and DiffSketcher \cite{xing2023diffsketcher} adopt diffusion-based approaches, with VectorFusion focusing on optimizing in the latent space, making it more light-weight and efficient compared to DiffSketcher that performs its optimization in the image space.
% In this paper, we explore text-conditioned sketch synthesis. Distinct from the existing approaches, we show that complex sketches can be synthesized as linear transformations on simple geometric primitives, proposing a model that achieves the same while tracking the evolution of the final sketch from the initial set of primitive shapes via an optimization step, inaugurating a fresh avenue of research, named \emph{explainable sketch synthesis}. In principle, ``tracking'' can be done for all the synthesis-through-optimization methods that use stroke-based initialization. Our novelty is in our minimalism -- complex sketches can be synthesized as linear transformations of primitives (which can also, of course, be tracked).
% }
% \anjan{have to differentiate between purely generative models and optimization based generative models.}
In this paper, we explore text-conditioned sketch synthesis. Distinct from the existing approaches, we show that complex sketches can be synthesized as simple linear transformations on simple geometric primitives, with our novelty lying in the minimalism of our formulation. 
% \anjan{have to add DiffSketcher and explain it is very different compared to our model.}

\myparagraph{Synthesis through Optimization:} Instead of directly training a network to generate images, a different strategy called \emph{activation maximization} optimizes a random image to match a target during evaluation \cite{Nguyen2016ImageSynthGAN}. CLIPDraw \cite{Frans2022CLIPDraw} advances this approach by using a CLIP language-image encoder \cite{Radford2021CLIP,Galatolo2021Caption2Image} to lessen the disparity between the synthesized image and a specified description, focusing on broad features rather than fine details. Although synthesis through optimization often results in unnatural or misleading images \cite{Nguyen2015DNNFool}, employing `natural image priors' can maintain authenticity \cite{Nguyen2016ImageSynthGAN,Nguyen2017PlugPlayGAN}, which often involve the restrictive and computationally intensive use of GANs. In this paper, we extend the capabilities of CLIPDraw to synthesize sketches of objects and scenes through geometric transformations of primitive shapes, contrasting from CLIPDraw's approach, which utilizes complex and abstract sketches created with hard-to-analyze, arbitrarily-shaped B\'{e}zier curves.
% that are more difficult to analyze.

\myparagraph{Vector Graphics:} We leverage the differentiable renderer for vector graphics pioneered by \cite{Li2022Diffvg}, a tool no longer confined to vector-specific datasets thanks to recent advancements. The advent of CLIP \cite{Radford2021CLIP}, which fosters improved visual text embedding, has spurred the development of robust sketch synthesis techniques including CLIPDraw \cite{Frans2022CLIPDraw}, CLIP-CLOP \cite{Mirowski2022Clipclop}, and CLIPascene \cite{Vinker2022CLIPascene}. The recently introduced VectorFusion \cite{Jain2023VectorFusion} also integrates a differentiable renderer with a diffusion model, aiding in the production of vector graphics creations like iconography and pixel art.

\myparagraph{Simplified representations for sketches:} Current sketch research on making neural sketch representations transparent and simplified is significantly limited, primarily concentrating on interpreting human sketches through stroke-level abstraction \cite{Alaniz2022PMN} and stroke location inversion \cite{Qu2023SketchXAI}.
However, as generative AI finds increasingly growing use in content creation, breaking down the abstraction inherent in such models for sketches \cite{Frans2022CLIPDraw, Vinker2022CLIPasso, Vinker2022CLIPascene} becomes ever more important. We present an approach that not only synthesizes sketches in terms of mathematically simple primitives, but also provides human-understandable insights into the representation space of foundation models like CLIP \cite{Radford2021CLIP}.

% Method
\vspace{6pt}
\tocless{
\section{CLIPDraw++}
\label{sec:method}
}
% problem statement in mathematical term
% what exactly we are trying to solve
% This work attempts to map the AI sketching to human sketching mimicking the human behavior of sketching. The methodology used in ClipDraw++ is described below.
    
\begin{figure*}[!t]
\includegraphics[width=\linewidth]{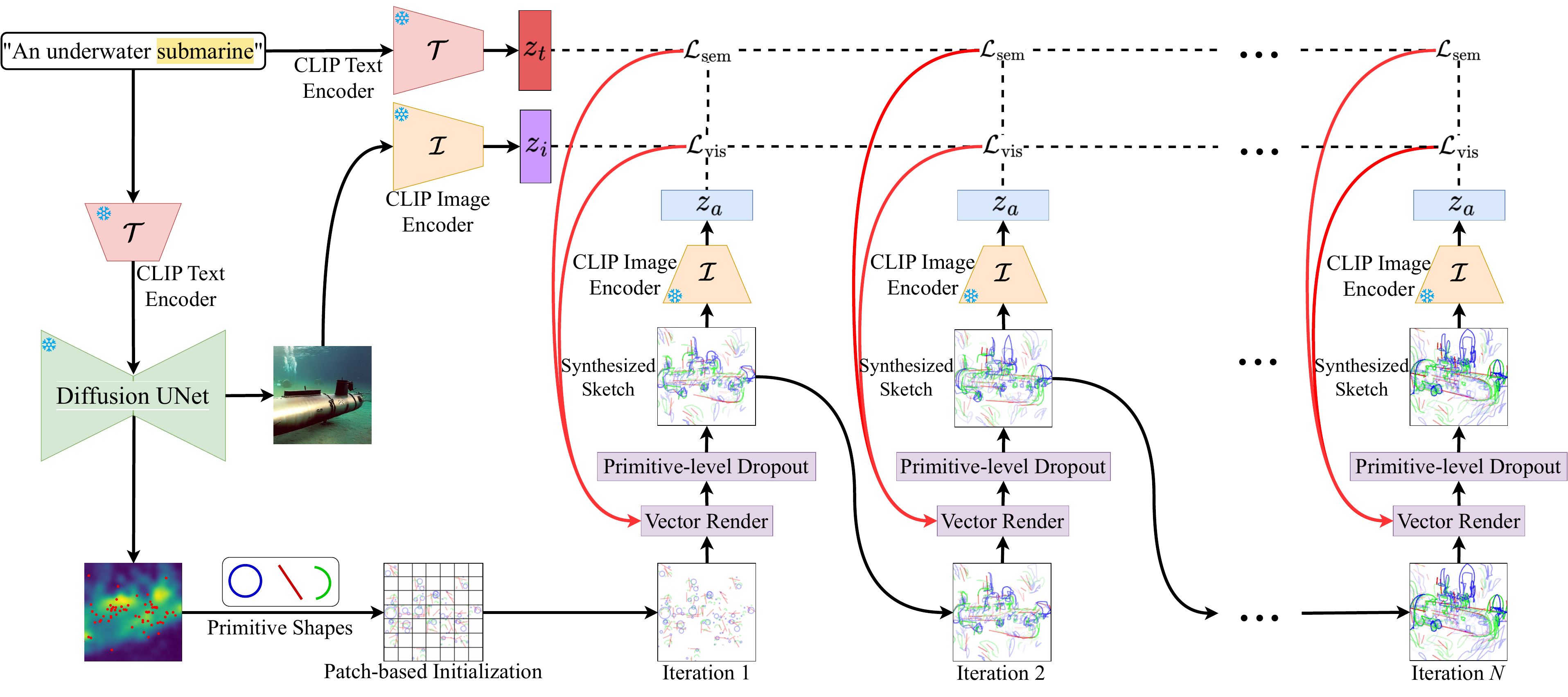}
\caption{CLIPDraw++ comprises strategic canvas initialization, which utilizes diffusion-based cross-attention maps and a patch-wise arrangement of primitives, along with a primitive-level dropout (PLD). The proposed model, coupled with the use of a pre-trained image ($\mathcal{I}$) and text ($\mathcal{T}$) encoders from the CLIP model for similarity maximization, positions itself as an efficient and user-friendly tool in the realm of AI-driven sketch synthesis. The \colorbox{Goldenrod}{highlighted} word is used to create the cross-attention maps.} 
% It efficiently synthesizes vector sketches from text descriptions using basic geometric primitives like straight lines, circles, and half circles.}
% by synthesizing vector sketches from textual descriptions using simple geometric primitives like straight lines, circles, and half circles.}
% CLIPDraw++ synthesizes vector sketches from textual descriptions using simple geometric primitives like straight lines, circles, and half circles. 
% CLIPDraw++ comprises strategic canvas initialization, which utilizes diffusion-based cross-attention maps and a patch-wise arrangement of primitives, along with a primitive-level dropout (PLD). The proposed model, coupled with the use of a pre-trained image ($\mathcal{I}$) and text ($\mathcal{T}$) encoders from the CLIP model for similarity maximization, positions itself as an efficient and user-friendly tool in the realm of AI-driven explainable sketch generation.}
% These components work together to produce sketches that are not only clean and comprised of essential strokes but also demonstrate improved efficiency and quality. Our approach, coupled with the use of a pre-trained image ($\mathcal{I}$) and text ($\mathcal{T}$) encoders from the CLIP model for similarity maximization, positions CLIPDraw++ as an efficient and user-friendly tool in the realm of AI-driven sketch generation.}
\vspace{-10pt}
\label{model_diagram}
\end{figure*}

In this work, we aim to synthesize sketches by expressing them as a collection of linearly transformed
% achieve explainability in machine synthesized sketches by expressing them as a collection of 
primitives, which could be evolved from an initial canvas into a set of strokes in a final sketch in a trackable manner based on an input text. We begin by populating a canvas with primitive shapes like straight lines, circles, and half-circles, detailed in \cref{ssec:prim-init}. We then track their progress during the sketch synthesis process outlined in \cref{ssec:optim-sketch-synth}, guided by the training criteria specified in \cref{ssec:train-crit}. An illustration of the proposed approach is provided in \cref{model_diagram}.

% \abhra{
\vspace{5pt}
\tocless{
\subsection{Sketch Synthesis from Primitives}
\label{ssec:explain_sketch_syn}}
Given an initial canvas $\mathbb{C}$ composed of a set of primitives $\left\{p_1, p_2,..., p_n\right\}$ as $\mathbb{C} = p_1 \oplus p_2 ... \oplus p_n$, 
% follows,
% \begin{equation*}
%     \mathbb{C} = p_1 \oplus p_2 ... \oplus p_n.
% \end{equation*}
we qualitatively show that any semantically meaningful sketch $\mathbb{Y}$ can be constructed as a transformation $f\left(\cdot\right)$ on $\mathbb{C}$ such that $\mathbb{S} = f(\mathbb{C}) = f_1(p_1) \oplus f_2(p_2) ... \oplus f_n(p_n)$,
% \begin{equation*}
%     \mathbb{S} = f(\mathbb{C}) = f_1(p_1) \oplus f_2(p_2) ... \oplus f_n(p_n),
% \end{equation*}
where $f_1, f_2,..., f_n$ are linear transformations on the primitives, and $\oplus$ denotes the composition of the primitives into a single canvas. In other words, any primitive $p \in \mathbb{C}$ maps to a concept $y \in \mathbb{Y}$, the target sketch as $y = F \cdot p$,
% follows:
% \begin{equation*}
%     y = F \cdot p,
% \end{equation*}
% % 
% \noindent 
where $F$ is a matrix encoding the linear transformation. The learning problem thus becomes finding the best approximation $\Tilde{F}$ of $F$ such that $p$ is appropriately transformed to depict some target concept in $\mathbb{Y}$. Thus, given an initial canvas and a target sketch, across all sub-concepts $y_i \in \mathbb{Y}$ in the target, the following objective needs to be optimized:
\vspace{-5pt}
\begin{equation}
    \min_{\mathcal{F}} l(\mathbb{C}_\mathcal{F}, \mathbb{Y}) = \arg \min_{i, \mathcal{F}} \sum_j || y_i - \Tilde{F} \cdot p_j ||
    \label{eqn:opt}
\vspace{-10pt}
\end{equation}
% 
% \anjan{Abhra, can you please check what $c_i$ is? Also, try to avoid the letter $c$ because we are using $c$ to annotate number of control points later} 
\noindent where $\mathcal{F} = \{\Tilde{F}_1, \Tilde{F}_2, ..., \Tilde{F}_n\}$ is a set of linear transformations applied on $\{p_1, p_2, ..., p_n\}$ from $\mathbb{C}$ respectively to obtain an approximation of the target sketch $\mathbb{Y}$, and $\mathbb{C}_\mathcal{F}$ denotes the sketch obtained under the set of transformations $\mathcal{F}$ applied on $\mathbb{C}$.
% }

% \abhra{
However, the target concept sketch $\mathbb{Y}$ is not available, since the main objective is to synthesize it from a textual description. We thus choose a vision-language model, specifically, CLIP \cite{Radford2021CLIP}, and use its latent representations as a proxy for the target sketch. The representation space of CLIP is unified, \ie, natural language sentences and their corresponding visual counterparts have the same embedding.
% \shyam{We denote the CLIP image encoder and text encoder as $\mathcal{I}\left(\cdot\right)$ and $\mathcal{T}\left(\cdot\right)$, respectively.} 
So, a sketch that captures the same semantics conveyed through a text prompt should have the same embedding in the representation space of CLIP. We thus make the CLIP text and sketch embeddings act as proxy representations for the ground-truth concept $\mathbb{Y}$ sketch and the synthesized sketch $\mathbb{C}_\mathcal{F}$ respectively and optimize the following objective which is equivalent to \cref{eqn:opt}:

\vspace{-10pt}
\begin{equation}
\small
    \min_{\mathcal{F}} l(\mathbb{C}_{\mathcal{F}}, x_t) = \arg \min_{\mathcal{F}} \left|\left| \mathcal{T}(x_t) - \mathcal{I} \left ( \bigcup_i \Tilde{F}_i \cdot p_i \right) \right |\right|
    \label{eqn:opt_clip}
    \vspace{-5pt}
\end{equation}
Thus in the formulation of \cref{eqn:opt_clip}, the text prompt $x_t$ serves as a proxy for the ground truth $\mathbb{Y}$. The functions $\mathcal{I}$ and $\mathcal{T}$ denote the image and text encoders of CLIP, respectively.

\tocless{
\subsection{Primitive-based Canvas Initialization}
\label{ssec:prim-init}}
We initialize the sketch canvas by first identifying the essential landmark points derived from the textual input. Subsequently, these landmarks are filled using primitive shapes like straight lines, circles, and half-circles.

% Why this is better 
% The CLIP loss function exhibits a pronounced non-convex characteristic, rendering the optimization process particularly sensitive to its initial conditions. This sensitivity is especially pronounced in scenarios involving multiple instances, where precise stroke placement becomes crucial for accentuating the overall semantic content conveyed through free-hand sketching.
% Due to the significant discrepancy in scale between optimization techniques like CLIP and the diffusion process, the former cannot generate sketches efficiently within a reduced number of iterative steps. To address this issue,
\myparagraph{Identifying Landmarks using Attention Maps:}
CLIPDraw++ initializes a canvas using primitives, which requires identifying important landmarks on that canvas based on the text input.
% \shyam{This context is used repeatedly} 
For that, we employ the DAAM~\cite{tang2022daam} generated token wise cross-attention %\anjan{is it only cross-attention and no self-attention? please confirm} \shyam{yes sir, it's cross attention maps only, should we also need to mention about the DAAM paper here} \abhra{yes, if you are using DAMM in your implementation}
mechanism as an integral component derived from the UNet architecture from latent diffusion model~\cite{Rombach2022LatentDiffusion}. Specifically, we capture attention features separately from both upper and lower sample blocks, merging them into a unified attention map. This composite attention map is then normalized using the softmax function to create a probabilistic distribution map.
% for further processing.
% the attention features separately from both the upper and lower sample blocks, subsequently merging into a unified attention feature set. To facilitate further processing, we employ the softmax function to normalize the composite attention map, thereby rendering it a probabilistic distribution map. 
%This distribution map serves as the basis for our subsequent task of selecting $n$ positions for each of the first \anjan{clarify this `first' with more words} control points, denoted as $p_i^j$, associated with Bézier curves.
This distribution map serves as the basis for our subsequent task of selecting $k$ positions.
The magnitude of each point within this attention-derived distribution map is leveraged as a weight parameter to guide the selection process. This
% sophisticated \abhra{I would not use ``sophisticated''} 
% approach
ensures that the selection of positions is influenced by the saliency and significance of the underlying attention-based features. To improve convergence towards semantic depictions, we place the primitive shapes or the initial strokes based on the salient regions (here patch) of the target image.

% CLIPDraw++ incorporates an initialization method based on a Latent-diffusion model~\cite{Rombach2022LatentDiffusion}, enhancing its initial setup. We employ the cross-attention mechanism as an integral component derived from the U-Net architecture. Specifically, we capture the attention features separately from both the upper and lower sample blocks, subsequently amalgamating these distinct attention representations into a unified attention feature set. To facilitate further processing, we employ the softmax function to normalize the composite attention map, thereby rendering it a probabilistic distribution map. This distribution map serves as the basis for our subsequent task of selecting $n$ positions for each of the first control points, denoted as $p_i^j$, associated with Bézier curves. Importantly, the magnitude of each respective point within this attention-derived distribution map is leveraged as a weight parameter to guide the selection process. This sophisticated approach ensures that the selection of control points is influenced by the saliency and significance of the underlying attention-based features. To improve convergence towards semantic depictions, we place the initial strokes based on the salient regions (here patch) of the target image.

\myparagraph{Patch-based Initialization:}
%\anjan{currently the purpose of this process is not clear, please clarify and try answering the following questions: (1) what is the motivation of patch-based initialization? (2) what problem or challenge are we solving with this? (3) how is this related with the attention map based initialization?}
Point-based initialization is effective for similar stroke types~\cite{Frans2022CLIPDraw, Vinker2022CLIPasso}, but determining the right primitive for a specific point on the attention map can be difficult.
% works for similar stroke type~\cite{Frans2022CLIPDraw, Vinker2022CLIPasso}. However, discerning the appropriate primitive for a given point on the attention map can be challenging. 
Additionally, placing different primitives at a single location is problematic as it may result in clutter due to high point density, leading to uneven primitive distribution and messy sketches. To address this, our CLIPDraw++ introduces primitives in fixed ranges or `patches', representing all points within, rather than at precise attention map locations. It divides a $224 \times 224$ canvas into patches of $32 \times 32$.
% or $56 \times 56$.
% , promoting uniform shape distribution. 
Each patch receives a mix of basic primitives: straight lines, circles, and semi-circles, promoting uniform shape distribution around the attention map local maxima.
% to populate the canvas systematically. 
The benefits of patch-based over point-based initialization are discussed in \cref{ssec:ablation} and illustrated in \cref{fig:patch-init}.

\myparagraph{Initializing Sketch Canvas with Primitives:} We define a sketch as a set of $n$ strokes $\left\{s_1, \ldots, s_n\right\}$ appearing in a canvas. In order to elucidate the origins and evolution of these strokes, we initialize our canvas with primitive shapes, such as straight lines, circles, and semi-circles. Each primitive shape is created using a two-dimensional
% Bézier curve
shape
that employs two to four control points, represented as $s_i=$ $\{p_i^j\}_{j=1}^c=\{(x_i, y_i)^j\}_{j=1}^c$ and an opacity attribute $\alpha_{i}$; where $c\in\left\{2,\:3,\:4\right\}$ denotes the number of control points. For example, straight line has 2 control points, while semi-circle and circle have 3 and 4. We incorporate the position of each
% Bézier
control point and opacity of the strokes into the optimization process and use the semantic knowledge in CLIP to guide the synthesis of a sketch
% CLIP semantics understanding to draw a sketch 
from a textual description. The parameters of the strokes are fed to a differentiable rasterizer $\mathcal{R}$, which forms the raster sketch $\mathcal{S}=\mathcal{R}((s_1, \alpha_1),\ldots,(s_n, \alpha_n))=\mathcal{R}((\{p_1^j\}_{j=1}^c, \alpha_1), \ldots,(\{p_1^j\}_{j=1}^c, \alpha_n))$. 
% where $c \in \{2,\:3,4\}$.

% \subsection{Primitive Initialization}
% Followed by the work of CLIPasso~\cite{Vinker2022CLIPasso} and VectorFusion~\cite{Jain2023VectorFusion} and it has been stated that primitive initialization plays a crucial role in text or image-to-sketch generation. Here we have two fundamental concerns choice/selection of strokes and their localization based on certain criteria.

\tocless{
\subsection{Optimizing Sketch Synthesis}
\label{ssec:optim-sketch-synth}}

Unwanted strokes can introduce noise into a sketch, making the removal of unnecessary strokes vital for automated sketch creation. This section outlines procedures for eliminating noisy strokes, drawing inspiration from existing machine learning techniques and human sketching practices.

\myparagraph{Primitive-level Dropout:} Dropout~\cite{Srivastava2014Dropout} is a regularization technique for neural networks where random subsets of neurons are temporarily deactivated during training. This procedure reduces overfitting by preventing co-adaptation of feature detectors and promoting a more robust network representation. 
Inspired by the success of dropout in learning robust representation, we propose \emph{primitive-level dropout} (PLD) in our CLIPDraw++ model. This technique focuses on optimizing the use of each sketch primitive and removing any that contribute unnecessary noise.
% At its core, primitive-level dropout revolves around maximizing the effective use of every primitive within a sketch, while simultaneously identifying and eliminating any superfluous primitives that may introduce unwanted noise in the canvas.
The intuition behind our approach is that limiting the number of available primitives compels the model to efficiently use each one to capture the semantics described in the texts, avoiding their wastage in noisy strokes. By selecting random smaller subsets of primitives for each iteration, the model is encouraged to utilize every primitive across all iterations for meaningful sketch representation, thereby minimizing their use in creating unnecessary noise.
% The intuition behind our approach is that, with fewer available primitives, the model would be forced to find the best possible use for each primitive to encode the semantics in the text, as it cannot afford to waste any of them in noisy strokes. If we pick random smaller subsets of the full primitive set in each iteration to synthesize the sketch, we would effectively be forcing the model to use each of the primitives in the full primitive set (across all iterations) to depict meaningful concepts in the sketch, thereby reducing primitives that are wasted in noisy strokes.

% During the optimization process, at every iterative step, 
In each step of the optimization process, a specific number of primitives, represented by $\mathcal{P}$ and determined through a probability distribution, are intentionally removed, after which a gradient step is undertaken. Formally, a random subset of the original primitives $\mathbb{C}$ is selected in each iteration to create a reduced canvas $\Tilde{\mathbb{C}}$ as follows:
% Formally, instead of the original set of primitives $\mathbb{C}$, in each iteration, we select a random subset of them to form a reduced canvas $\Tilde{\mathbb{C}}$ as follows:
%
\vspace{-5pt}
\begin{equation*}
    \vectorname{d} \sim \text{Bernoulli}(1 - \mathcal{P}); \;\; \Tilde{\mathbb{C}} = \mathbb{C} \cdot \vectorname{d}^T,
    \vspace{-5pt}
\end{equation*}
where $\vectorname{d}$ is an $n$-dimensional row vector of Bernoulli random variables, each of whose elements are 1 with probability $(1 - \mathcal{P})$, and 0 otherwise. Multiplying $\vectorname{d}$ with $\mathbb{C}$ masks out the primitives whose indices correspond to the elements of $\vectorname{d}$ that are 0, while the others are retained.
Subsequently, these $\mathcal{P}$ primitives are reintroduced into the optimization loop at the conclusion of the iteration, with another random subset of $\mathcal{P}$ primitives being removed in the next iteration. The cyclical approach of introducing, excluding, and then reintroducing primitives in CLIPDraw++ offers a balanced optimization, ensuring sketches are not overwhelmed with strokes but still retain vital elements.
% \abhra{
\def\thefootnote{1}
Without loss of generality\footnote{In general, each concept would be depicted by multiple primitives, but the central claim of this formalization would still hold.}, consider the simplified scenario where the number of primitives $n$ is equal to the number of target concepts $y \in \mathbb{Y}$, and the following holds:
\vspace{-5pt}
\begin{equation}
    \forall \text{ } y \in \mathbb{Y}, \exists \text{ } p \in \mathbb{C}, f \in \mathcal{F} \mid y = f(p)
    \label{eqn:sufficiency}
    \vspace{-5pt}
\end{equation}
In other words, each concept in the target text $\mathbb{Y}$ can be depicted uniquely as a transformation of a certain primitive, \ie, there exists a one-to-one mapping between $\mathbb{C}$ and $\mathbb{Y}$ under $\mathcal{F}$. Now, consider adding $\eta$ additional primitives to the primitive set such that $\mathbb{C}$ now becomes $\left\{p_1, p_2,..., p_n, p_{n+1}, ..., p_{n+\eta}\right\}$. However, the given premise states that generating $\mathbb{Y}$ was achievable using $\left\{ p_1, p_2,..., p_n \right\}$ only. Therefore the evolution of the additional primitives $\left\{p_{n+1}, ..., p_{n+\eta}\right\}$ is not constrained by the optimization objective, leaving open the possibility of them lying around the canvas as noisy strokes with no clear meaning. To guarantee sufficiency, \ie, the condition in \cref{eqn:sufficiency}, we always overestimate the number of primitives required to visualize a certain text prompt, to ensure the synthesized sketch is complete with the required details. However, as argued formally, this overestimation could leave room for some noisy strokes crowding up the canvas. Primitive-level dropout ensures that this does not happen by randomly removing $\eta$ (proportional to the estimated noise rate) primitives in each iteration, thereby forcing all of the remaining primitives to contribute towards representing something meaningful. The empirical benefits of this PLD approach are discussed in \cref{ssec:ablation} and can be seen in \cref{fig:pld-eiffel}.
% \anjan{have to mention a portion of text, which I will when it is ready} 
% \anjan{have to fix \cref{fig:Patch} is mentioned before \cref{fig:PLD}}
% }

% \abhra{I think that the following paragraph can be removed.}
% Primitive level dropout refines the sketches, reducing noise, and leading to superior outcomes,
% \abhra{guaranteeing that each primitive encodes something meaningful in the final sketch, rather than being wasted as noisy strokes}.
% much like an artist's iterative process in perfecting their work \abhra{the intuition behind this is not clear, so can be removed in my opinion}.
% \anjan{we require ablation on this, applying dropout should generate clearer sketches.} \nits{Reference Fig. ~\ref{fig:PLD}}

% The concept of primitive-level dropout in CLIPDraw++ entails a deliberate emphasis on the comprehensive utilization of all primitives contained within the sketch, concomitant with the strategic removal of extraneous primitives that function as deleterious noise within the canvas. In each iteration,  $\mathcal{P}$ primitives, based on a probability distribution are deliberately excluded from active participation in the optimization process, after which a gradient step is undertaken. Subsequently, these $\mathcal{P}$ primitives are reintroduced into the optimization loop at the conclusion of the iteration. This judicious integration of primitive-level dropout serves to promote a smoother and less noisy evolution of the sketch, thereby facilitating enhanced synthesis outcomes.

\myparagraph{Initializing Primitives with Diminished Opacity:} In traditional sketching, artists often begin with a light outline or faint layout, serving as a foundation for the artwork. This initial phase sets the broader structure and composition. As the artwork advances, artists intensify strokes, especially focusing on crucial elements to make them prominent, ensuring each stroke adds value to the overall piece. Drawing a parallel to the digital realm, in CLIPDraw++, we have curated a similar methodology. Here, the initialization of primitive shapes starts with a low opacity value, denoted as $\alpha$. This can be likened to the faint layout artists create. As the system begins its optimization process, based on relevance and significance, the opacity of certain primitives is incrementally increased. This mirrors the artist's method of iteratively intensifying strokes that are deemed crucial to the sketch's integrity.
% \abhra{
Formally, in each backward pass, we update the opacity value of a primitive $p$ as $\alpha_p \leftarrow \nabla_\mathcal{I}(\mathcal{L}_\text{total}); \;\; \alpha_p > K$,
%
% \begin{equation*}
%     \alpha_p \leftarrow \nabla_\mathcal{I}(\mathcal{L}_\text{total}); \;\; \alpha_p > K
% \end{equation*}
%
where $\mathcal{L}_\text{total}$ is the optimization criterion of CLIPDraw++ from \cref{eqn:total_loss} and $K$ is an empirical constant. We retain $p$ in the canvas if the inequality is met and drop it otherwise.
% }
In essence, CLIPDraw++ attempts to replicate the thoughtful and incremental approach artists employ, blending the nuances of human artistry with the precision of machine optimization. The advantage of initializing strokes with diminished opacity is discussed in Sec. C.2 and demonstrated in Fig. 7 of the \supp.
\vspace{5pt}
\tocless{
\subsection{Training Criteria}
\label{ssec:train-crit}}

Our training criteria evaluate the alignment between input text prompts and the synthesized sketches, including their augmented versions. In order to measure the similarity, we employ the pre-trained text and image encoders from the CLIP model~\cite{Radford2021CLIP}.
% , which has undergone extensive training on diverse image-text pairs. 
The ability of CLIP to encode information from both natural images (here sketches) and texts, eliminates the need for additional training. To measure the alignment of a given text and the synthesized sketch we use cosine similarity as $\text{sim}(x, y) = \frac{x \cdot y}{||x||\cdot||y||}$.
% \myparagraph{Augmentations:} The primary aim of image augmentation lies in the preservation of recognizability in graphical representations when subjected to diverse distortions. Absent the employment of image augmentation strategies, the optimization-driven synthesis methodologies frequently yield outcomes that, although numerically aligned with predetermined objectives, manifest as adversarial images perceptually obfuscated to human observers. In the scope of our present study, we employ a carefully orchestrated sequence of transformation functions, namely \texttt{torch.transforms.RandomPerspective} and \texttt{torch.transforms.RandomResizedCrop}, to manifest this imperative augmentation procedure. This serves a dual purpose: it incentivizes plausible images to exhibit diminished loss, while concurrently imparting a heightened loss to implausible renditions.

\myparagraph{Semantic Loss:} Our primary objective is to amplify the semantic similarity between the text prompts and the synthesized sketches. 
% Given the notable proficiency of the text and image encoders from the CLIP model in comprehending and representing high-level semantic nuances, it becomes an instrumental tool for this task. 
To measure the semantic congruence between a text prompt $P$ and rasterized version of the synthesized sketch $S$, we introduce the semantic loss function, $\mathcal{L}_\text{sem}$, designed to ensure semantic coherence and capture discrepancies between the two modalities.
\vspace{-8pt}
\begin{equation*}
\mathcal{L}_\text{sem} = - \sum\limits_{i = 0}^{M} \text{sim}\left(\mathcal{T}(P),\:\mathcal{I}\left(T_{f}\left(\mathcal{S}\right)\right)\right)
\vspace{-8pt}
\end{equation*}
where $\mathcal{T}_{f}$ indicates randomized affine transformations, and $M$ denotes the number of transformed variations produced through augmentations. 
% For our purposes, we specifically use \texttt{RandomPerspective} and \texttt{RandomResizedCrop} as our selected image transformations.
% sequence of image transformations and $K$ is the number of transformed variations under consideration. 
% These image transformations enhance model generalization and robustness by artificially expanding and diversifying the training dataset through various transformations of the original images. Specifically, in this work, we employ \texttt{RandomPerspective} and \texttt{RandomResizedCrop} as our chosen image transformations.
% Here the image transformation function aims to maintain image recognizability despite various distortions. Without it, optimization techniques often produce images that, while numerically accurate, appear adversarially distorted to humans. In our study, we use transformation functions like \texttt{torch.transforms.RandomPerspective} and \texttt{torch.transforms.RandomResizedCrop} for augmentation, which both reduces loss for plausible images and increases loss for implausible ones.
% Notably, CLIP excels at capturing high-level semantic attributes - building upon this foundation, to generate vector sketches that align with specific textual prompt $P$, we define $\mathcal{L}_{semantic}$ loss as follows:

\begin{figure*}[t!]
\centering
\includegraphics[width=1.0\textwidth]{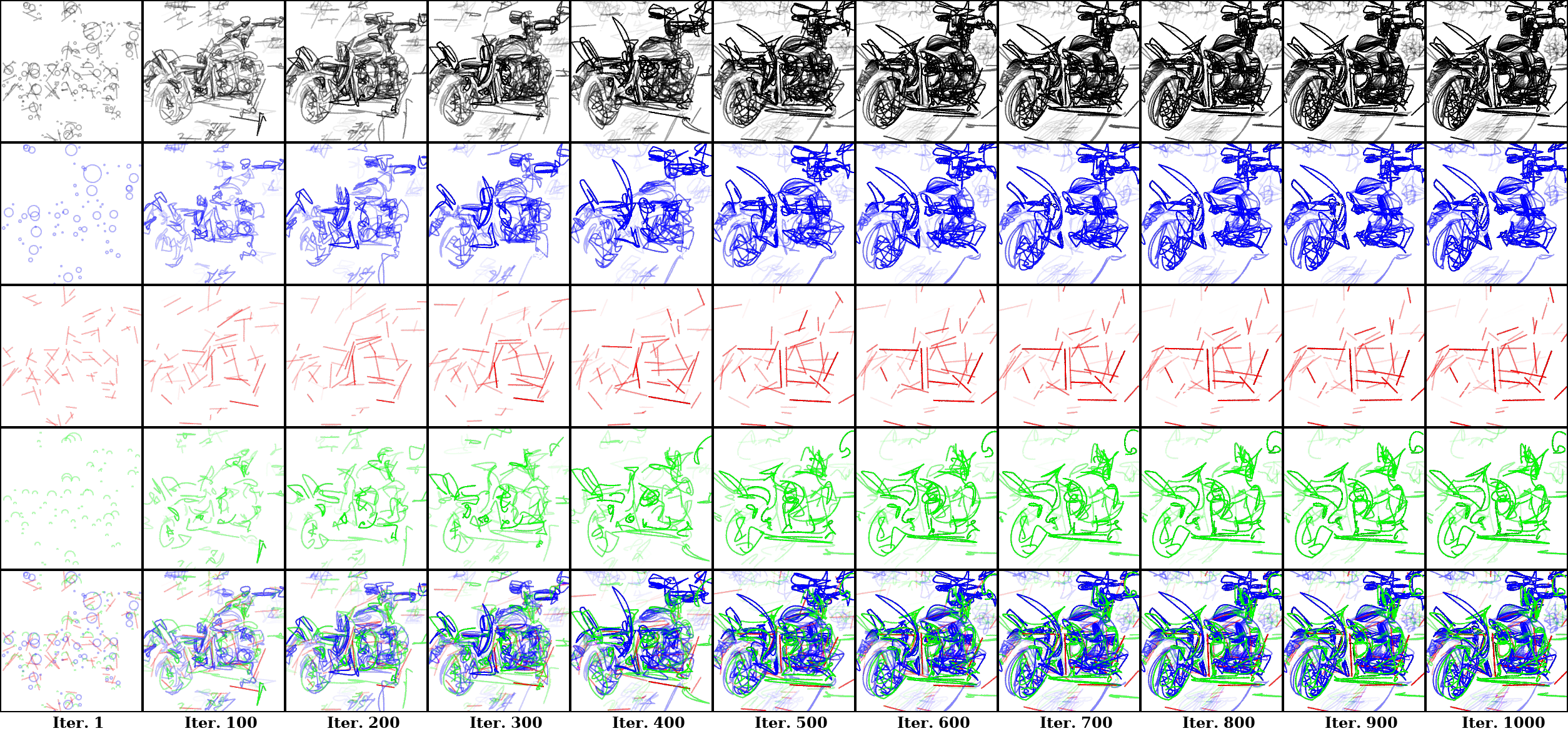}
\caption{For the text prompt ``A standing \colorbox{Goldenrod}{motorcycle}'', our CLIPDraw++ tracks the evolution of each primitive shape during optimization: the first row shows a black-and-white synthesized sketch, the next three rows display the development of \textcolor{blue}{circles}, \textcolor{red}{straight lines}, and \textcolor{green}{semi-circles}, and the final row combines these three rows' compositions. Here ``motorcycle'' is used to create the cross-attention maps.}
\vspace{-5pt}
\label{fig:fulltrack-motorcycle}
\end{figure*}

\myparagraph{Visual Loss:} While semantic loss focuses on high-level semantic cues, it may overlook important low-level spatial attributes such as pose and structure. Therefore, to complement the semantic loss, we introduce a criteria $\mathcal{L}_\text{vis}$ measuring the geometric congruence between the image generated by the diffusion UNet and the sketch synthesized by our model for the same input text, which is computed as:
\vspace{-5pt}
\begin{equation*}
\mathcal{L}_\text{vis} = - \sum\limits_{i=0}^{M}  
 \text{sim}\left(\mathcal{I}\left(I_{0}\right), \mathcal{I}(T_{f}\left(\mathcal{S}\right))\right)
 \vspace{-5pt}
\end{equation*}
where $I_{0}$ denotes the final image generated by frozen UNet. 
% \vspace{1pt}
% To bridge this gap and enable finer control over the output appearance, we introduce a method that measures the geometric similarity between the LDM generated image and the sketch. This added dimension allows us to better preserve essential structural and spatial characteristics. To further enhance the synthesis quality and maintain fidelity to the output image from the frozen LDM, we incorporate a visual loss. This unique combination ensures that the generated sketches not only align semantically with the text prompt but also exhibit perceptual and geometric coherence. In doing so, we successfully address the limitations associated with relying solely on semantic loss, resulting in a more comprehensive and nuanced approach to image synthesis.
% \par
% where $I_{0}$ denotes the final image generated by frozen LDM.

\myparagraph{Total Loss:}
The total loss, $\mathcal{L}_\text{total}$ is the summation of two loss functions (semantic loss $\mathcal{L}_\text{sem}$ and visual loss $\mathcal{L}_\text{vis}$) explained above, each weighted by their respective coefficients, $\lambda_\text{sem}$ and $\lambda_\text{vis}$. These two loss functions balance our sketch synthesis process: semantic loss aligns vector sketches with textual prompts, while visual loss maintains low-level spatial features and perceptual coherence. This combination effectively captures the intricate relationship between semantic fidelity and geometric accuracy.
% The combination of these two loss functions balances the imperative alignment of generated vector sketches with textual prompts, as achieved through semantic loss, with the preservation of essential low-level spatial features and perceptual coherence, as ensured by the visual loss. It effectively encapsulates the nuanced interplay between semantic fidelity and geometric accuracy in our sketch synthesis process.
%
\vspace{-3pt}
\begin{equation}
    \mathcal{L}_\text{total} = \lambda_\text{sem} \mathcal{L}_\text{sem} + \lambda_\text{vis} \mathcal{L}_\text{vis}
    \label{eqn:total_loss}
    % \vspace{-3pt}
\end{equation}

% \begin{algorithm}[!ht]
% \SetKwInOut{Input}{Input}
% \SetKwInOut{Output}{Output}

% \Input{A text prompt $P$; Loss weights $\lambda_s$ \& $\lambda_v$; Initial Opacity $\alpha_{init}$; Number of iteration $N$, Pre-trained \texttt{CLIP(ViT-B/32)} and \texttt{Latent-diffusion} model.}
% \Output{A rendered sketch $\mathcal{S}$}

% \tcp{Cross-attention embeddings for text prompt from LDM}
% $CrossAttn \longleftarrow$ \texttt{Latent\_diffusion\_model\:}($P$) \\

% $\mathcal{A} \longleftarrow$ Set of primitive types - \emph{straight line, circle, semi-circle}\\
% $\mathcal{D} \longleftarrow$ Differentiable vector rasterizer\\
% $\mathcal{CLIP} \longleftarrow$ \\

% $\mathcal{S} \longleftarrow$ Black canvas of size $\left(224px \times 224px\right)$\\

% \tcp{}
% \tcp{}
% \tcp{}

% \For{epoch $\leftarrow 1$ \KwTo $N$}{
%     $p \sim \mathcal{P}, \; (s_1, s_2) \sim \mathcal{S} \; \mid \; p \xleftrightarrow{} (s_1, s_2)$\\
    
%     $(\vectorname{x}^1_p, \vectorname{x}^1_s) \longleftarrow \Gamma(p, s_1)$ \\
    
%     \tcp{Outputs are treated as constants}
%     with no gradient: \\
%      \Indp $(\vectorname{x}^2_p, \vectorname{x}^2_s) \longleftarrow \Gamma(p, s_2)$ \\
%      \Indm
    
%     $\mathcal{L}_{\text{teacher}} \longleftarrow \textbf{\texttt{XAQC}}(\vectorname{x}^1_p, \vectorname{x}^2_s, \mathcal{Q})$ \\
    
%     $\mathcal{Q} \longleftarrow \mathcal{Q}$.enqueue$(\vectorname{x}^2_s)$\\
%     $\Gamma \longleftarrow \Gamma - \eta \nabla_\Gamma \mathcal{L}_{\text{teacher}}$
% }
% \caption{\textsc{CLIPDraw++}: Optimization algorithm}

% \label{alg:teacher}
% \end{algorithm}

% Experiment
\tocless{
\section{Experiments}
\label{sec:expts}
}

In this section, we use CLIPDraw++ to synthesize sketches from linearly transformed primitives like circles, straight lines, and semi-circles. We also compare CLIPDraw++ with related methods and conduct ablations to evaluate its components. More results are in the \supp.
% assess various components of CLIPDraw++ through a comprehensive ablation study to confirm their effectiveness.
% various interesting behaviors of CLIPDraw++ are highlighted through a variety of examples. With the exception of Section 4.1, example images are picked to best convey the behavior in consideration. The focus is on qualitative observations, unusual behavior, or recurring trends in CLIPDraw++ image synthesis.

\vspace{4pt}
\tocless{
\subsection{Sketch Generation from Primitives} 
\label{ssec:primit-explan}}

% \vspace{-5pt}
As shown in \cref{fig:fulltrack-motorcycle}, our CLIPDraw++ model offers the ability to synthesize sketches whose strokes are linearly transformed primitive shapes like circles (second row from the top), straight lines (third row), and half circles (fourth row). These individual strokes can be tracked through their evolution in successive iterations of the optimization process. Notably, the model intuitively represents different parts of the synthesized sketch with appropriate primitive shapes. For example, the chassis and handlebars of the synthesized \emph{motorcycle} in \cref{fig:fulltrack-motorcycle} are rendered with straight lines, while the wheels are depicted using circles and semicircles. 
Our model skillfully captures the dynamics of shape or scene evolution, displaying varying levels of flexibility based on the degree of freedom, which is linked to the number of control points in a shape.
In the \emph{farm} example (refer to \supp{} Fig. 1 (b)), it assigns straight lines to simpler structures like a house's roof and walls, while more complex elements like grass and crops are made with semi-circles, providing more flexibility. Even more intricate structures, like trees, are rendered using circles, the most flexible shape, illustrating the model's skill in using various primitives for different levels of complexity. This strategic use of shapes enhances the model's ability to create detailed, nuanced sketches. Additional examples of sketch generation with primitive level tracking and overall sketch level tracking are respectively shown in Fig. 1 and Figs. 2-5 of the \supp.
\definecolor{customgreen}{RGB}{213, 232, 212}
\definecolor{customblue}{RGB}{204, 255, 255}

\begin{figure*}[t!]
\centering
\includegraphics[width=1.0\textwidth]{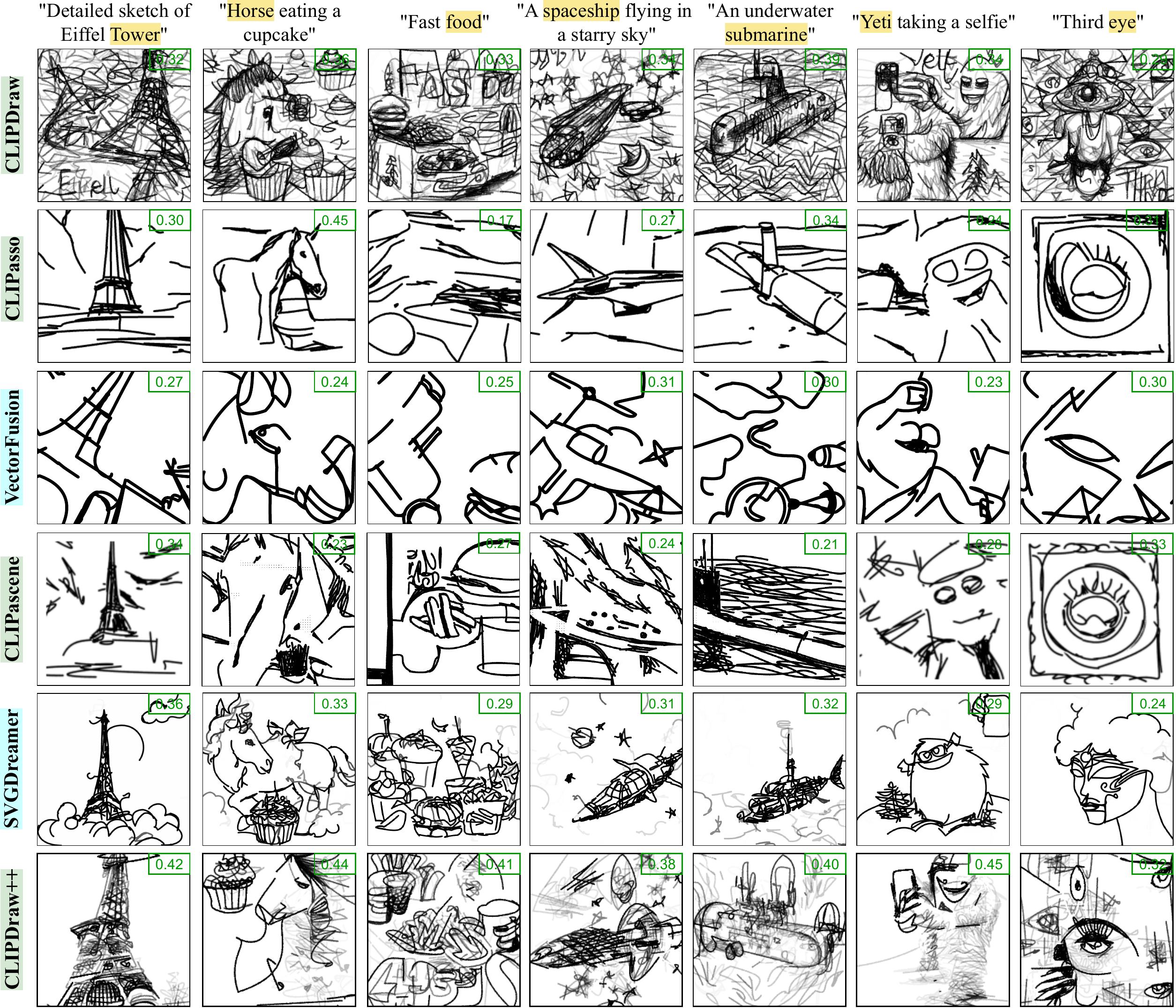}
% \vspace{-15pt}
\caption{Qualitative comparison with quantitative measure between CLIP based approaches \colorbox{customgreen}{CLIPDraw}, \colorbox{customgreen}{CLIPasso}, \colorbox{customgreen}{CLIPascene}, and diffusion based approaches \colorbox{customblue}{VectorFusion}, \colorbox{customblue}{SVGDreamer}, with our CLIPDraw++. The numbers in green at the top-right corner of each image indicate the CLIP-T score wrt the text prompt. The \colorbox{Goldenrod}{highlighted} words are used to create cross-attention maps.}
\label{fig:comparison}
\vspace{-3pt}
\end{figure*}

\vspace{5pt}
\tocless{
\subsection{Comparison}
\label{ssec:compar}}

Our CLIPDraw++ model is compared with five related methods: \textbf{CLIPDraw} \cite{Frans2022CLIPDraw}, which optimizes Bézier curves for CLIP-guided sketches; \textbf{CLIPasso} \cite{Vinker2022CLIPasso}, which simplifies sketches with Bézier curves; \textbf{CLIPascene} \cite{Vinker2022CLIPascene}, which generates scene sketches from CLIP embeddings; \textbf{VectorFusion} \cite{Jain2023VectorFusion}, which uses a diffusion model for vector sketches; and \textbf{SVGDreamer} \cite{xing2024svgdreamer}, which creates clearer sketches via a stroke-based diffusion approach. All models are tested with their original settings for fair comparison.

\begin{table}[!h]
% \vspace{-5pt}
\centering
\resizebox{0.47\textwidth}{!}{
\begin{tabular}{l|c|c|c|c|c}
\toprule
\textbf{Method / Metric} & \textbf{CS} $\uparrow$ & \textbf{PSNR} $\uparrow$ & \textbf{CLIP-T} $\uparrow$ & \textbf{BLIP} $\uparrow$ & \textbf{Conf.} $\uparrow$\\ \midrule
CLIPDraw \cite{Frans2022CLIPDraw} & 0.2578 & 28.1740 & 0.3114 & 0.2611 & 0.49 \\
CLIPasso \cite{Vinker2022CLIPasso} & 0.2250 & 27.5000 & 0.2850 & 0.2783 & 0.45 \\
VectorFusion \cite{Jain2023VectorFusion} & 0.2283 & 28.3277 & 0.2949 & 0.3894 & 0.44 \\
CLIPascene \cite{Vinker2022CLIPascene} & 0.2000 & 27.0231 & 0.2746 & 0.2551 & 0.42\\
% BigGAN & 0.2076 & 28.0923 & 0.2965 & 0.4017 & 0.53 \\  
SVGDreamer \cite{xing2024svgdreamer} & 0.2688 & \textbf{28.7384} & 0.3132 & 0.4102 & \textbf{0.61} \\ \midrule
\rowcolor[gray]{0.9} \textbf{CLIPDraw++} (Ours) & \textbf{0.2763} & 28.6417 & \textbf{0.3365} & \textbf{0.4222} & 0.58 \\ \bottomrule
\end{tabular}
}
\caption{Quantitative comparison with existing methods.}
% \vspace{-12pt}
\label{tab:clip_t}
\end{table}

In \cref{tab:clip_t}, we present quantitative experiments to validate our approach using different evaluation metrics following existing literature \cite{Frans2022CLIPDraw,xing2024svgdreamer}, including Cosine Similarity (CS) \cite{huang2024t2i},
Peak Signal-to-Noise Ratio (PSNR) \cite{hore2010image}, CLIP-T Score \cite{Radford2021CLIP}, BLIP score \cite{li2022blip}, 
and confusion score (Conf.) \cite{xing2023diffsketcher}. Our CLIPDraw++ has significantly outperformed all other methods in CS, CLIP-T, and BLIP metrics, and achieves the second best in PSNR and Conf, demonstrating its effectiveness. The higher scores in CS, CLIP-T, and BLIP indicate that CLIPDraw++ generates sketches closely aligned with the text prompts. Although SVGDreamer achieves slightly higher PSNR and confusion scores due to its stroke-based diffusion approach optimized directly in image space, our method is more efficient and achieves realistic sketches without extensive optimization. Furthermore, our high PSNR indicates less supersaturation, and the confusion score underscores the realism of our generated sketches.

% \begin{table*}[!htbp]
% \vspace{-20pt}
% \centering
% \resizebox{\textwidth}{!}{
% \begin{tabular}{l|c|c|c|c|c|c|c|c}
% \hline
% Method / Metric       & CS $\uparrow$ & FID $\downarrow$ & PSNR $\uparrow$ & CLIP-T $\uparrow$ & BLIP $\uparrow$ & Aes. $\uparrow$ & HPS $\uparrow$ & Conf. $\uparrow$\\ \hline
% CLIPDraw    &     0.2578     &   79.6512  &   28.1740   &   0.3114     &    0.2611       &  3.8097        &  0.2928  & 0.49 \\
% VectorFusion         &   0.2283       &   111.3729  &  28.3277    &    0.2949    &      0.3894     &     \textbf{4.6043}      & 0.2627  & 0.44  \\
% BigGAN &    0.2076      & \textbf{32.9452}  &   28.0923    &    0.2965    &     0.4017      &     3.9544      &   \textbf{0.3627} & 0.53 \\ \hline
% CLIPDraw++ (our)            &    \textbf{0.2763}      &   73.3281  &   \textbf{28.6417}  &    \textbf{0.3365}    &     \textbf{0.4222}      &    4.4926       &   0.3298 & \textbf{0.58} \\ \hline
% \end{tabular}
% }
% \caption{Quantitative evaluation metrics}
% \vspace{-30pt}
% \label{tab:clip_t}
% \end{table*}

As demonstrated in \cref{fig:comparison}, our CLIPDraw++ produces sketches that are noticeably cleaner and semantically closer (as indicated by CLIP-T score) than those from CLIPDraw, likely due to our method's use of primitive-level dropout and the initialization of primitives at reduced opacity. Compared to CLIPasso, which simplifies sketches using B\'{e}zier curves but often loses fine details, CLIPDraw++ retains both clarity and detail without excessive smoothing. In contrast to VectorFusion, which employs a diffusion model for vector sketches but struggles with abstract and less coherent representations, our method ensures more precise and semantically rich outputs. CLIPascene, while capable of generating scene-level sketches, often introduces clutter due to less effective control of primitive placement, whereas CLIPDraw++ maintains clarity through structured primitive initialization. Although SVGDreamer produces visually clearer sketches via a stroke-based diffusion approach, its optimization time is significantly longer, and it occasionally lacks semantic accuracy. In contrast, CLIPDraw++ consistently delivers clean sketches with accurate details and semantics. The reduced noise in our sketches is attributed to the minimized control points, learnable opacity for primitives, and primitive-level dropout, thus reducing the need for manual intervention and parameter adjustments. 
\tocless{
\subsection{Model Ablations}
\label{ssec:ablation}}
In this section, we present selected ablation studies of our model. Additional ablation results are in Sec. C of the \supp.

\begin{figure}[h]
% \vspace{-5pt}
    \centering
    \includegraphics[width=0.45\textwidth]{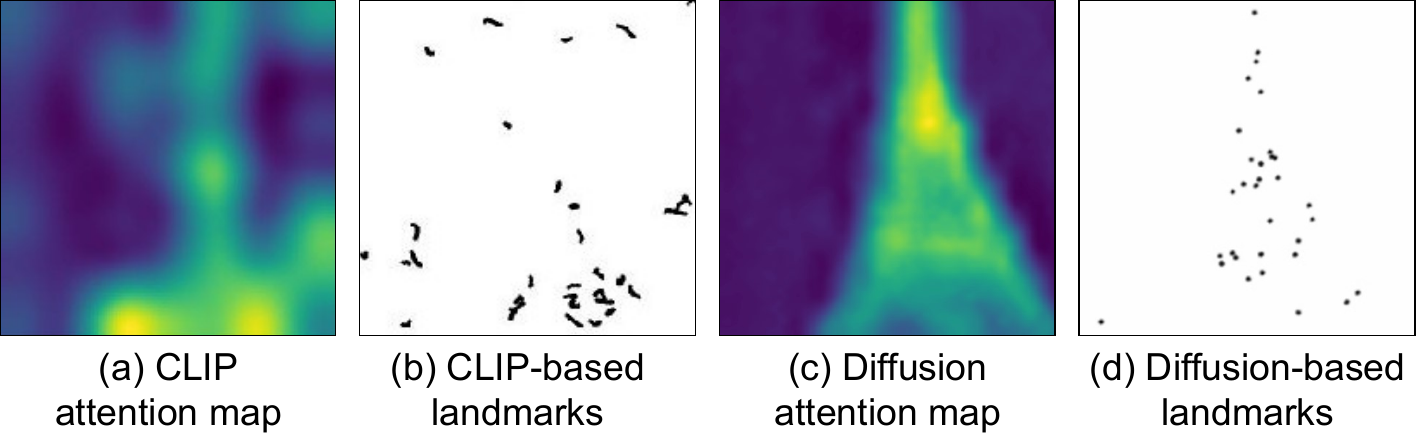}
    % \vspace{-1pt}
    \caption{CLIP and diffusion attention maps and initializations.}
    % \vspace{-15pt}
    \label{fig:canvas-init}
\end{figure}

\myparagraph{CLIP-based vs Diffusion-based Initialization:} The non-convex nature of the optimization in CLIPDraw++ is sensitive to how primitives are initialized. We explore two methods: one using the CLIP attention map (\cref{fig:canvas-init} (a)) and the other based on the latent diffusion model's attention map (\cref{fig:canvas-init} (c)). Local maxima in these maps (\cref{fig:canvas-init} (b) and \cref{fig:canvas-init} (d)) identify key landmark points. Our findings show that the CLIP attention map lacks precision, spreading focus across irrelevant regions, while the diffusion model's attention offers more localized and detailed information. This precision is why we prefer the diffusion-based attention map for initializing sketches.

\begin{figure}[!ht]
% \vspace{-7pt}
\centering
\includegraphics[width=0.49\textwidth]{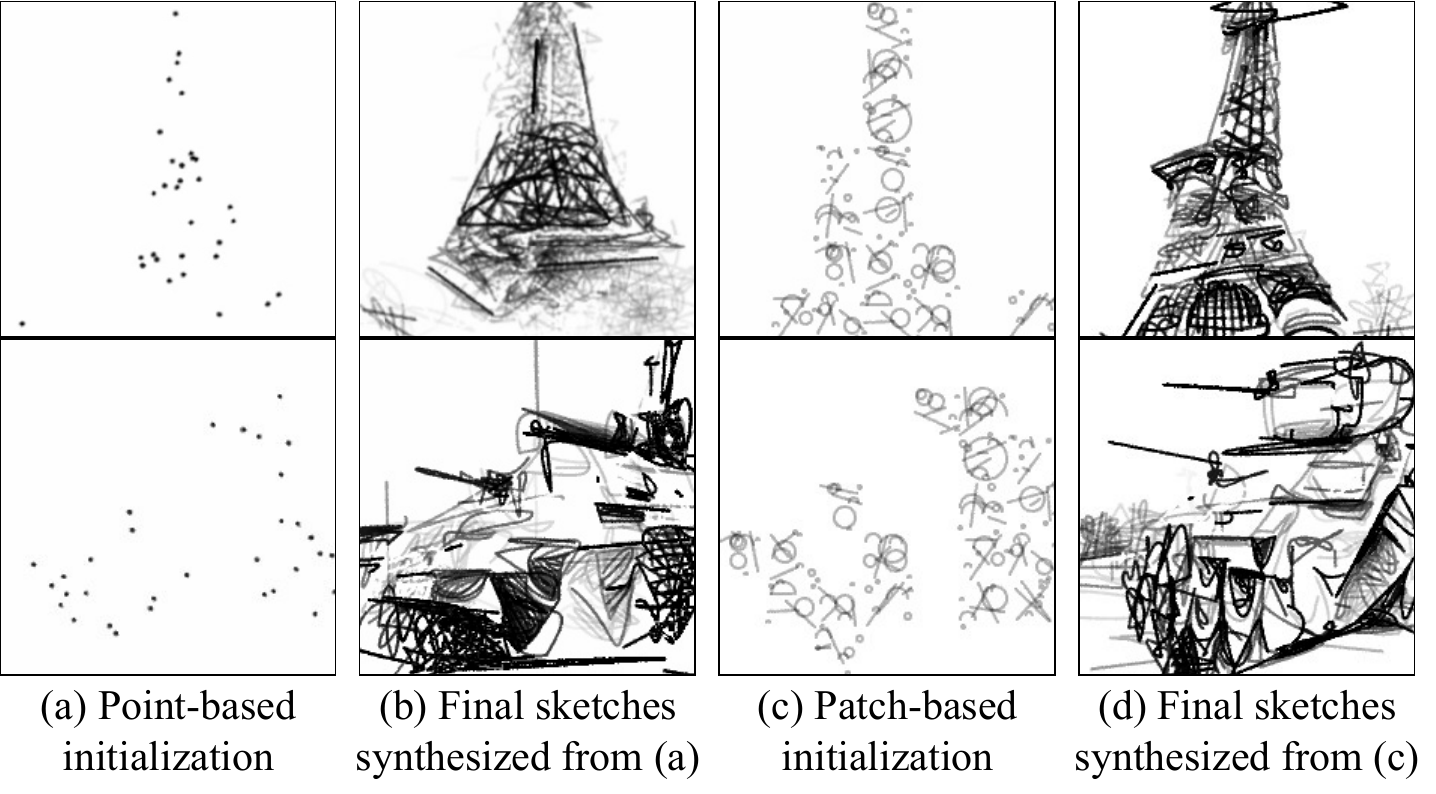}
% \vspace{-1pt}
\caption{Comparison of results for `Eiffel Tower' and `Tank' for point-based and patch-based initialization.}
% \vspace{-10pt}
\label{fig:patch-init}
\end{figure}

\myparagraph{Patch-based Initialization:}
In CLIPDraw++, we use a patch-based approach for stroke initialization, placing primitives within patches rather than directly at landmark points on the attention map. This method prevents the clutter and messiness typical of point-based initialization. As shown in \cref{fig:patch-init}, sketches created with patch-based initialization (\cref{fig:patch-init} (d)) are clearer and more coherent compared to those from point-based initialization (\cref{fig:patch-init} (b)). Distributing primitives within a set range of attention local maxima (based on patch size) avoids excessive constraints and helps maintain clarity. This gives the optimizer a clearer view of the canvas, enabling more effective retention, evolution, or removal of primitives, resulting in cleaner sketches.

\begin{figure}[h]
    \centering
    % \vspace{-5pt}
    \includegraphics[width=\linewidth]{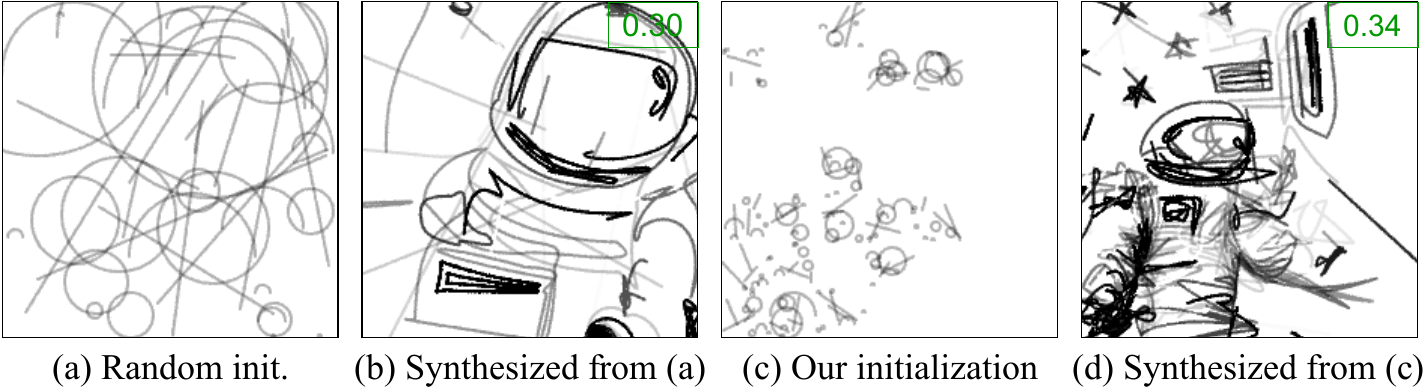}
    % \vspace{1pt}
    \caption{Random vs. patch-based (Ours) strokes initialization.}
    \label{fig:random_vs_patch_init}
\end{figure}

In \cref{fig:random_vs_patch_init}, we demonstrate the superiority of our patch-based initialization compared to random initialization, with the former adhering to our strategy and the latter to the CLIPDraw approach for initially setting up canvases. Both qualitatively and quantitatively, our patch-based method outperforms the random approach.

\begin{figure}[h]
    \centering
    \includegraphics[width=\linewidth]{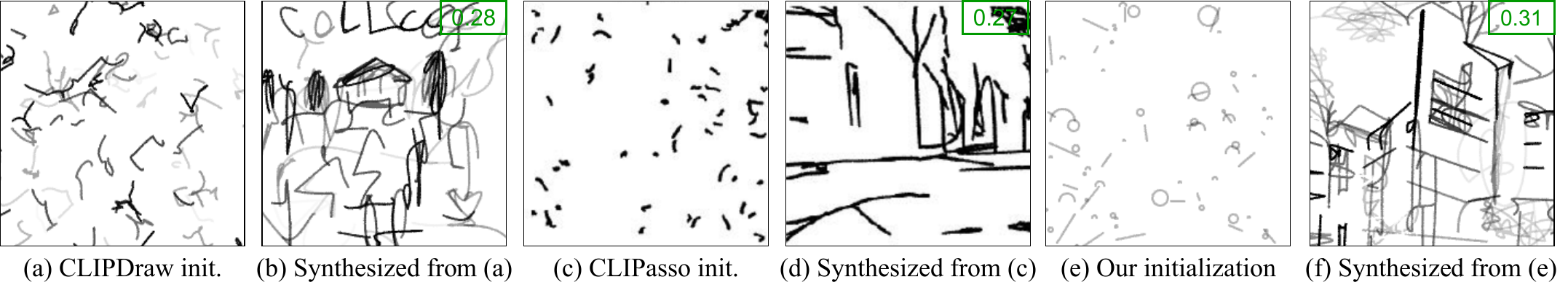}
    \caption{Our patch-based initialization in comparison with CLIPDraw and CLIPasso initialization with the same number of strokes.}
    \label{fig:clipasso_vs_patch_init}
\end{figure}

\begin{figure*}[!h]
\centering
\includegraphics[width=0.99\textwidth]{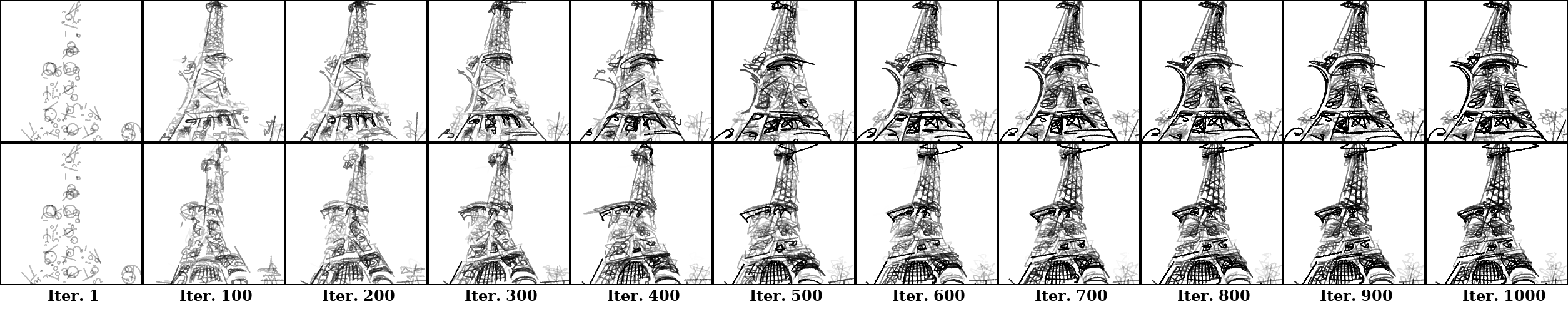}
\caption{Effectiveness of primitive-level dropout (PLD) for the text prompt ``Detailed sketch of \colorbox{Goldenrod}{Eiffel Tower}''. The sketches in the top row are synthesized without PLD, while the ones in the bottom row are synthesized with PLD.}
\vspace{-5pt}
\label{fig:pld-eiffel}
\end{figure*}

In \cref{fig:clipasso_vs_patch_init}, we compare our patch-based initialization against the random initialization of CLIPDraw \cite{Frans2022CLIPDraw} and CLIPasso \cite{Vinker2022CLIPasso}, using an equal number of Bézier curves/primitives for canvas initialization. Our patch-based technique surpasses CLIPDraw and CLIPasso in both quality and performance, even with the same number of strokes.

% \abhra{From here:}
% The patch-based process enhances the canvas's overall granularity, which is pivotal in facilitating a more accurate and effective extraction and portrayal of semantic content within the model. By smoothing out the distribution of strokes across the canvas, this method ensures that each part of the sketch contributes meaningfully to the overall image, enhancing both its aesthetic appeal and its fidelity to the source material.
% \abhra{To here. The above sentences in summary say that patch-based initialization leads to better sketches. But the justification for why that is the case lacks precision. Also, the statement ``By smoothing out the distribution of strokes across the canvas" is not really correct, as patch-based initialization smooths out the distribution only within local regions, and not across the canvas. Suggested rewrite: ``

\begin{figure*}[!h]
% \vspace{-18pt}
    \centering
    \includegraphics[width=0.99\textwidth]{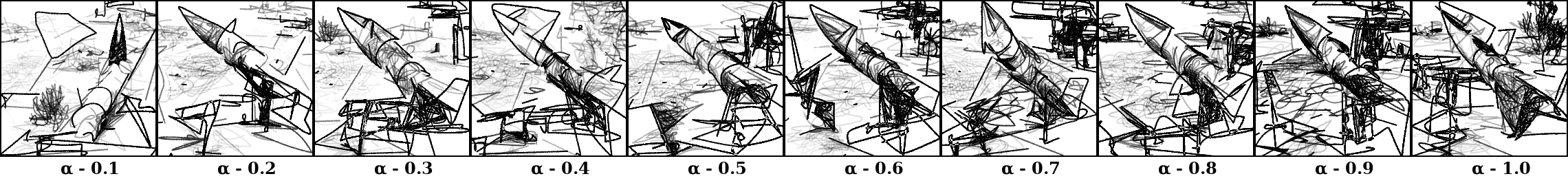}
    \caption{Effectiveness of initializing primitives with diminished opacity, indicated by lower $\alpha$ values, is notable. Initiating primitives with lower $\alpha$ in the prompt ``A \colorbox{Goldenrod}{missile} ready for launch'' yields cleaner final sketches compared to higher $\alpha$.}
    \label{fig:dim-opacity}
    \vspace{-6pt}
\end{figure*}

\begin{figure}[h]
\vspace{-5pt}
\begin{center}
\includegraphics[width=0.465\textwidth]{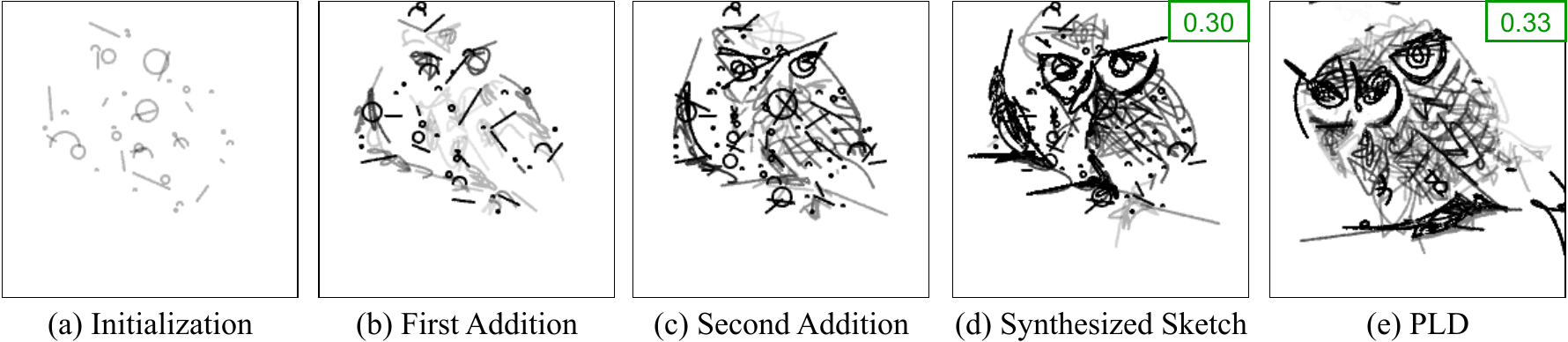}
\end{center}
\vspace{-5pt}
\caption{Comparison of efficacy of PLD with the strategy of sequentially adding extra primitives as required.}
\label{fig:pld_adding_primi}
\vspace{-7pt}
\end{figure}

\myparagraph{Primitive-level Dropout:} Primitive-level dropout enhances the use of primitives by ensuring each encodes a specific concept, reducing noisy strokes that add no meaning. PLD also speeds up convergence by quickly identifying relevant strokes. As shown in \cref{fig:pld-eiffel}, sketches with PLD (bottom row) are cleaner and more realistic, while those without PLD (top row) are noisier. To further assess PLD's effectiveness, we compared it with a strategy of gradually adding new strokes during optimization. In \cref{fig:pld_adding_primi}, ablation shows that PLD outperforms this approach in both visual quality and CLIP-T score. We suspect that abruptly adding new primitives disrupts the loss landscape's smoothness, making optimization harder and resulting in noisier sketches. More results are in Sec. C.1 and Fig. 6 of the \supp.

% Primitive-level dropout enhances the utilization of primitive shapes. It facilitates the optimization process in a way that forces every primitive to independently encode some specific concept in the sketch. As a result, the number of noisy strokes in the sketch, that do not contribute towards depicting anything meaningful, is reduced. Moreover, PLD expedites convergence by reducing the number of steps required for structure creation, thanks to its ability to promptly detect relevant strokes. As demonstrated in \cref{fig:pld-eiffel}, the sketches synthesized with PLD (bottom row) are clean and realistic, while the ones synthesized without PLD (top row) are noisy. 

% To better showcase the effectiveness of our proposed PLD, we compare it with a more intuitive solution by providing limited number of strokes at the early steps of optimization and adding new strokes at the last steps if necessary until the resulting sketch looks plausible. In \cref{fig:pld_adding_primi}, we conduct this ablation study on PLD by controlling the number of primitives, adding more if needed later, which demonstrated PLD's superiority in both visual quality and CLIP-T score. We conjecture that the abrupt introduction of new primitives into the canvas disrupts the smoothness of the loss landscape (by violating Lipschitz continuity), which the optimizer cannot cope with, resulting in noisier sketches. More results on the effectiveness of PLD are in Sec. 3.1 and Fig. 6 of the supplementary.

% \begin{wrapfigure}{r}{0.5\textwidth}

\begin{figure}[h]
% \vspace{-5pt}
\begin{center}
\includegraphics[width=\linewidth]{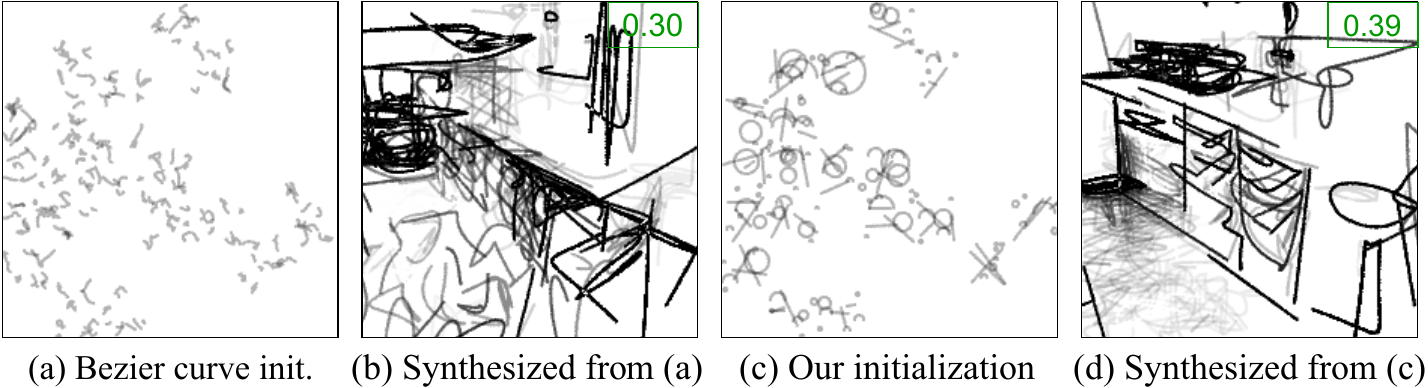}
\end{center}
\vspace{-7pt}
\caption{Advantage of using primitives over B\'{e}zier curves.}
\label{fig:primitive_vs_bezier}
% \vspace{-pt}
\end{figure}

\myparagraph{Advantage of Primitives over B\'{e}zier Curves:} In \cref{fig:primitive_vs_bezier}, we compare the effectiveness of using primitive shapes like straight lines, circles, and semi-circles, with B\'{e}zier curves, where we use B\'{e}zier curves in combination with our remaining novel methodological components, and observe that the B\'{e}zier curves still result in noisier sketches (lower CLIP-T scores) compared to our primitives. We conjecture that since B\'{e}zier curves are more general objects, they suffer from the lack of any geometric inductive bias (which, in this case, is that, sketches can just as well be expressed via simpler primitives like straight lines, circles, and semi-circles). Therefore, they require wastefully more strokes, making the synthesis task \emph{harder to optimize}, and one that results in significantly more \emph{noisy} sketches.

\myparagraph{Sketching with Diminished Opacity:} We start with highly transparent strokes (low opacity or a low $\alpha$ value) and gradually increase the opacity of only the essential strokes needed to convey the text prompt's semantics.
In addition to primitive-level dropout, this approach further minimizes the presence of superfluous strokes in the synthesized sketches. As shown in \cref{fig:dim-opacity}, the final sketches initialized with lower $\alpha$ values are less noisy compared to the ones that are initialized with higher $\alpha$. By mimicking human drawing behaviour in this way, our approach demonstrates the potential to yield sketches that are more precise and finely crafted than those produced by conventional methods.

% Conclusions
% \vspace{-2pt}
\tocless{
\section{Conclusion}
}

% In this work, we have introduced the notion of explainable sketch synthesis through optimization, a process that begins with a canvas featuring simple geometric primitives like straight lines, circles, and semicircles, thereby enhancing the transparency of sketch creation in the realm of explainable AI.
% 
% Our solution to this innovative task is the CLIPDraw++ model. Its core innovations, including the synthesis of highly expressive sketches in an explainable manner via simple linear transformations (computed through optimization) on such  primitives,
% % novel approach to explainable sketch generation,
% strategic sketch canvas initialization for synthesizing clean sketches, and the introduction of primitive-level dropout for producing sketches
% with low noise,
% % only with necessary strokes,
% collectively enhance the model's efficiency and output quality.
% 
We introduced CLIPDraw++, a model for sketch synthesis through optimization using simple geometric primitives like straight lines, circles, and semicircles. 
Our model creates highly expressive sketches through simple linear transformations on these primitives, incorporating techniques such as strategic sketch canvas initialization for synthesizing clean sketches and primitive-level dropout (PLD) for producing sketches with low noise, collectively enhance the model's efficiency and output quality. 
The extensive experiments and ablation studies underscore the model's superiority over existing methods, showcasing its ability to produce aesthetically appealing and semantically rich sketches. CLIPDraw++ excels in AI-driven art creation by merging advanced optimization with intuitive design principles.

\clearpage

{
\small
\bibliographystyle{ieeenat_fullname}
\bibliography{main}

\begin{thebibliography}{34}
\providecommand{\natexlab}[1]{#1}
\providecommand{\url}[1]{\texttt{#1}}
\expandafter\ifx\csname urlstyle\endcsname\relax
  \providecommand{\doi}[1]{doi: #1}\else
  \providecommand{\doi}{doi: \begingroup \urlstyle{rm}\Url}\fi

\bibitem[Alaniz et~al.(2022)Alaniz, Mancini, Dutta, Marcos, and Akata]{Alaniz2022PMN}
Stephan Alaniz, Massimiliano Mancini, Anjan Dutta, Diego Marcos, and Zeynep Akata.
\newblock Abstracting sketches through simple primitives.
\newblock In \emph{ECCV}, 2022.

\bibitem[Bhunia et~al.(2022)Bhunia, Khan, Cholakkal, Anwer, Khan, Laaksonen, and Felsberg]{Bhunia2022DoodleFormer}
Ankan~Kumar Bhunia, Salman Khan, Hisham Cholakkal, Rao~Muhammad Anwer, Fahad~Shahbaz Khan, Jorma Laaksonen, and Michael Felsberg.
\newblock {DoodleFormer: Creative Sketch Drawing with Transformers}.
\newblock In \emph{ECCV}, 2022.

\bibitem[Chan et~al.(2022)Chan, Durand, and Isola]{Chan2022LineDrawings}
Caroline Chan, Frédo Durand, and Phillip Isola.
\newblock Learning to generate line drawings that convey geometry and semantics.
\newblock In \emph{CVPR}, 2022.

\bibitem[Chen et~al.(2020)Chen, Su, Gao, Xia, and Fu]{Chen2020DeepFaceDrawing}
Shu-Yu Chen, Wanchao Su, Lin Gao, Shihong Xia, and Hongbo Fu.
\newblock Deepfacedrawing: Deep generation of face images from sketches.
\newblock \emph{ACM Trans. Graph.}, 2020.

\bibitem[Frans et~al.(2022)Frans, Soros, and Witkowski]{Frans2022CLIPDraw}
Kevin Frans, Lisa Soros, and Olaf Witkowski.
\newblock {CLIPDraw: Exploring Text-to-Drawing Synthesis through Language-Image Encoders}.
\newblock In \emph{NeurIPS}, 2022.

\bibitem[Galatolo et~al.(2021)Galatolo, Cimino, and Vaglini]{Galatolo2021Caption2Image}
Federico Galatolo, Mario Cimino, and Gigliola Vaglini.
\newblock {Generating Images from Caption and Vice Versa via CLIP-Guided Generative Latent Space Search}.
\newblock In \emph{ICIPVE}, 2021.

\bibitem[Gao et~al.(2022)Gao, Man, and Wang]{Gao2022AIArt}
Ze Gao, Sihuang Man, and Anqi Wang.
\newblock Ai art and design creation industry: The transformation from individual production to human-machine symbiosis.
\newblock In \emph{WAC}, 2022.

\bibitem[Ha and Eck(2018)]{Ha2018NeuralSketch}
David Ha and Douglas Eck.
\newblock {A Neural Representation of Sketch Drawings}.
\newblock In \emph{ICLR}, 2018.

\bibitem[Hore and Ziou(2010)]{hore2010image}
Alain Hore and Djemel Ziou.
\newblock Image quality metrics: Psnr vs. ssim.
\newblock In \emph{ICPR}, 2010.

\bibitem[Huang et~al.(2023)Huang, Sun, Xie, Li, and Liu]{huang2024t2i}
Kaiyi Huang, Kaiyue Sun, Enze Xie, Zhenguo Li, and Xihui Liu.
\newblock T2i-compbench: A comprehensive benchmark for open-world compositional text-to-image generation.
\newblock \emph{NeurIPS}, 36, 2023.

\bibitem[Jain et~al.(2023)Jain, Xie, and Abbeel]{Jain2023VectorFusion}
Ajay Jain, Amber Xie, and Pieter Abbeel.
\newblock {VectorFusion: Text-to-SVG by Abstracting Pixel-Based Diffusion Models}.
\newblock In \emph{CVPR}, 2023.

\bibitem[Jumper et~al.(2021)Jumper, Evans, Pritzel, Green, Figurnov, Ronneberger, Tunyasuvunakool, Bates, Žídek, Potapenko, Bridgland, Meyer, Kohl, Ballard, Cowie, Romera-Paredes, Nikolov, Jain, Adler, Back, Petersen, Reiman, Clancy, Zielinski, Steinegger, Pacholska, Berghammer, Bodenstein, Silver, Vinyals, Senior, Kavukcuoglu, Kohli, and Hassabis]{Jumper2021Protein}
John Jumper, Richard Evans, Alexander Pritzel, Tim Green, Michael Figurnov, Olaf Ronneberger, Kathryn Tunyasuvunakool, Russ Bates, Augustin Žídek, Anna Potapenko, Alex Bridgland, Clemens Meyer, Simon A.~A. Kohl, Andrew~J. Ballard, Andrew Cowie, Bernardino Romera-Paredes, Stanislav Nikolov, Rishub Jain, Jonas Adler, Trevor Back, Stig Petersen, David Reiman, Ellen Clancy, Michal Zielinski, Martin Steinegger, Michalina Pacholska, Tamas Berghammer, Sebastian Bodenstein, David Silver, Oriol Vinyals, Andrew~W. Senior, Koray Kavukcuoglu, Pushmeet Kohli, and Demis Hassabis.
\newblock {Highly accurate protein structure prediction with AlphaFold}.
\newblock \emph{Nature}, 2021.

\bibitem[Li et~al.(2022)Li, Li, Xiong, and Hoi]{li2022blip}
Junnan Li, Dongxu Li, Caiming Xiong, and Steven Hoi.
\newblock Blip: Bootstrapping language-image pre-training for unified vision-language understanding and generation.
\newblock In \emph{ICML}, 2022.

\bibitem[Li et~al.(2019)Li, Lin, Mech, Yumer, and Ramanan]{Li2019PhotoSketching}
Mengtian Li, Zhe Lin, Radomir Mech, Ersin Yumer, and Deva Ramanan.
\newblock {Photo-Sketching: Inferring Contour Drawings from Images}.
\newblock In \emph{WACV}, 2019.

\bibitem[Li et~al.(2020)Li, Luk\'{a}\v{c}, Micha\"{e}l, and Ragan-Kelley]{Li2022Diffvg}
Tzu-Mao Li, Michal Luk\'{a}\v{c}, Gharbi Micha\"{e}l, and Jonathan Ragan-Kelley.
\newblock Differentiable vector graphics rasterization for editing and learning.
\newblock In \emph{SIGGRAPH Asia}, 2020.

\bibitem[Li et~al.(2017)Li, Song, Hospedales, and Gong]{Li2017SketchSynth}
Yi Li, Yi-Zhe Song, Timothy~M. Hospedales, and Shaogang Gong.
\newblock {Free-Hand Sketch Synthesis with Deformable Stroke Models}.
\newblock \emph{IJCV}, 2017.

\bibitem[Liu et~al.(2020)Liu, Yu, and Yu]{Liu2020UnsupSketch}
Runtao Liu, Qian Yu, and Stella Yu.
\newblock Unsupervised sketch to photo synthesis.
\newblock In \emph{ECCV}, 2020.

\bibitem[Mirowski et~al.(2022)Mirowski, Banarse, Malinowski, Osindero, and Fernando]{Mirowski2022Clipclop}
Piotr Mirowski, Dylan Banarse, Mateusz Malinowski, Simon Osindero, and Chrisantha Fernando.
\newblock {CLIP-CLOP: CLIP-Guided Collage and Photomontage}.
\newblock In \emph{ICCC}, 2022.

\bibitem[Murdock(2021)]{Murdock2021BigGAN}
Ryan Murdock.
\newblock {The Big Sleep: BigGANxCLIP.ipynb}, 2021.

\bibitem[Nguyen et~al.(2015)Nguyen, Yosinski, and Clune]{Nguyen2015DNNFool}
Anh Nguyen, Jason Yosinski, and Jeff Clune.
\newblock {Deep Neural Networks are Easily Fooled: High Confidence Predictions for Unrecognizable Images}.
\newblock In \emph{CVPR}, 2015.

\bibitem[Nguyen et~al.(2016)Nguyen, Dosovitskiy, Yosinski, Brox, and Clune]{Nguyen2016ImageSynthGAN}
Anh Nguyen, Alexey Dosovitskiy, Jason Yosinski, Thomas Brox, and Jeff Clune.
\newblock {Synthesizing the preferred inputs for neurons in neural networks via deep generator networks}.
\newblock In \emph{NIPS}, 2016.

\bibitem[Nguyen et~al.(2017)Nguyen, Clune, Bengio, Dosovitskiy, and Yosinski]{Nguyen2017PlugPlayGAN}
Anh Nguyen, Jeff Clune, Yoshua Bengio, Alexey Dosovitskiy, and Jason Yosinski.
\newblock {Plug \& Play Generative Networks: Conditional Iterative Generation of Images in Latent Space}.
\newblock In \emph{CVPR}, 2017.

\bibitem[Paszke et~al.(2019)Paszke, Gross, Massa, Lerer, Bradbury, Chanan, Killeen, Lin, Gimelshein, Antiga, et~al.]{paszke2019pytorch}
Adam Paszke, Sam Gross, Francisco Massa, Adam Lerer, James Bradbury, Gregory Chanan, Trevor Killeen, Zeming Lin, Natalia Gimelshein, Luca Antiga, et~al.
\newblock Pytorch: An imperative style, high-performance deep learning library.
\newblock In \emph{NeurIPS}, 2019.

\bibitem[Qu et~al.(2023)Qu, Gryaditskaya, Li, Pang, Xiang, and Song]{Qu2023SketchXAI}
Zhiyu Qu, Yulia Gryaditskaya, Ke Li, Kaiyue Pang, Tao Xiang, and Yi-Zhe Song.
\newblock {SketchXAI: A First Look at Explainability for Human Sketches}.
\newblock In \emph{CVPR}, 2023.

\bibitem[Radford et~al.(2021)Radford, Kim, Hallacy, Ramesh, Goh, Agarwal, Sastry, Askell, Mishkin, Clark, et~al.]{Radford2021CLIP}
Alec Radford, Jong~Wook Kim, Chris Hallacy, Aditya Ramesh, Gabriel Goh, Sandhini Agarwal, Girish Sastry, Amanda Askell, Pamela Mishkin, Jack Clark, et~al.
\newblock Learning transferable visual models from natural language supervision.
\newblock In \emph{ICML}, 2021.

\bibitem[Rombach et~al.(2022)Rombach, Blattmann, Lorenz, Esser, and Ommer]{Rombach2022LatentDiffusion}
Robin Rombach, Andreas Blattmann, Dominik Lorenz, Patrick Esser, and Björn Ommer.
\newblock {High-Resolution Image Synthesis with Latent Diffusion Models}.
\newblock In \emph{CVPR}, 2022.

\bibitem[Silver et~al.(2016)Silver, Huang, Maddison, Guez, Sifre, Driessche, Schrittwieser, Antonoglou, Panneershelvam, Lanctot, Dieleman, Grewe, Nham, Kalchbrenner, Sutskever, Lillicrap, Leach, Kavukcuoglu, Graepel, and Hassabis]{Silver2016AlphaGo}
David Silver, Aja Huang, Chris~J. Maddison, Arthur Guez, Laurent Sifre, George van~den Driessche, Julian Schrittwieser, Ioannis Antonoglou, Veda Panneershelvam, Marc Lanctot, Sander Dieleman, Dominik Grewe, John Nham, Nal Kalchbrenner, Ilya Sutskever, Timothy Lillicrap, Madeleine Leach, Koray Kavukcuoglu, Thore Graepel, and Demis Hassabis.
\newblock {Mastering the game of Go with deep neural networks and tree search}.
\newblock \emph{Nature}, 2016.

\bibitem[Srivastava et~al.(2014)Srivastava, Hinton, Krizhevsky, Sutskever, and Salakhutdinov]{Srivastava2014Dropout}
Nitish Srivastava, Geoffrey Hinton, Alex Krizhevsky, Ilya Sutskever, and Ruslan Salakhutdinov.
\newblock Dropout: A simple way to prevent neural networks from overfitting.
\newblock \emph{JMLR}, 2014.

\bibitem[Tang et~al.(2022)Tang, Pandey, Jiang, Yang, Kumar, Lin, and Ture]{tang2022daam}
Raphael Tang, Akshat Pandey, Zhiying Jiang, Gefei Yang, Karun Kumar, Jimmy Lin, and Ferhan Ture.
\newblock What the daam: Interpreting stable diffusion using cross attention.
\newblock \emph{arXiv preprint arXiv:2210.04885}, 2022.

\bibitem[Tong et~al.(2020)Tong, Chen, Ni, and Wang]{Tong2020SketchGenVectorFlow}
Zhengyan Tong, Xuanhong Chen, Bingbing Ni, and Xiaohang Wang.
\newblock {Sketch Generation with Drawing Process Guided by Vector Flow and Grayscale}.
\newblock In \emph{AAAI}, 2020.

\bibitem[Vinker et~al.(2022{\natexlab{a}})Vinker, Alaluf, Cohen-Or, and Shamir]{Vinker2022CLIPascene}
Yael Vinker, Yuval Alaluf, Daniel Cohen-Or, and Ariel Shamir.
\newblock {CLIPascene: Scene Sketching with Different Types and Levels of Abstraction}.
\newblock In \emph{arXiv}, 2022{\natexlab{a}}.

\bibitem[Vinker et~al.(2022{\natexlab{b}})Vinker, Pajouheshgar, Bo, Bachmann, Bermano, Cohen-Or, Zamir, and Shamir]{Vinker2022CLIPasso}
Yael Vinker, Ehsan Pajouheshgar, Jessica~Y. Bo, Roman~Christian Bachmann, Amit~Haim Bermano, Daniel Cohen-Or, Amir Zamir, and Ariel Shamir.
\newblock {CLIPasso: Semantically-Aware Object Sketching}.
\newblock In \emph{SIGGRAPH}, 2022{\natexlab{b}}.

\bibitem[Xing et~al.(2023)Xing, Wang, Zhou, Zhang, Yu, and Xu]{xing2023diffsketcher}
XiMing Xing, Chuang Wang, Haitao Zhou, Jing Zhang, Qian Yu, and Dong Xu.
\newblock Diffsketcher: Text guided vector sketch synthesis through latent diffusion models.
\newblock In \emph{NeurIPS}, 2023.

\bibitem[Xing et~al.(2024)Xing, Zhou, Wang, Zhang, Xu, and Yu]{xing2024svgdreamer}
Ximing Xing, Haitao Zhou, Chuang Wang, Jing Zhang, Dong Xu, and Qian Yu.
\newblock Svgdreamer: Text guided svg generation with diffusion model.
\newblock In \emph{CVPR}, 2024.

\end{thebibliography}
}

% \clearpage

\maketitlesupplementary

\renewcommand\thesection{\Alph{section}}
\renewcommand\thesubsection{\thesection.\arabic{subsection}}

\renewcommand{\cftsecleader}{\cftdotfill{\cftdotsep}}
\renewcommand{\contentsname}{Table of Contents}
\renewcommand{\cftaftertoctitle}{\vspace{-8pt}\par\noindent\rule{\linewidth}{1pt}\vspace{-5pt}\par}

\renewcommand{\cftpnumalign}{l}
\setlength{\cftsecindent}{10pt}
\setlength{\cftsubsecindent}{30pt}
\setlength{\cftsecnumwidth}{17pt}
\tableofcontents

\setcounter{section}{0}

\setcounter{figure}{0}
\setcounter{table}{0}

\section{Implementation Details}
\label{sec:implementation_details}
In this section, we provide detailed information on how we have implemented our method. The implementation is carried out using \texttt{PyTorch}~\cite{paszke2019pytorch} and makes use of the differentiable rasterizer framework \texttt{diffvg}~\cite{Li2022Diffvg}.

\subsection{Initialization details} 
\textit{``What is the procedure to initialize the first canvas?''} -- to answer this question can be given in the following three steps:
\ding{182} We first extract the crucial salient regions of the canvas by incorporating diffusion attentive attribution maps (DAAM)~\cite{tang2022daam}, leveraging the pre-trained Latent Diffusion model~\cite{Rombach2022LatentDiffusion}. DAAM upscale and aggregate cross-attention word–pixel scores in the denoising subnetwork. For more details please refer to ~\cite{tang2022daam}.
\ding{183} From the DAAM-generated final attention map, we sample $k$ positions. Here we have considered the value of $k$ to be 32. Thereafter, we follow patch-wise initialization as described within Sec. \mainsec{3.2} in the main paper. Here, we divide the whole canvas into $32\times32$ patches and select only patches where any of $k$ points lie within, we do not initialize the primitives within these patches where there are no points belonging to them. A patch is denoted by $P_{i, j}$, where $i$ and $j$ represent the row and column, respectively. Here, $\forall (x,y) \in P_{i, j}$ lies within $(x^s_{i,j}, y^s_{i,j})$, and $(x^e_{i,j}, y^e_{i,j})$ where $s$ denotes the top-left/start point and $e$ denotes bottom-right/end point. A point, ($x, y$) belongs to patch $P_{i', j'}$ where $(i', j') = (\left[x/32\right], \left[y/32\right])$.
% $$ patch_x = x/patch\_size $$
% $$ patch_y = y/size(patch) $$
% $$
% patch = patch_x + patch_y \times N
% $$
% $N$: total patches
\ding{184} After choosing which patches to initialize the primitives, we randomly place exactly 3 primitives, (one from each type, straight line, circle, and semi-circle) within selected patches. Our primitives are based on the SVG (Scalable Vector Graphics) path but constrained to certain geometric shapes, described as follows:

\begin{itemize}
\item \textbf{\textit{Line:}} A line can be represented by two control points, \textsc{$l_{1} (x_1, y_1)$} and \textsc{$l_{2} (x_2, y_2)$}. To initialize a line, $L: (l_1, l_2)$ at the desired patch, $P_{i,j}$ we follow:
\begin{align*}
    x_1 = \mathtt{random.randint}(x^s_{i,j}, x^e_{i,j})\\
    y_1 = \mathtt{random.randint}(y^s_{i,j}, y^e_{i,j})\\
    x_2 = \mathtt{random.randint}(x^s_{i,j}, x^e_{i,j})\\
    y_2 = \mathtt{random.randint}(y^s_{i,j}, y^e_{i,j})
\end{align*}
The length of line $L = \sqrt{(x_2 - x_1)^2 + (y_2 - y_1)^2}$.
    
\item \textbf{\textit{Circle:}} A circle, $C$ within a patch $P_{i,j}$ is represented by it's centre $\mathbf{c}\:(x_{c}, y_{c})$, and the radius $\mathbf{r}$ where
\begin{align*}
    x_{c} = \mathtt{random.randint}(x^s_{i,j}, x^e_{i,j}) \\
    y_{c} = \mathtt{random.randint}(y^s_{i,j}, y^e_{i,j}) \\
    r = \mathtt{random.randint}(1, r_{max}).
\end{align*}
Here $r_{max}$ is the maximum size of the radius so that the circle should remain within the patch, $P_{i,j}$ which is defined in the below pseudo-code:
\begin{lstlisting}[language=Python]
# randomly initialize (x_c, y_c)
max_radius_x = patch_size/2
max_radius_y = patch_size/2
if x_c < start_x + patch_size/2:
    max_radius_x = x_c - start_x
else:
    max_radius_x = end_x - x_c
if y_c < start_y + patch_size/2:
    max_radius_y = y_c - start_y
else:
    max_radius_y = end_y - y_c
max_radius = min(max_radius_x, max_radius_y)
\end{lstlisting}

where $\text{patch\_size}=32$; (x\_c, y\_c) is the center of the circle; and (start\_x, start\_y) = $\left(x^s_{i,j}, y^s_{i,j}\right)$ and (end\_x, end\_y) = $\left(x^e_{i,j}, y^e_{i,j}\right)$.

\item \textbf{\textit{Semi-circle:}} For the semi-circle, the mathematical explanation remains the same as the circle, the only change that occurs is in the SVG path and these can be approximated using a cubic Bezier curve.
    
\end{itemize}

The SVG path of these primitive types to incorporate \texttt{diffvg} library is given as:
\begin{lstlisting}[language=Python]
import pydiffvg # diffvg library
# SVG line path
line_path = pydiffvg.from_svg_path(f'M {x1},{y1} L {x2},{y2}')
# SVG circle path
circle_path = pydiffvg.from_svg_path(f'M {x_c - r},{y_c} a {r},{r} 0 1,1 {r*2},0 a {r},{r} 0 1,1 {-1*r*2},0')
# SVG semi-circle path
semi_circle_path = pydiffvg.from_svg_path('M {x_c - r},{y_c} a {r},{r} 0 1,1 {r*2},0')
\end{lstlisting}

This way we initialize the first sketch canvas using a set of simple geometric primitives before the optimization process begins.

\subsection{Augmentation details} The primary aim of image augmentation is to preserve recognizability in the presence of various distortions. Following the implementation of CLIPDraw~\cite{Frans2022CLIPDraw}, we utilize a series of transformation functions, namely \texttt{torch.transforms.RandomPerspective} and \texttt{torch.transforms.RandomResizedCrop}, on both the given image and generated sketch before passing them as inputs to the CLIP model for loss computation. In this context, the total number of augmentations, denoted as $N$, is set to 4. The incorporation of these augmentations serves to enhance the robustness of the optimization process against adversarial samples and contributes to an overall improvement in the quality of the generated sketch.

\subsection{Optimization details} Our optimization loop includes the following three steps -- 1) generating a sketch from vectorized primitives, 2) applying primitive-level dropout (PLD), and 3) computing loss function and back-propagate it to SVG parameters within \texttt{diffvg}. CLIPDraw++ optimization does not operate on width and color optimization, we keep them constant for the task of the sketch-synthesize. 
The optimization process typically takes around 2 min to run 1000 iterations on a single RTX 3090 GPU. Nevertheless, we attain a thorough grasp of the semantic and visual aspects of the provided textual prompt within 500 epochs. Yet, for the purpose of refining it and enhancing noise removal, we continue the process for an additional 500 iterations as shown in ~\cref{fig:primitive-full}. We employ the \texttt{Adam} optimizer along with the learning rate scheduler outlined in the following pseudo-code:
\begin{lstlisting}[language=Python]
for t in range(num_iter):
 ...
 if t == int(num_iter * 0.5):
    lr = 0.4
 if t == int(num_iter * 0.75):
    lr = 0.1
 ...
\end{lstlisting}

Here, \texttt{num\_iter} is the maximum number of optimization loops, we set it as 1000; \texttt{lr} is the learning rate, and we initially set it as 1.0.

\begin{tcolorbox}
    The words highlighted in yellow are used to generate the attention maps leveraging DAAM.
\end{tcolorbox}

\section{Sketch Generation from Primitives}
\label{sec:primitive_explanability}
As delineated in Sec. \mainsec{3.1} and Sec. \mainsec{4.1} of the main paper, our sketches are composed of linearly transformed primitives such as straight lines, circles, and semi-circles. These individual strokes can be tracked through their evolution in successive iterations of the optimization process.
% with an aim to enhance explainability by illustrating the individual evolution of each primitive type. 
In this context, we present the visualization of sketch generation with the primitive level tracking in \cref{fig:primitive-full} and overall sketch level tracking in Figs.~\ref{fig:track_1}, \ref{fig:track_2}, \ref{fig:track_3}, and \ref{fig:track_4} after each 100 iterations, contextualized by diverse text prompts.

In the illustrative portrayal of  \textit{``Floating musical notes from a piano''} (refer to \cref{piano}), the composition employs straight lines for the base structure, and a combination of circles and semi-circles for floating musical notes. Analogously, in the depiction of \textit{``Faucet''} (refer to \cref{Faucet}),  straight lines define the wash basin, while the faucet's spout predominantly assumes a circular form, with its juncture to the basin evolving gracefully from semi-circular elements. Turning attention to the \textit{``Supermarket''} scene (refer to \cref{Supermarket}), straight lines form simple structures like shelves, contrasting with circles and semi-circles that compose more intricate elements such as displayed items.

Moreover, we present comprehensive results of sketch synthesis and their traceable versions for various input text prompts in \cref{fig:track_1,fig:track_2,fig:track_3,fig:track_4}.

\begin{figure*}[!htbp]
    \centering
    \subfloat[Floating musical notes from a \highlight{piano}.]{\includegraphics[width=0.85\textwidth]{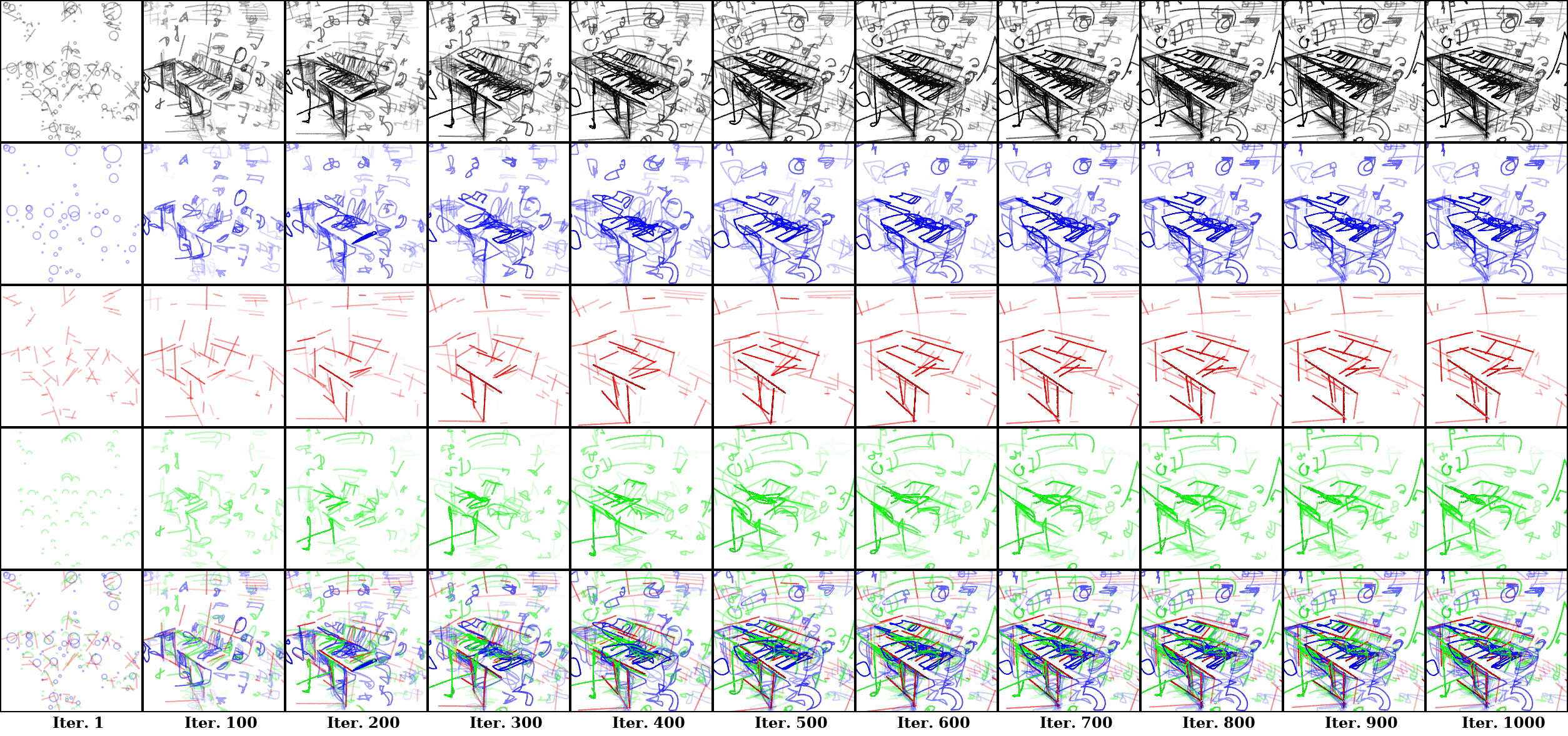} \label{piano}}

    \subfloat[A peaceful afternoon at the \hl{farm} sketch.]{\includegraphics[width=0.85\textwidth]{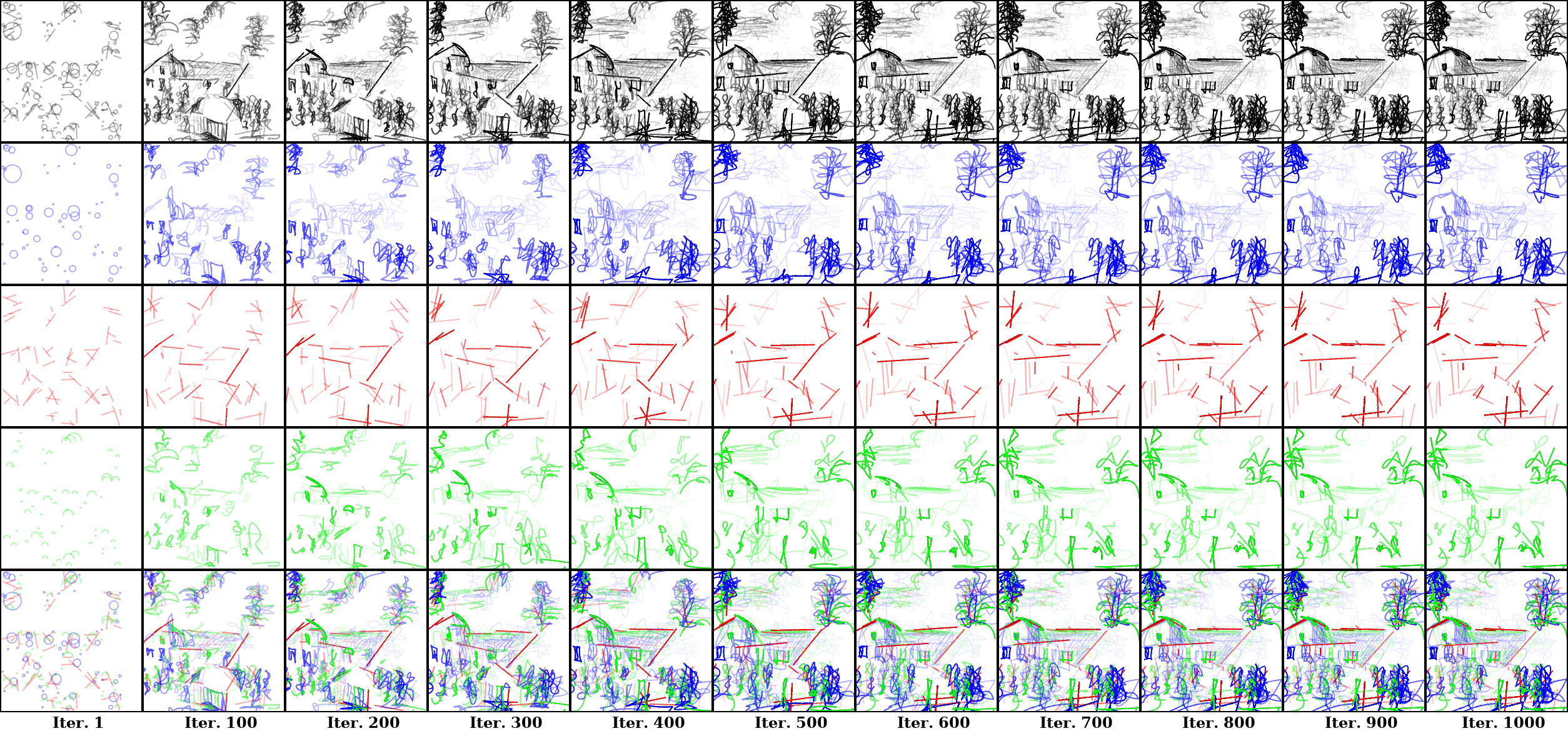}}

    \subfloat[Fast \highlight{food} with soft drinks.]{\includegraphics[width=0.85\textwidth]{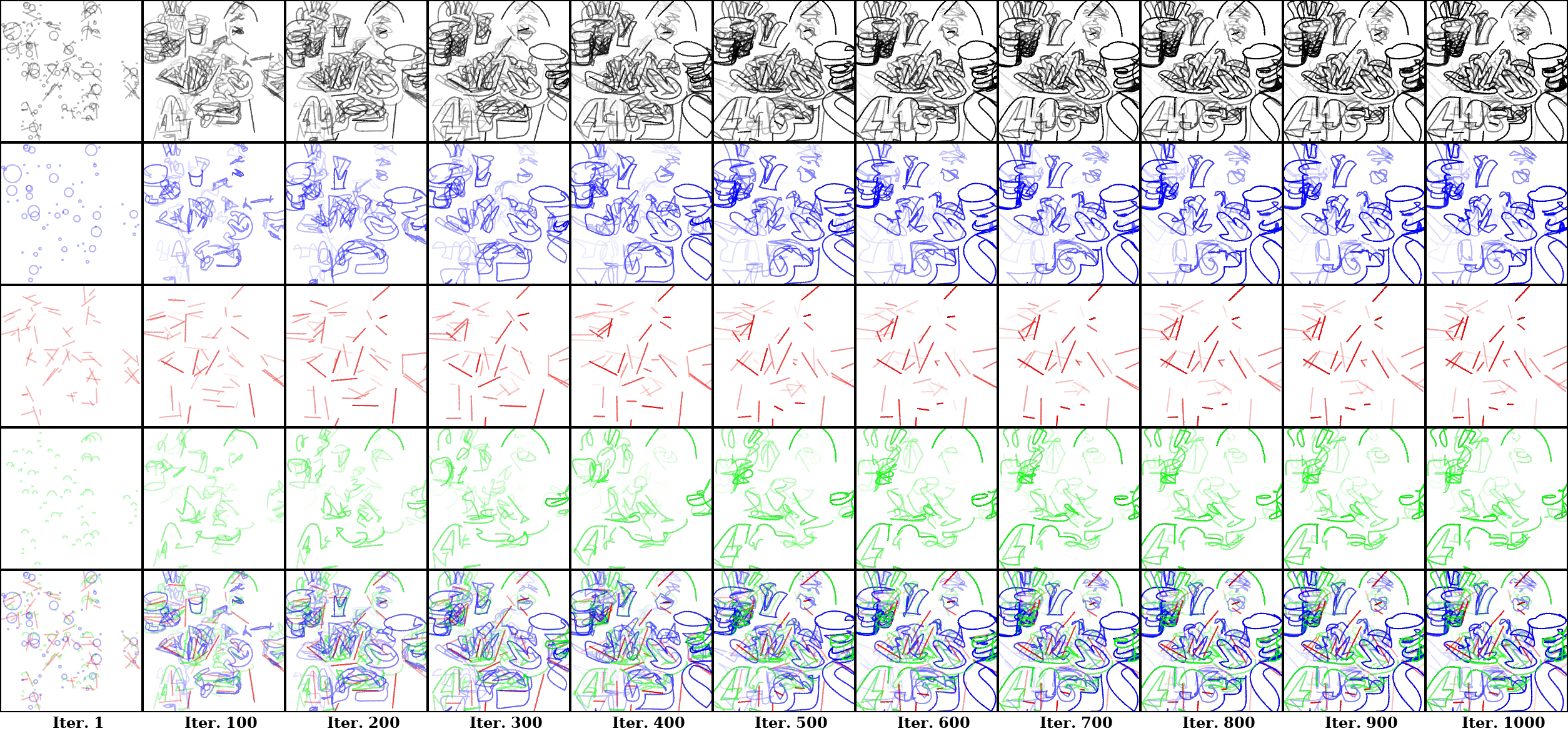}}
\end{figure*}

\begin{figure*}[!htbp]
    \centering
    \ContinuedFloat
    \subfloat[A sleek \hl{faucet} centered in a bathroom.]{\includegraphics[width=0.88\textwidth]{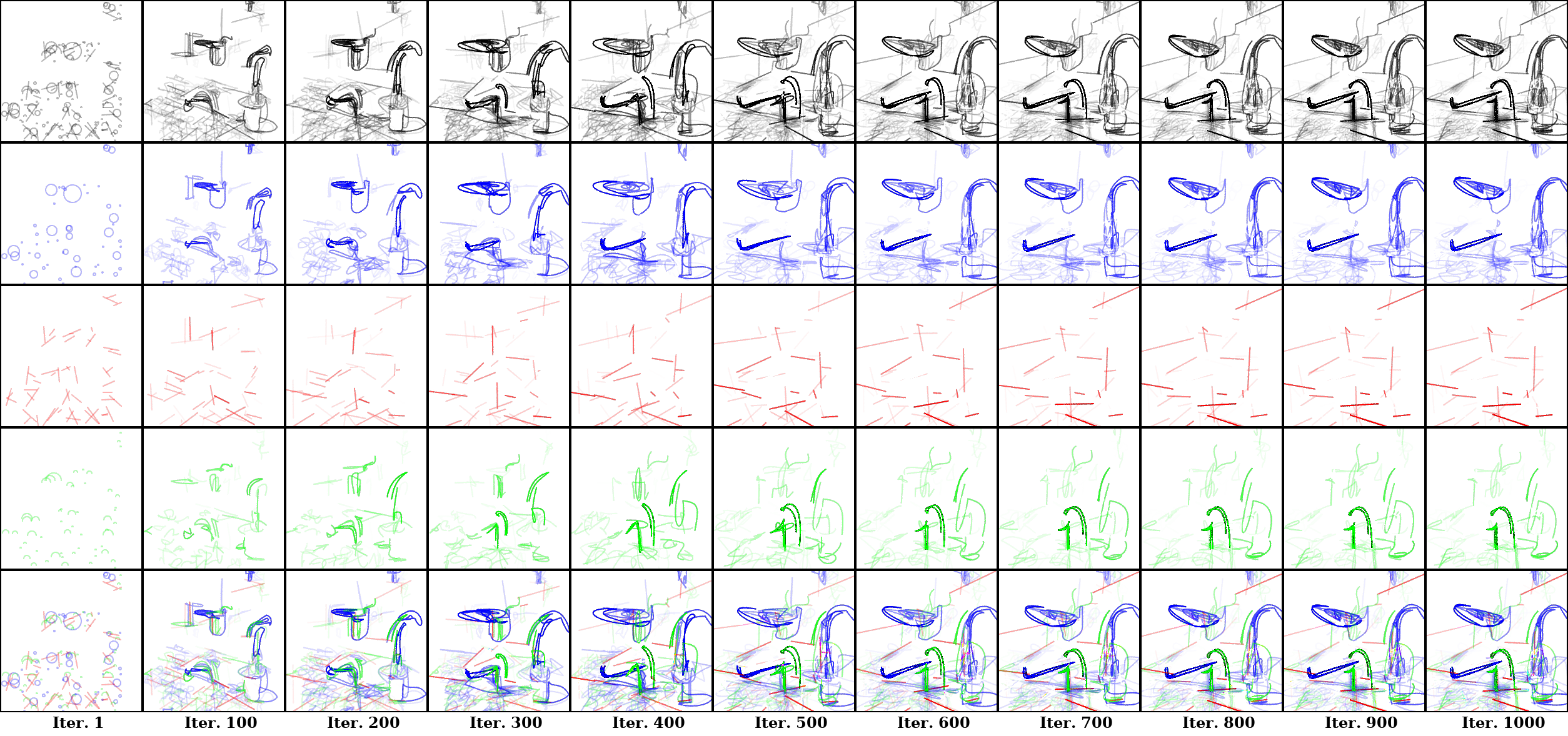} \label{Faucet}}

    \subfloat[Local \hl{supermarket} showcasing products.]{\includegraphics[width=0.88\textwidth]{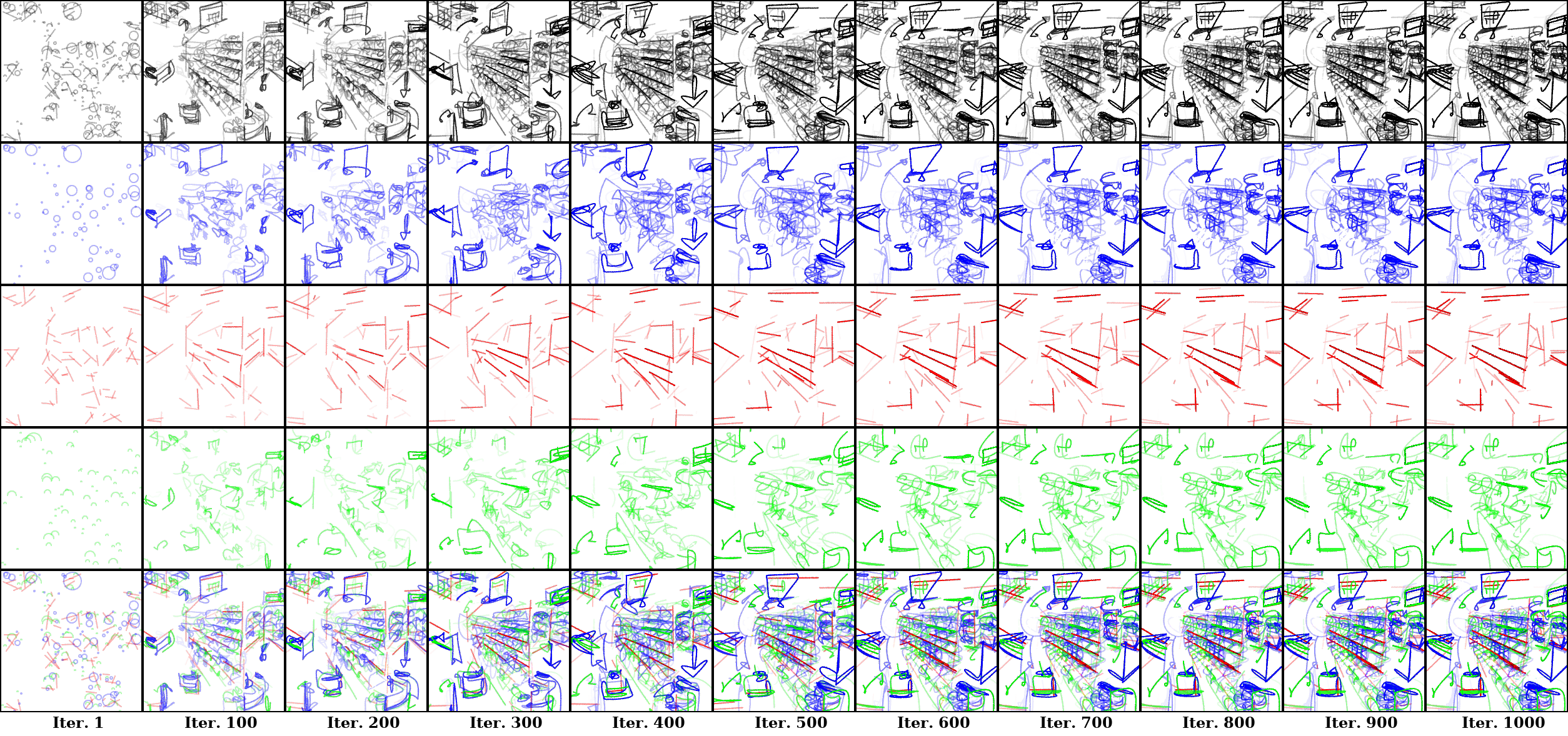} \label{Supermarket}}
    \caption{CLIPDraw++ illustrates the shape evolution of each primitive type in optimization:  first row - black-and-white synthesized sketch, next three rows - \textcolor{blue}{circles}, \textcolor{red}{straight lines}, and \textcolor{green}{semi-circles}, final row - combined compositions.}
    \label{fig:primitive-full}
\end{figure*}

\begin{figure*}[!htbp]
    \centering
    \subfloat[A sketch of a \hl{house} in the woods.]{\includegraphics[width=0.92\textwidth]{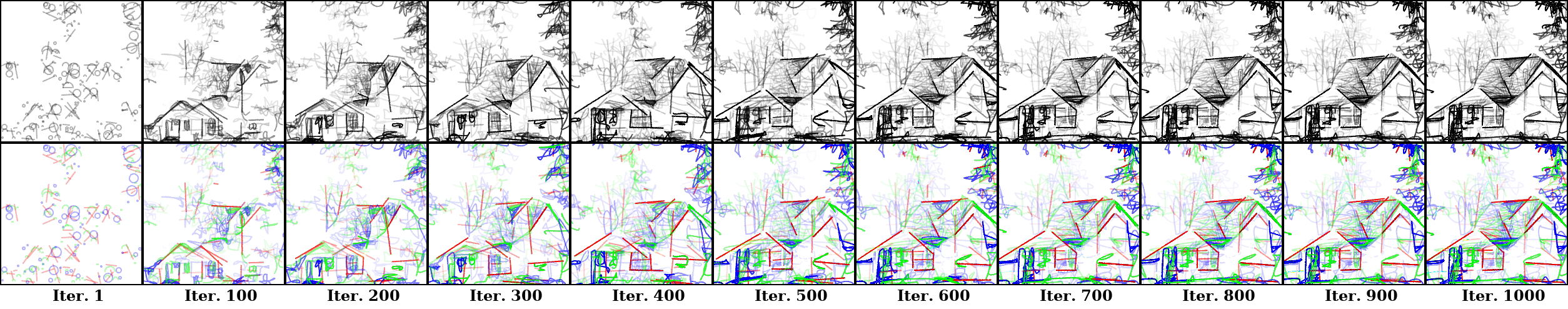}}
    
    \subfloat[A \hl{caravan} adventure life.]{\includegraphics[width=0.92\textwidth]{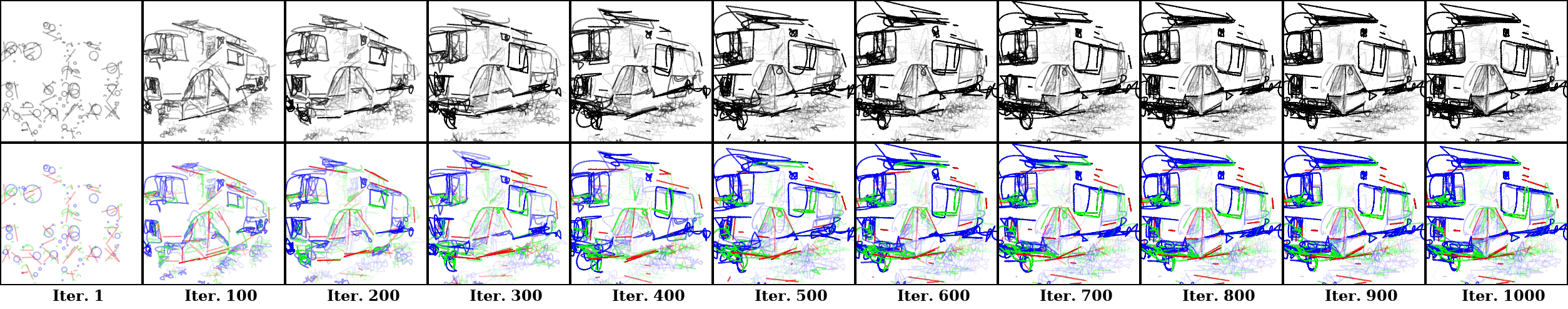}}

    \subfloat[A \hl{sideboard} amidst the room's rhythm.]{\includegraphics[width=0.92\textwidth]{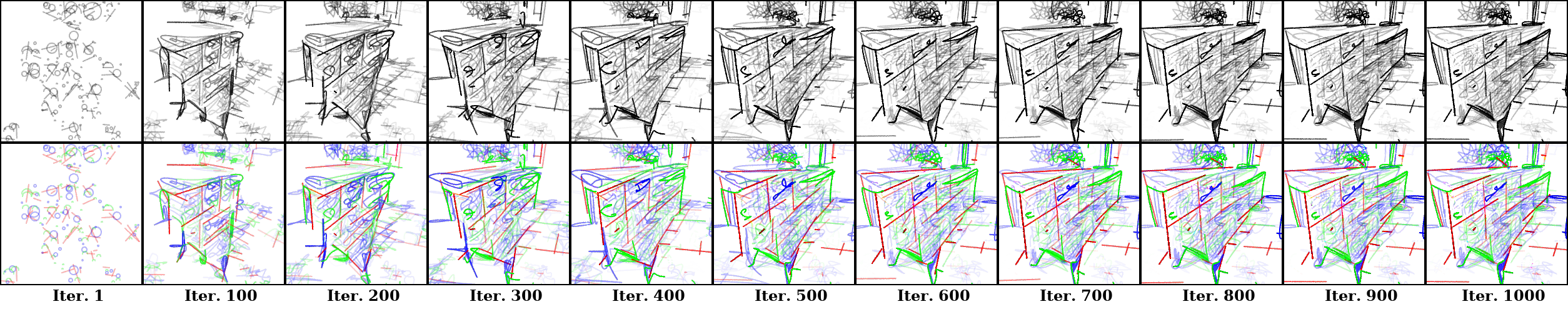}}

    \subfloat[Taste and aroma of \hl{fedelini}.]{\includegraphics[width=0.92\textwidth]{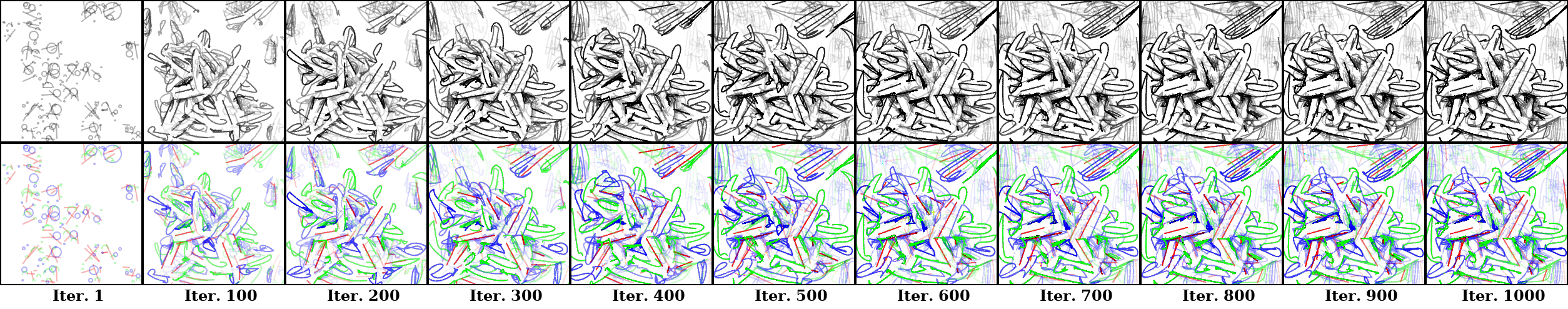}}

    \subfloat[Sketch of eggs laying down a \hl{nest}.]{\includegraphics[width=0.92\textwidth]{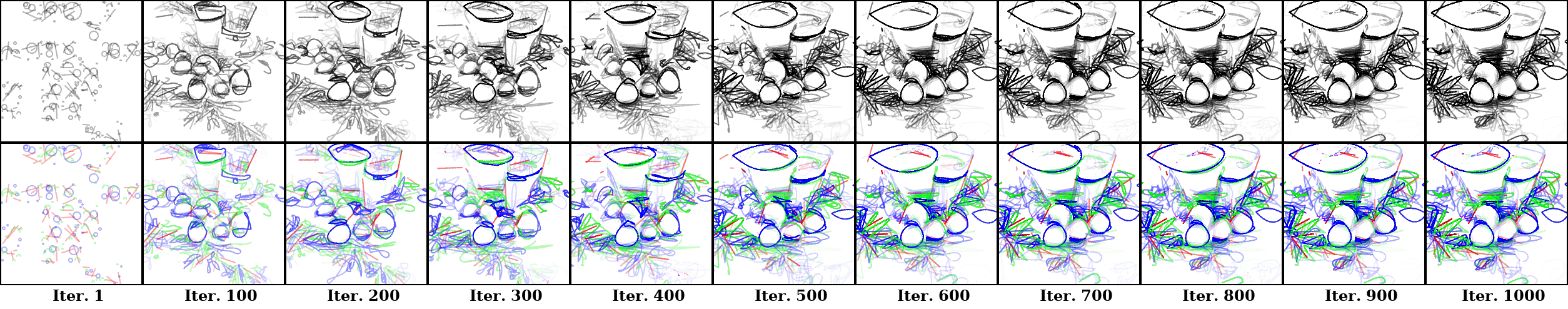}}
    
    \subfloat[A solid \hl{stone}.]{\includegraphics[width=0.92\textwidth]{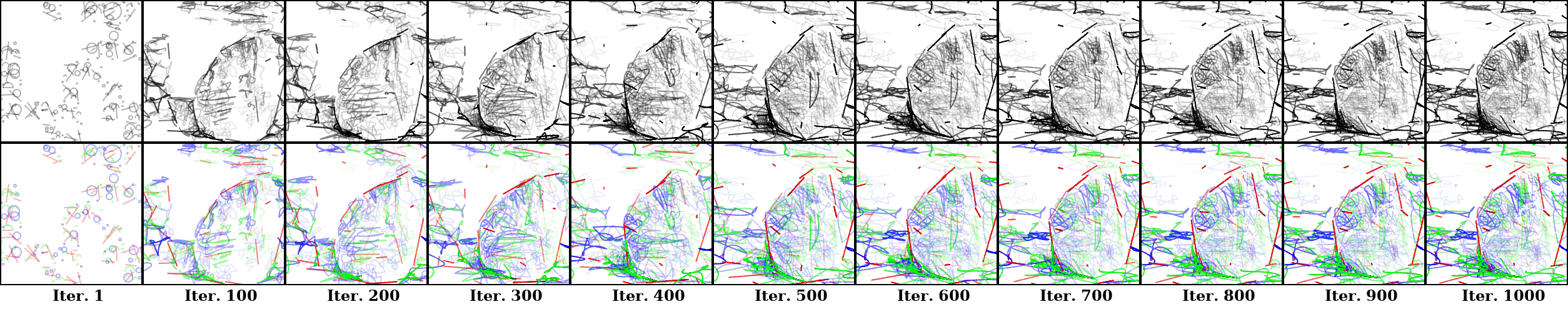}}
  
    \caption{Visualizations of synthesized sketches and its traceable version w.r.t. varying optimization iterations.}
    \label{fig:track_1}
\end{figure*}

\begin{figure*}[!htbp]
    \centering
    \subfloat[Peaks of a \hl{mountain} range.]{\includegraphics[width=0.92\textwidth]{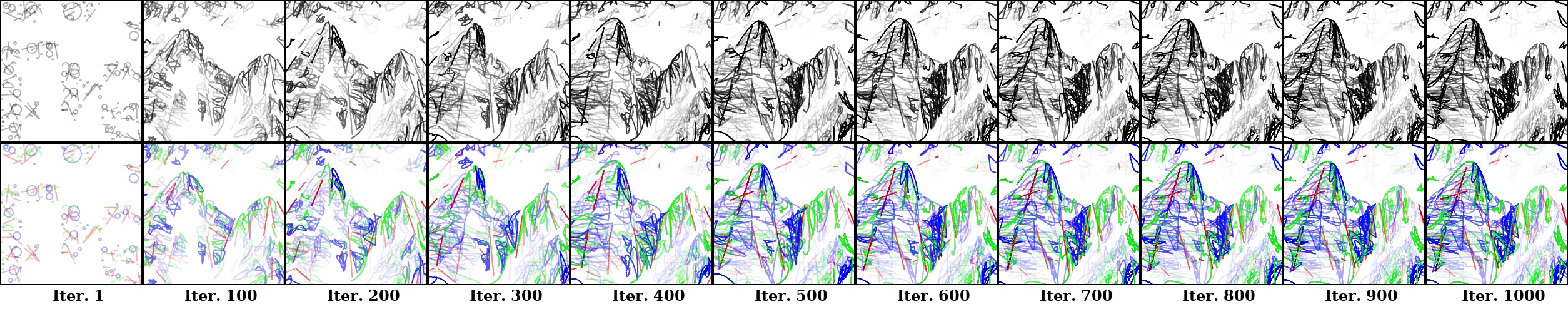}}
  
    \subfloat[A sketch of the mysterious \hl{octopus}.]{\includegraphics[width=0.92\textwidth]{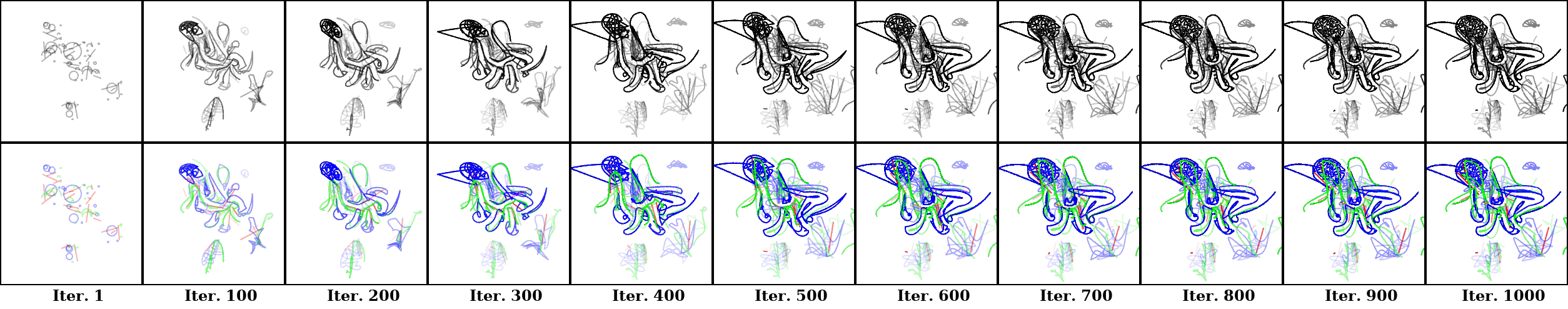}}
  
    \subfloat[An unique \hl{anteater}.]{\includegraphics[width=0.92\textwidth]{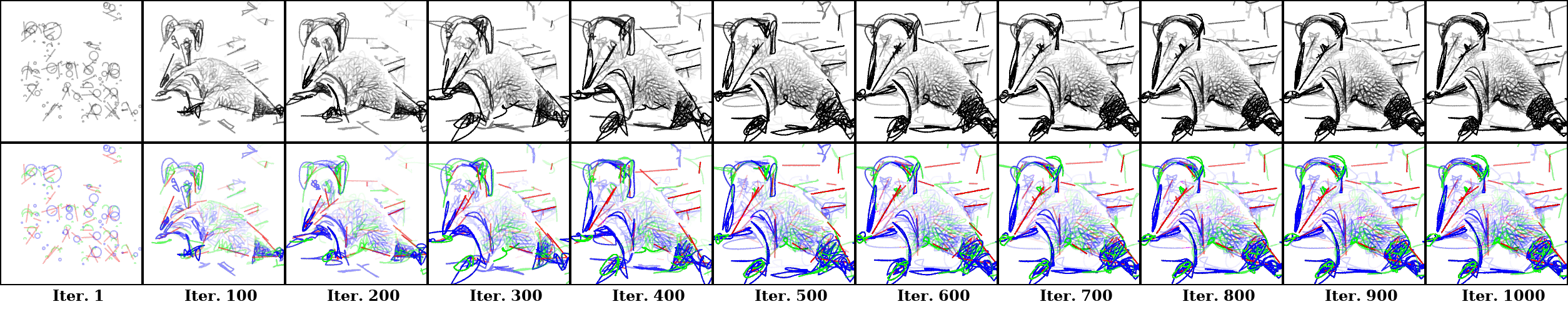}}
  
    \subfloat[The art of \hl{balance} scale.]{\includegraphics[width=0.92\textwidth]{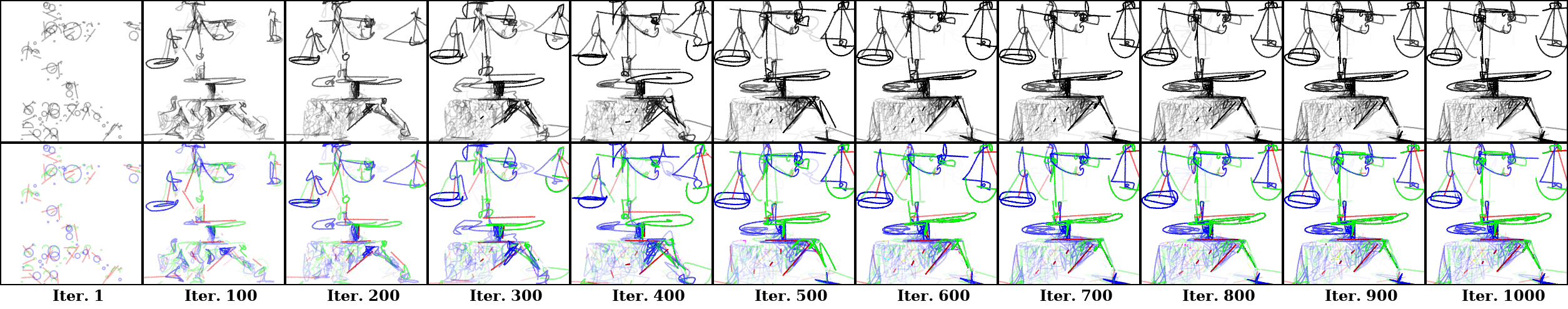}}

    \subfloat[A \hl{college} building surrounded by greenery.]{\includegraphics[width=0.92\textwidth]{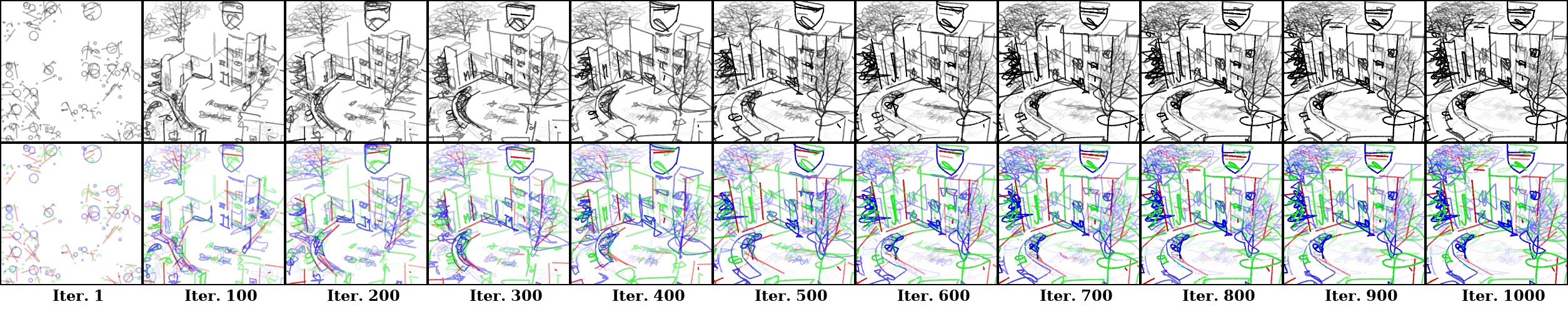}}

    \subfloat[A sketch of \hl{grapes} on the vine.]{\includegraphics[width=0.92\textwidth]{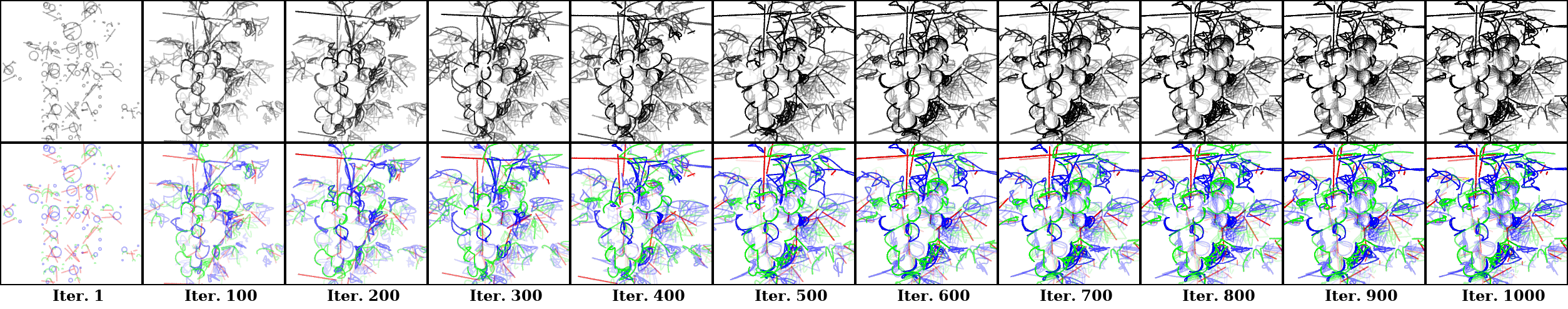}}
    
    \caption{Visualizations of synthesized sketches and its traceable version w.r.t. varying optimization iterations (continued to ~\cref{fig:track_1}).}
    \label{fig:track_2}
\end{figure*}

\begin{figure*}[!htbp]
    \centering
    \subfloat[A sketch of a simple \hl{basket}.]{\includegraphics[width=0.92\textwidth]{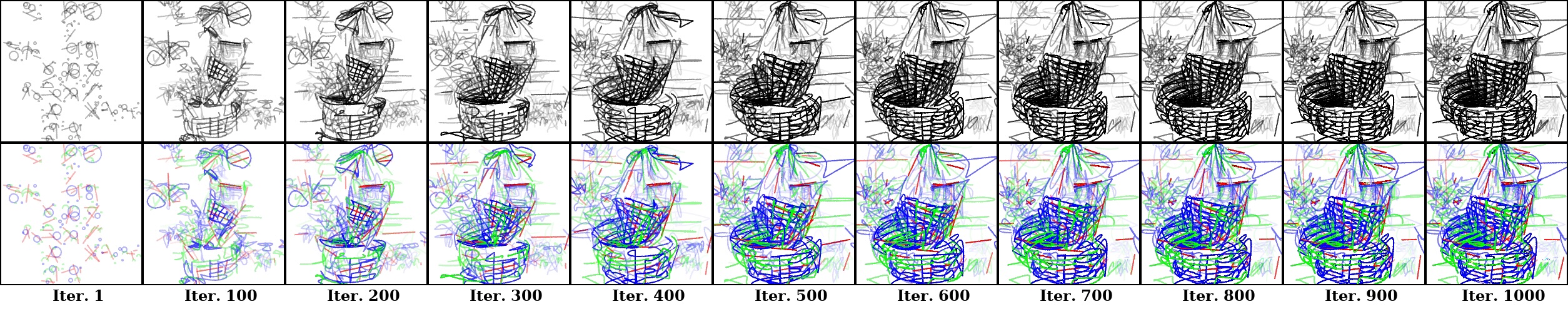}}
    
    \subfloat[A \hl{parrot's} solitude amidst nature.]{\includegraphics[width=0.92\textwidth]{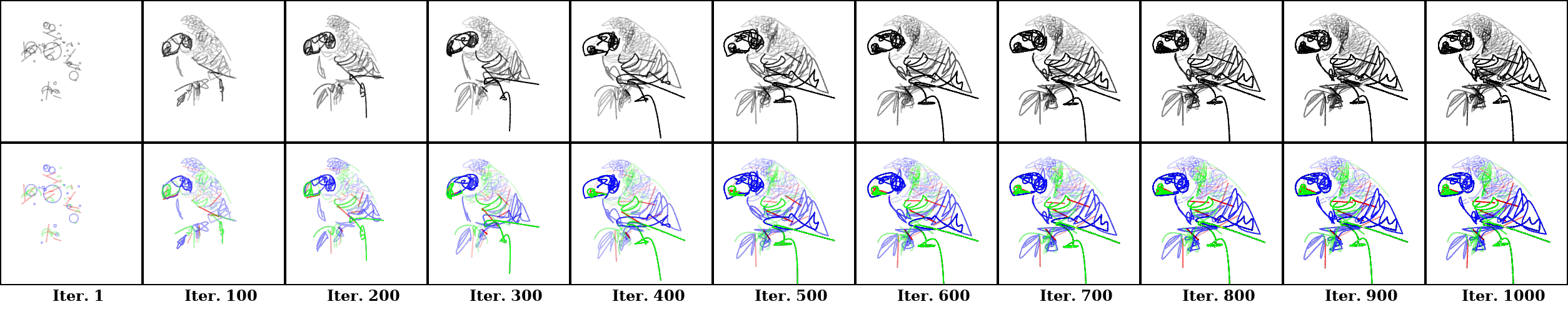}}

    \subfloat[Intense gaze of a \hl{chimpanzee}.]{\includegraphics[width=0.92\textwidth]{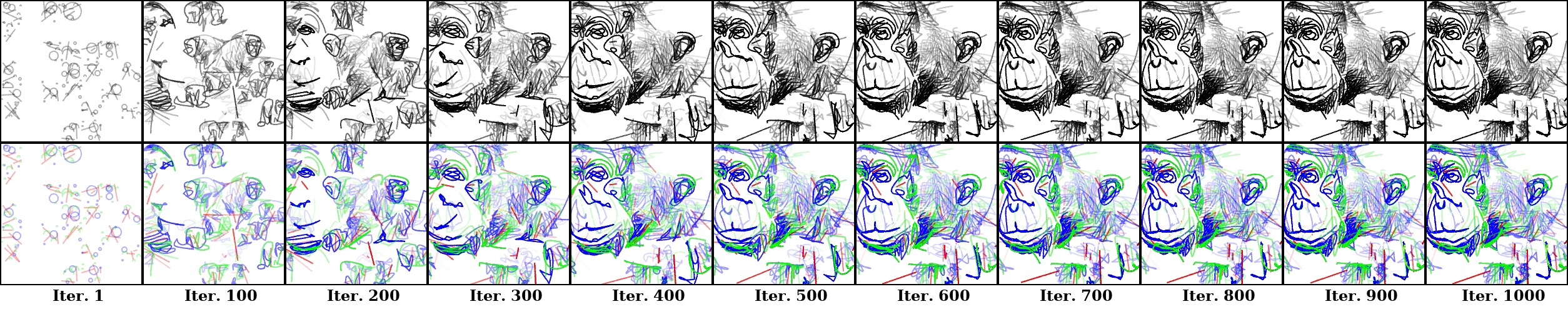}}

    \subfloat[A \hl{classroom} filled with students.]{\includegraphics[width=0.92\textwidth]{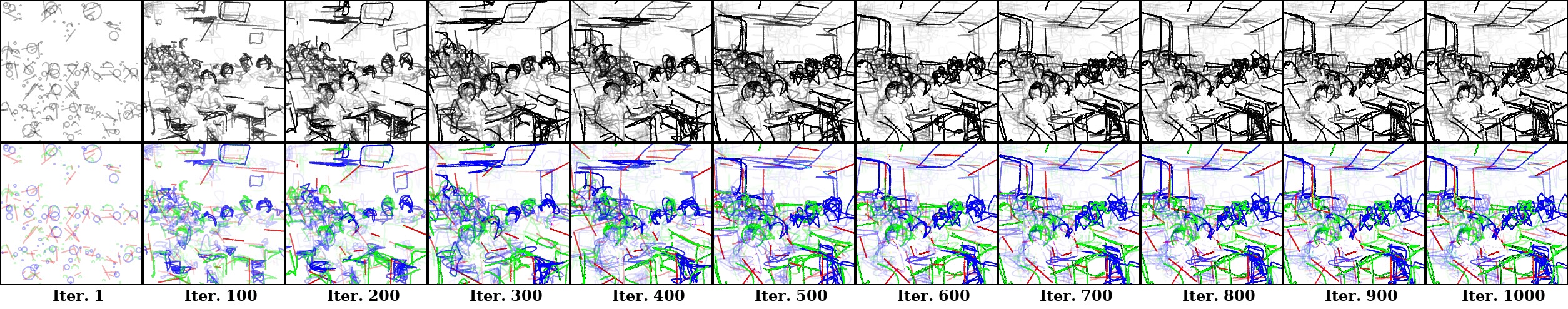}}

    \subfloat[The symphony of an \hl{accordion}.]{\includegraphics[width=0.92\textwidth]{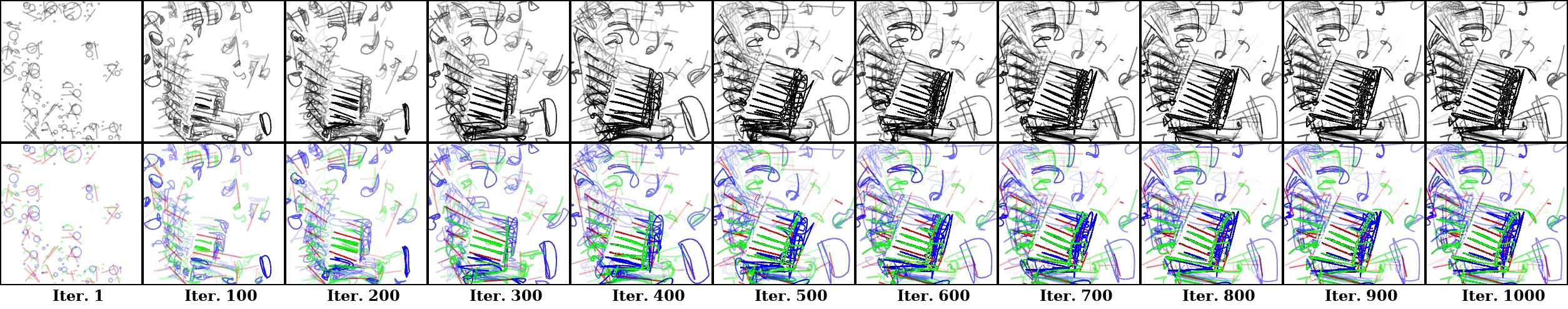}}

    \subfloat[A stroll through the park \hl{archway}.]{\includegraphics[width=0.92\textwidth]{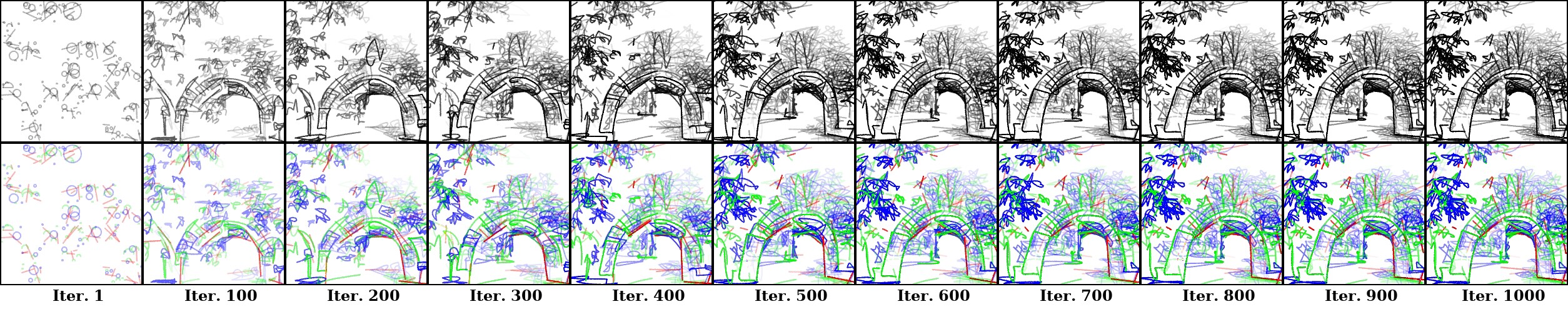}}

    \caption{Visualizations of synthesized sketches and its traceable version w.r.t. varying optimization iterations (continued to ~\cref{fig:track_1}).}
    \label{fig:track_3}
\end{figure*}

\begin{figure*}[!htbp]
    \centering
    \subfloat[\hl{Bees} hovering around.]{\includegraphics[width=0.92\textwidth]{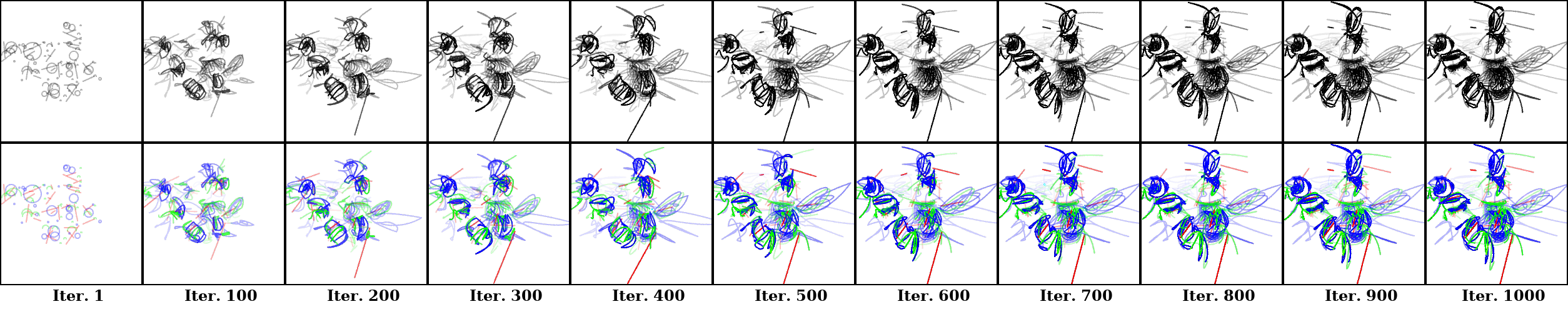}}
    
    \subfloat[A journey through the gallery of \hl{museum}.]{\includegraphics[width=0.92\textwidth]{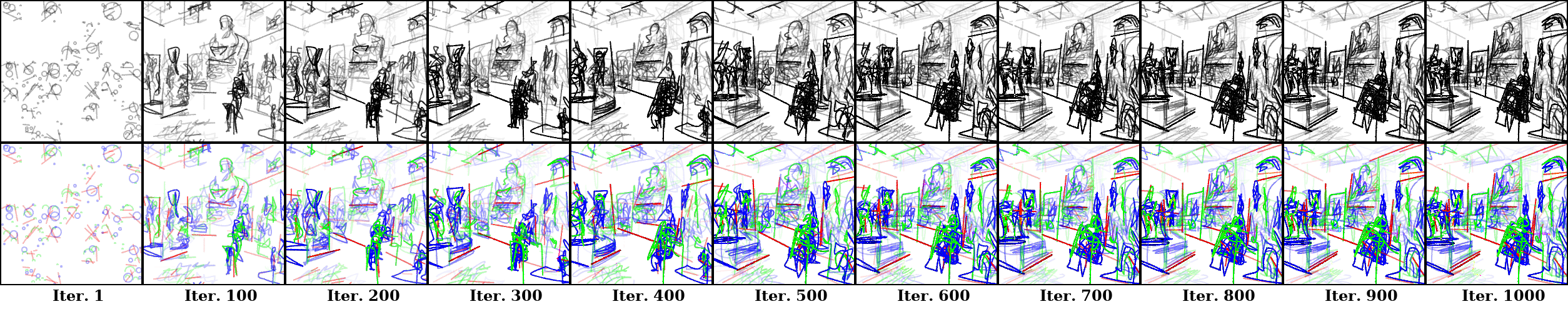}}

    \subfloat[A sketch of furious \hl{storm}.]{\includegraphics[width=0.92\textwidth]{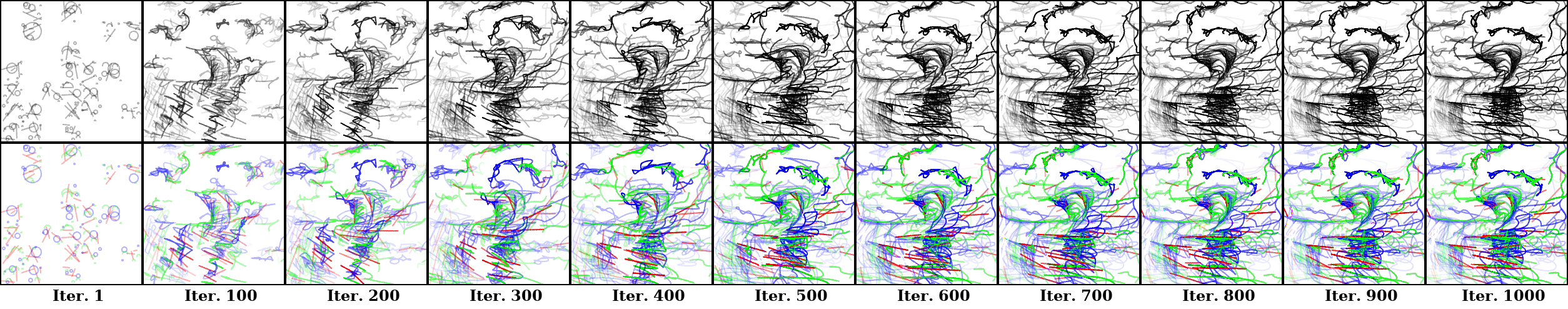}}
    
    \subfloat[A glimpse into the \hl{basement}.]{\includegraphics[width=0.92\textwidth]{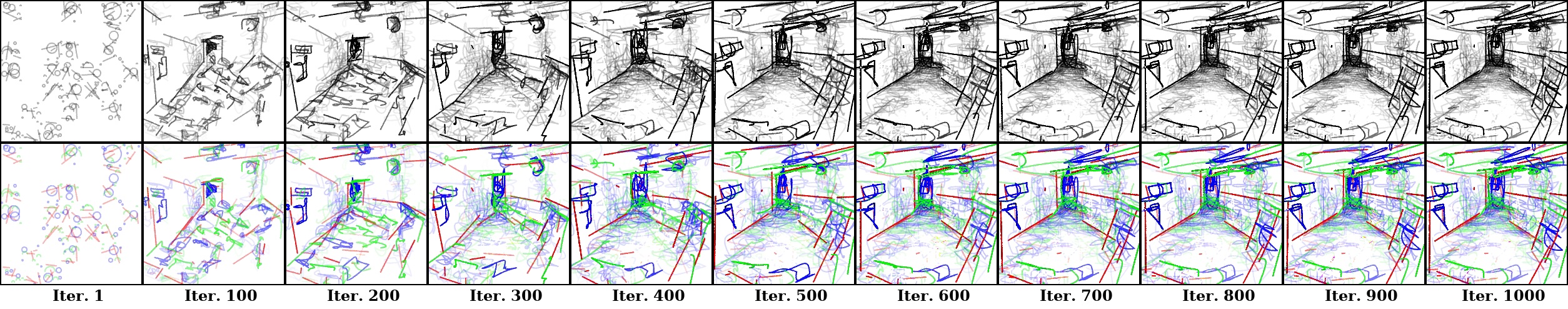}}

    \subfloat[A \hl{cactus} in a dry desert.]{\includegraphics[width=0.92\textwidth]{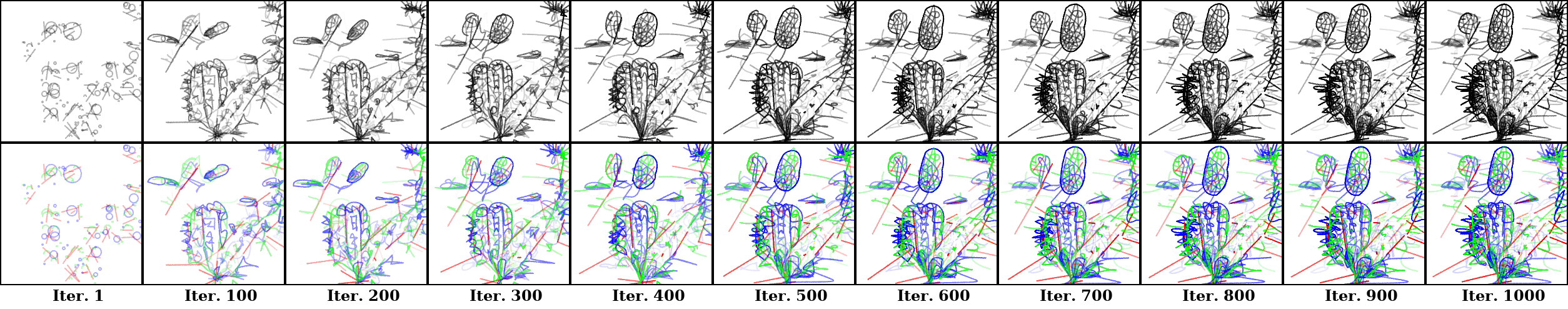}}

    \subfloat[\hl{Armenian} aura on a canvas of abstract impression.]{\includegraphics[width=0.92\textwidth]{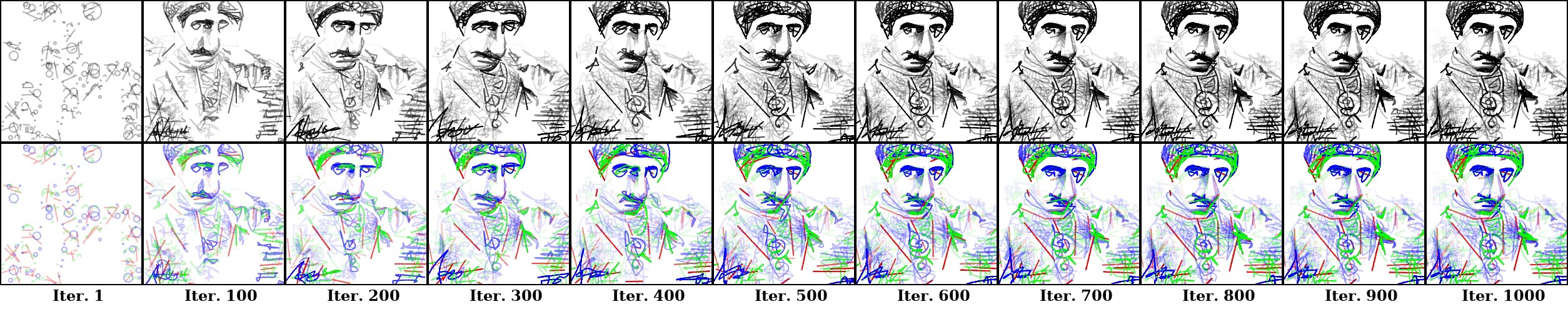}}
    
    \caption{Visualizations of synthesized sketches and its traceable version w.r.t. varying optimization iterations (continued to ~\cref{fig:track_1}).}
    \label{fig:track_4}
\end{figure*}

\newpage

\section{CLIPDraw++ Ablation Study}
\label{sec:ablation_study}
In this section, we showcase more examples to expound upon comprehensive ablation studies carried out on the different components of our CLIPDraw++ model.

\subsection{Impact of Primitive-level Dropout}
A thorough and intuitive mathematical explanation of Primitive-level Dropout (PLD) can be found in Sec. \mainsec{3.3} of the main paper. Within this section, we demonstrate the robust effectiveness of PLD by presenting additional examples with varying dropout probabilities: 0 (without PLD), 0.05, and 0.1. Our findings indicate that a dropout probability of 0.05 yields sketches that are visually and semantically more complete and coherent compared to those without PLD and with a higher dropout probability, such as 0.1.

\begin{figure*}[!h]
\includegraphics[width=\textwidth]{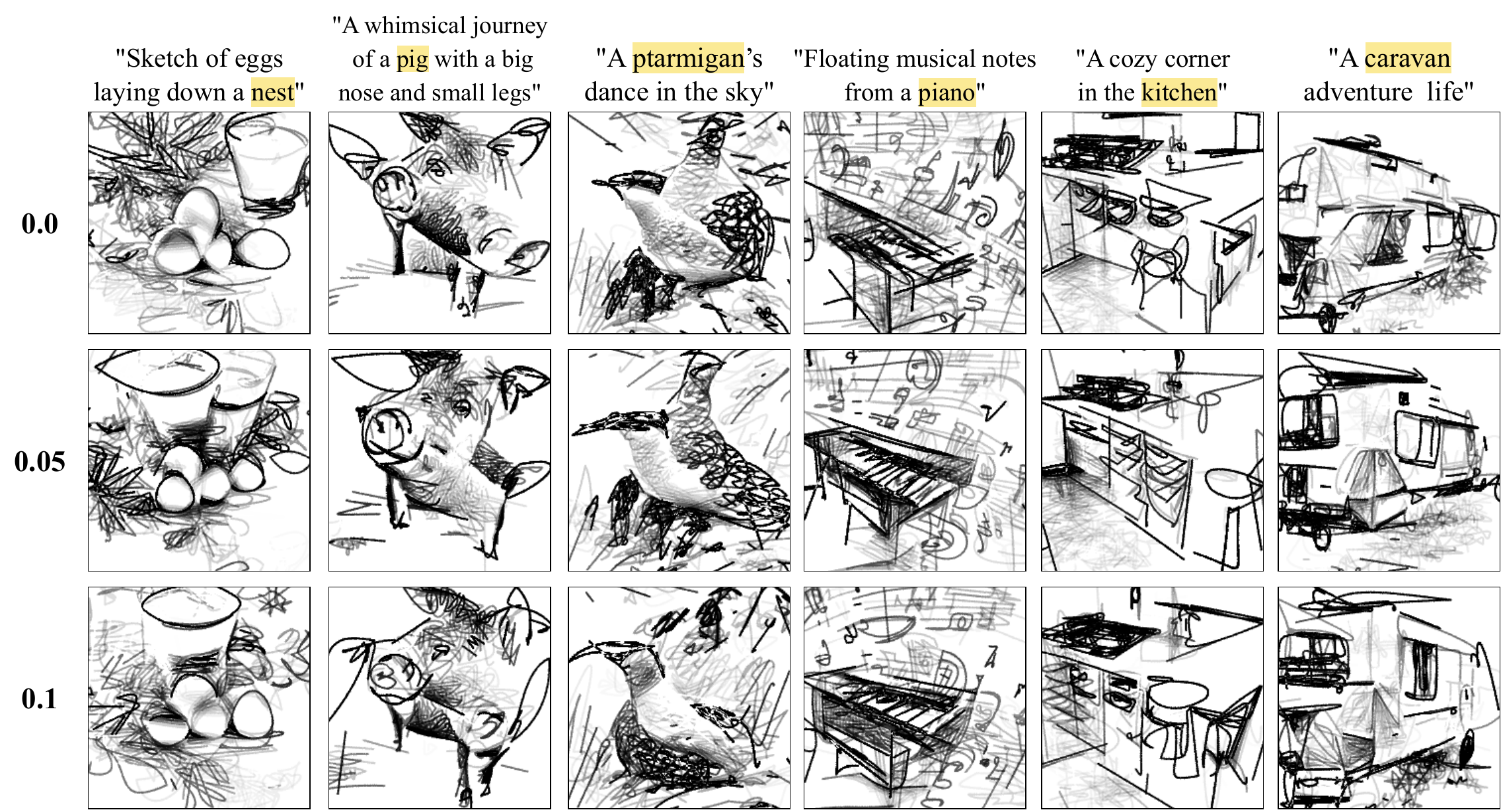}
\caption{Effectiveness of primitive-level dropout (PLD) across various text prompts with top-row sketches generated without dropout, middle-row sketches with a 0.05 dropout probability, and bottom-row sketches with a 0.1 dropout probability.}
\label{fig:PLD-full}
\end{figure*}

To illustrate, when using prompts such as \textit{``A whimsical journey of a pig with a big nose and small legs''} and \textit{``A ptarmigan’s dance in the sky''}, we obtain a good finishing of pig's face and ptarmigan, respectively at a dropout probability of 0.05. As depicted in \cref{fig:PLD-full}, sketches synthesized with a PLD of 0.05 generally exhibit cleanliness and realism, while those without PLD (top row) appear noisy and with a PLD of 1.0 occasionally exhibit incomplete structure in certain cases. This observation is also evident in \textit{``Floating musical notes from a piano''} and \textit{``A snug nook in the kitchen''} example where the structural completeness of the piano and the cozy kitchen corner are obtained with a PLD of 0.05. In the case of instances such as \textit{``eggnog''} and \textit{``caravan''} a PLD of 0.05 shows satisfactory outcomes but a PLD of 0.1 produces more promising results, highlighting the reliance of dropout probability on text prompts.

% \subsection{Starting with Diminished Opacity} 
% We initiate primitives with diminished opacity in order to mimic the human way of drawing with an aim to gradually increase the opacity of only the essential strokes needed to convey the text prompt’s semantics. 

\begin{figure*}[!h]
    \subfloat[Busy day in a multi-story \hl{mall}.]{\includegraphics[width=\textwidth]{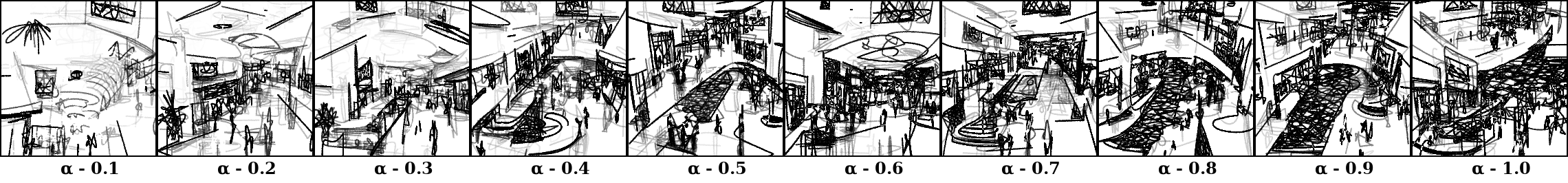}}

    \subfloat[A standing \highlight{motorcycle}.]{\includegraphics[width=\textwidth]{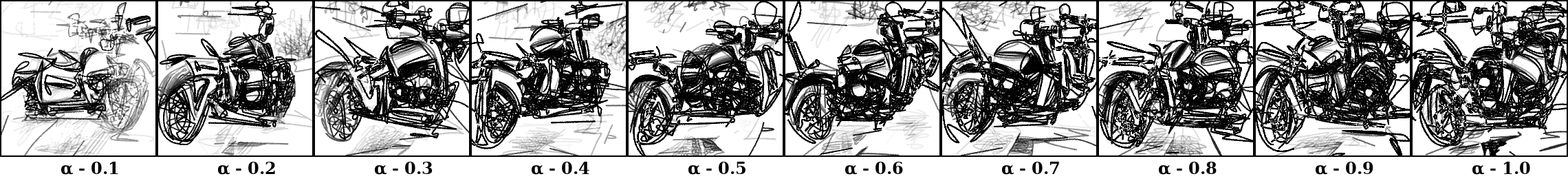}}

    \subfloat[A sketch of the mysterious \hl{octopus}.]{\includegraphics[width=\textwidth]{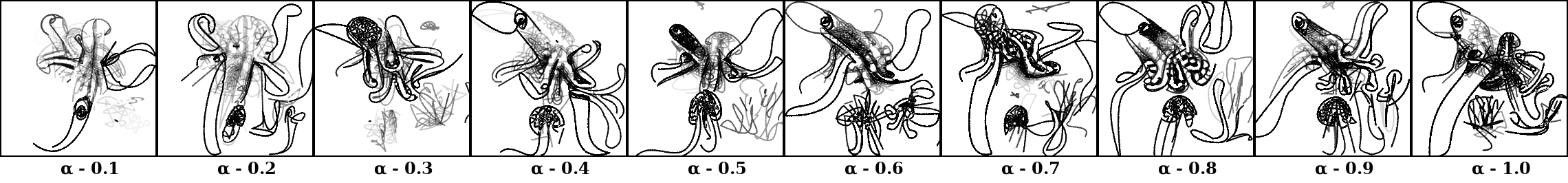}}

    \subfloat[A serene day in the \hl{park}.]{\includegraphics[width=\textwidth]{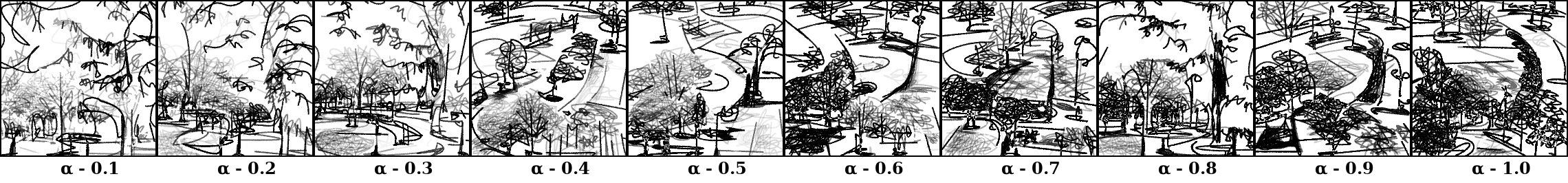}}

    \subfloat[Floating musical notes from a \highlight{piano}.]{\includegraphics[width=\textwidth]{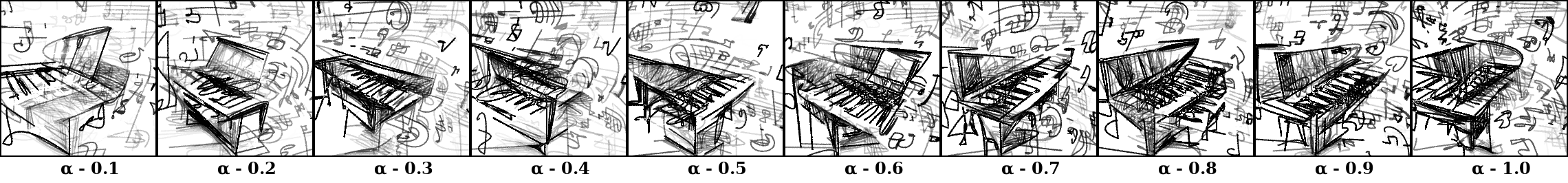}}
    
    \subfloat[A whimsical journey of a \hl{pig} with a big nose and small legs.]{\includegraphics[width=\textwidth]{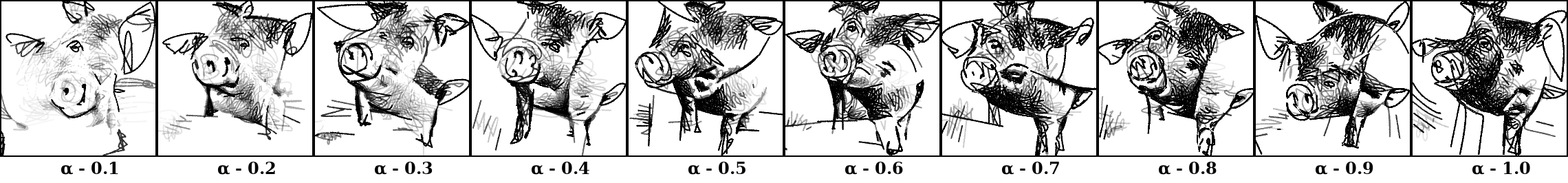}}

    \subfloat[Sketch of a \hl{sheep} at rest.]{\includegraphics[width=\textwidth]{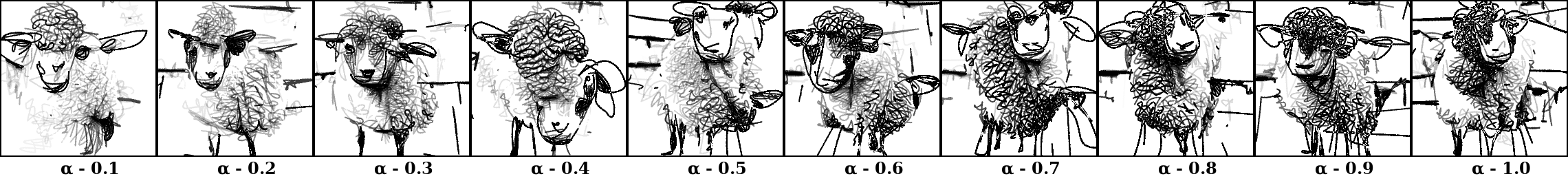}}

    \subfloat[A \hl{sideboard} amidst the room's rhythm.]{\includegraphics[width=\textwidth]{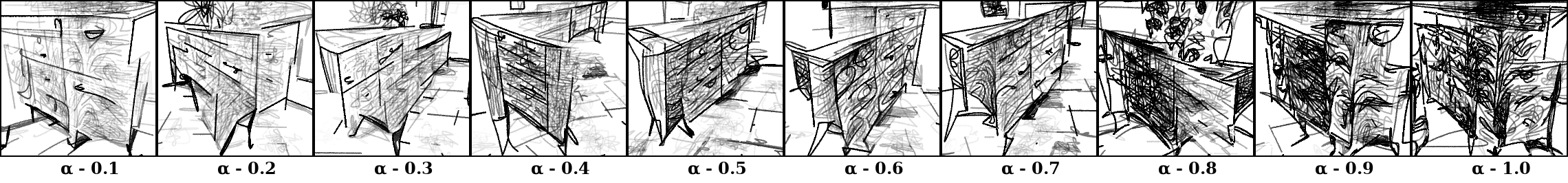}}
    
    \caption{Effectiveness of initializing primitives with diminished opacity. Initiating primitives with lower $\alpha$ yields cleaner final sketches compared to higher $\alpha$. We report optimal $\alpha$ value as 0.3.}
    \label{fig:int-full}
\end{figure*}

\subsection{Sketching with Diminished Opacity} 

% We start with highly transparent strokes (low opacity or a low $\alpha$ value) and gradually increase the opacity of only the essential strokes needed to convey the text prompt's semantics.
% In addition to primitive-level dropout, this approach further minimizes the presence of superfluous strokes in the synthesized sketches. As shown in \cref{fig:dim-opacity}, the final sketches which were initialized with lower $\alpha$ values are less noisy compared to the ones that are initialized with higher $\alpha$. By mimicking human drawing behaviour in this way, our approach demonstrates the potential to yield sketches that are more precise and finely crafted than those produced by conventional methods.

In ~\cref{fig:int-full}, we present additional results demonstrating the impact of initiating sketching with diminished opacity, showcasing its influence on various text prompts. Our findings indicate that initiating primitives with $\alpha$ value $0.3$ gives the best semantically aligned and clearer results across all prompts, values lower than this threshold compromise semantic integrity, while values higher than this threshold lead to noisy distorted, and altered sketches.

% \begin{figure*}[!h]
% % \vspace{-18pt}
%     \centering
%     \includegraphics[width=\textwidth]{Intensity/missile.png}
%     \caption{Effectiveness of initializing primitives with diminished opacity, indicated by lower $\alpha$ values, is notable. Initiating primitives with lower $\alpha$ in the prompt ``A \colorbox{Goldenrod}{missile} ready for launch'' yields cleaner final sketches compared to higher $\alpha$. }
%     \label{fig:dim-opacity}
% \end{figure*}

\subsection{Patch-based Initialization}
In CLIPDraw++, we adopted a patch-based approach for initializing strokes, placing primitives in patches to cover areas within a certain range, instead of just at the exact landmark points identified on the attention map. This method of patch-based initialization is designed to prevent the cluttering and messiness often seen with point-based initialization. As demonstrated in Fig. \mainsec{7} of main paper, sketches created through patch-based initialization stand out for their clarity and prominence. This approach allows for a more effective capture of the essential semantics of the input, resulting in cleaner, more coherent representations. In contrast, sketches originating from point-based initialization  tend to be muddled and unclear. Allowing primitives to be evenly distributed up to a certain distance of the attention local maxima (determined by the patch dimensions) prevents the model from being overly constrained, and also maintains clarity in each of the local regions at initialization. This helps the optimizer have a much clearer view of the canvas at the start, which, in turn, lets it retain, evolve, or drop primitives based on semantic requirements with greater ease -- all of which eventually contribute to much clearer output sketches.

\subsection{Analysis of Patch Size} 
\begin{figure*}[!htbp]
\includegraphics[width=\textwidth]{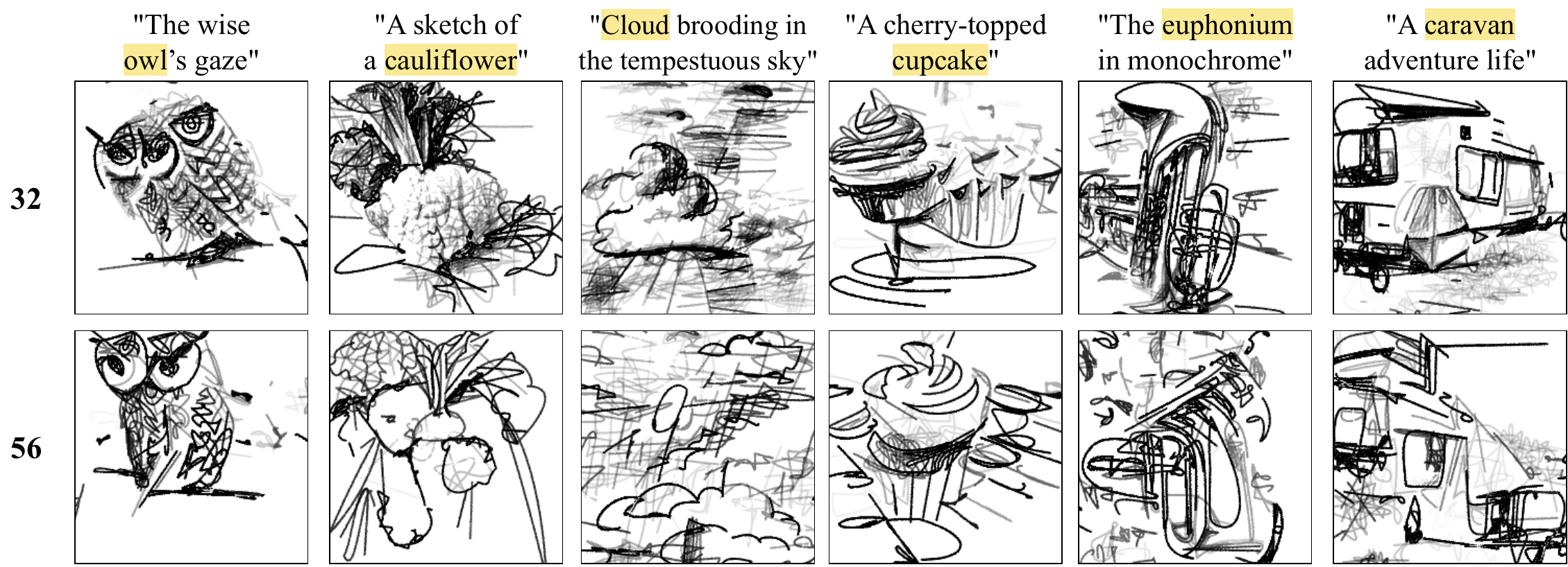}
    \caption{Comparative analysis of patch sizes ($32\times32$ vs. $56\times56$) in primitive initialization. The $32\times32$ patch consistently enhances visual details and textual coherence across most of the prompts.}
    \label{fig:patch-size}
\end{figure*}

The effectiveness of patch-based initialization in comparison to point-based initialization is detailed in Sec. \mainsec{3.2} and illustrated in Fig. \mainsec{7} of the main paper.
Here, we explore the influence of different patch sizes during primitive initialization. For this purpose, we divide a $224 \times 224$ canvas into smaller patches, with dimensions of $32 \times 32$ and $56 \times 56$. It is evident from the ~\cref{fig:patch-size} that, for the majority of the prompts, the $32\times 32$ patch exhibits more visual details and aligned textual semantics compared to the $56\times56$ patch size (the highest difference between them is being observed for prompts \textit{``A sketch of a cauliflower``} and \textit{``A caravan adventure life''}), with the exception of the \textit{``The wise owl’s gaze''} example where the $56\times56$ patch size displays more structural details of the owl.

\subsection{Effect of primitives count within a patch} 
\begin{figure*}
\includegraphics[width=\textwidth]{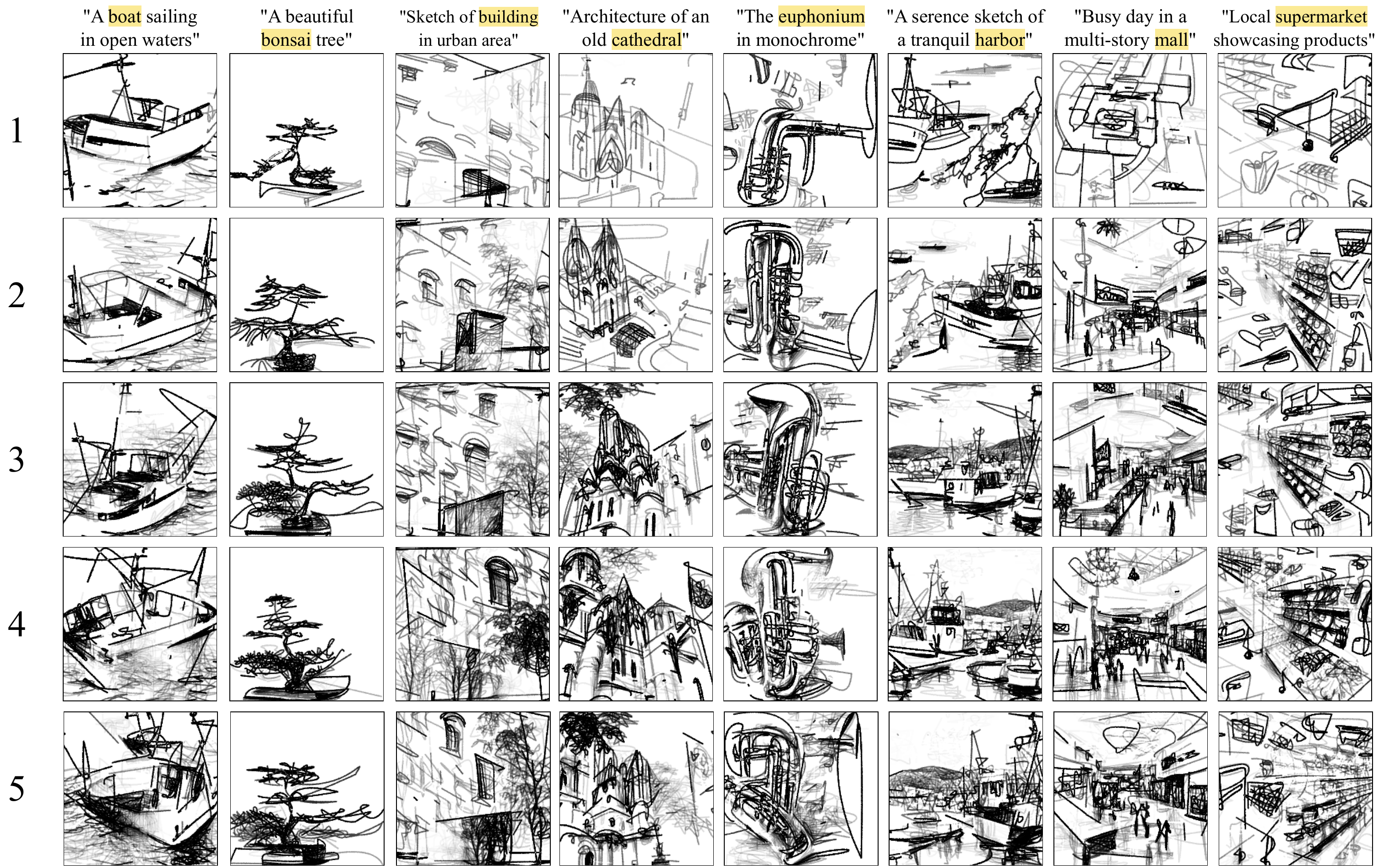}
    \caption{Exploring the impact of varying primitive counts for each type within selected patches on sketch outcomes with diverse text prompts in CLIPDraw++. (1) Few primitives yield abstract sketches (rows 1-2). (2) Optimal detail emerges with 3-4 primitives (rows 3-4). (3) Excessive counts (row 5) complicate optimization, hindering synthesis. The study recommends deploying 3 primitives per type for efficient synthesis, balancing complexity and resource utilization.}
    \label{fig:count}
\end{figure*}

CLIPDraw++ initializes the primitives within selected patches. In context, it becomes very crucial to study how many primitives from each type (straight line, circle, and semi-circle) we need to initialize within the selected patches.
In ~\cref{fig:count}, we delve into the impact of varying the number of primitives in each category on the synthesis of sketches using diverse text prompts. Here, our findings unfold as follows -- (1) When the primitive count is low for each type as seen in the first and second rows of ~\cref{fig:count}, CLIPDraw++ tends to generate abstract representations to the provided text prompts. (2) Increasing the primitive counts to around 3 or 4 for each type (third and fourth rows in ~\cref{fig:count}) results in more detailed drawings that incorporate additional features. (3) However, an increase in primitive counts beyond a certain threshold, as depicted in the fifth row of ~\cref{fig:count}, does not necessarily lead to improved sketch synthesis. Such increments introduce intricacies into the optimization process, making it more complex, time-consuming, and resource-intensive. Given the fact that for simpler prompts many of these primitives are superfluous for sketching, which may yield sub-optimal outcomes, as well.

Keeping the above-mentioned findings in mind, we consider deploying 3 primitives from each type within the designated patches, which aims to constrain the optimization process to a suitable image space while utilizing the minimum number of primitives or strokes necessary.

\begin{figure*}[!htbp]
    \centering
    \subfloat[A portrait of an \hl{actor}.]{\includegraphics[width=\textwidth]{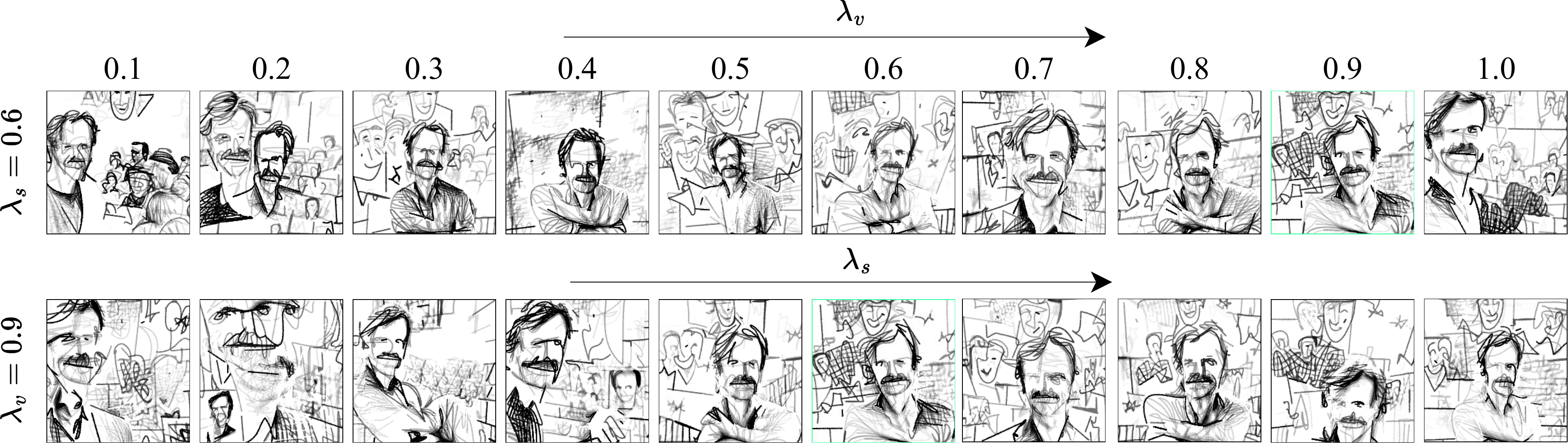} \label{actor_loss}}

    \subfloat[\hl{Apples} hanging in full bloom.]{\includegraphics[width=\textwidth]{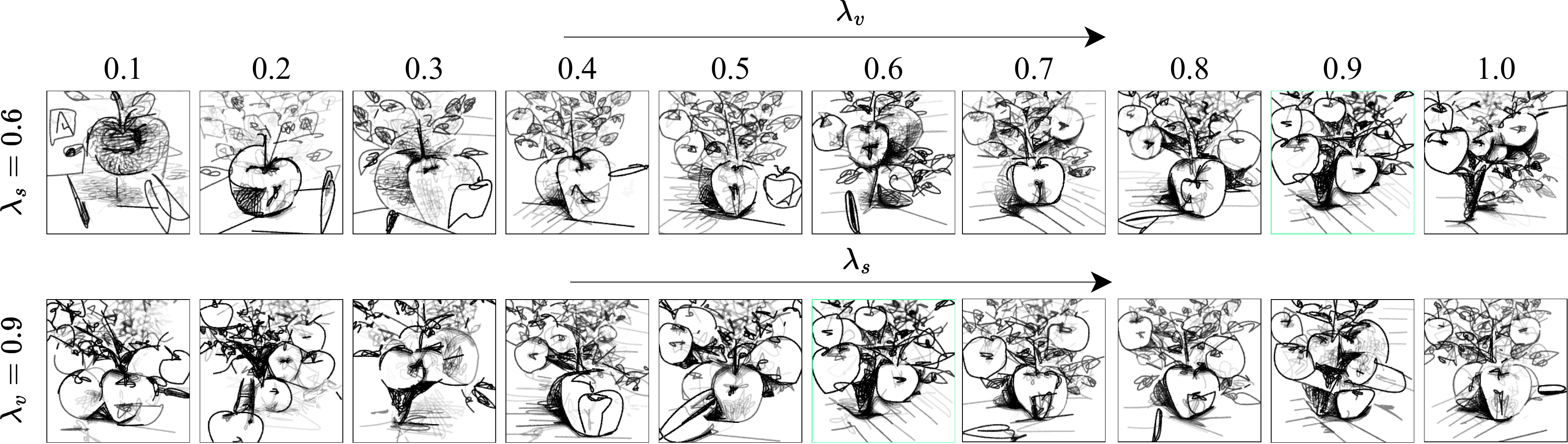} \label{apple_loss}}
    
    \caption{Ablation study on the impact of semantic and visual loss weights, $\lambda_{\text{sem}}$ and $\lambda_{\text{vis}}$ in the total loss function. Sketches reveal optimal values of 0.6 and 0.9, respectively, for $\lambda_{\text{sem}}$ and $\lambda_{\text{vis}}$.}
    \label{fig:loss-ablation}
\end{figure*}

\subsection{Loss Ablation and Analysis}
Our total loss function, $\mathcal{L}_{\text{total}}$ is composed of two loss functions, semantic loss: $\mathcal{L}_{\text{sem}}$ and visual loss: $\mathcal{L}_{\text{vis}}$, weighted by their respective coefficients or importance, $\lambda_{\text{sem}}$ and $\lambda_{\text{vis}}$. 
\begin{equation*}
    \mathcal{L}_{\text{total}} = \lambda_{\text{sem}}\mathcal{L}_{\text{sem}} + \lambda_{\text{vis}}\mathcal{L}_{\text{vis}}
\end{equation*}
A detailed explanation of the aforementioned loss function and its components are given in Sec. \mainsec{3.4} of the primary manuscript. Within ~\cref{fig:loss-ablation}, we present two sketches corresponding to the prompts \textit{``A portrait of an actor''} and \textit{``Apples hanging in full bloom''} while varying weight factors $\lambda_{\text{sem}}$ and $\lambda_{\text{vis}}$ individually. For the ``actor'' example in ~\cref{actor_loss}, we fix $\lambda_{\text{sem}}$ as 0.6 and vary $\lambda_{\text{vis}}$ values while in the second row, we fix $\lambda_{\text{vis}}$ and vary $\lambda_{\text{sem}}$ to show the impact of each components. Our analysis establishes the optimal values for $\lambda_{\text{sem}}$ and $\lambda_{\text{vis}}$ at 0.6 and 0.9, respectively. Utilizing these prescribed values yields a meticulous representation of the upper body structure for the actor (in \cref{actor_loss}) and an accurately contoured depiction of the apple (in \cref{apple_loss}) example.

\subsection{Variations}
\begin{figure*}[!htbp]
    \centering
    \includegraphics[scale=0.4]{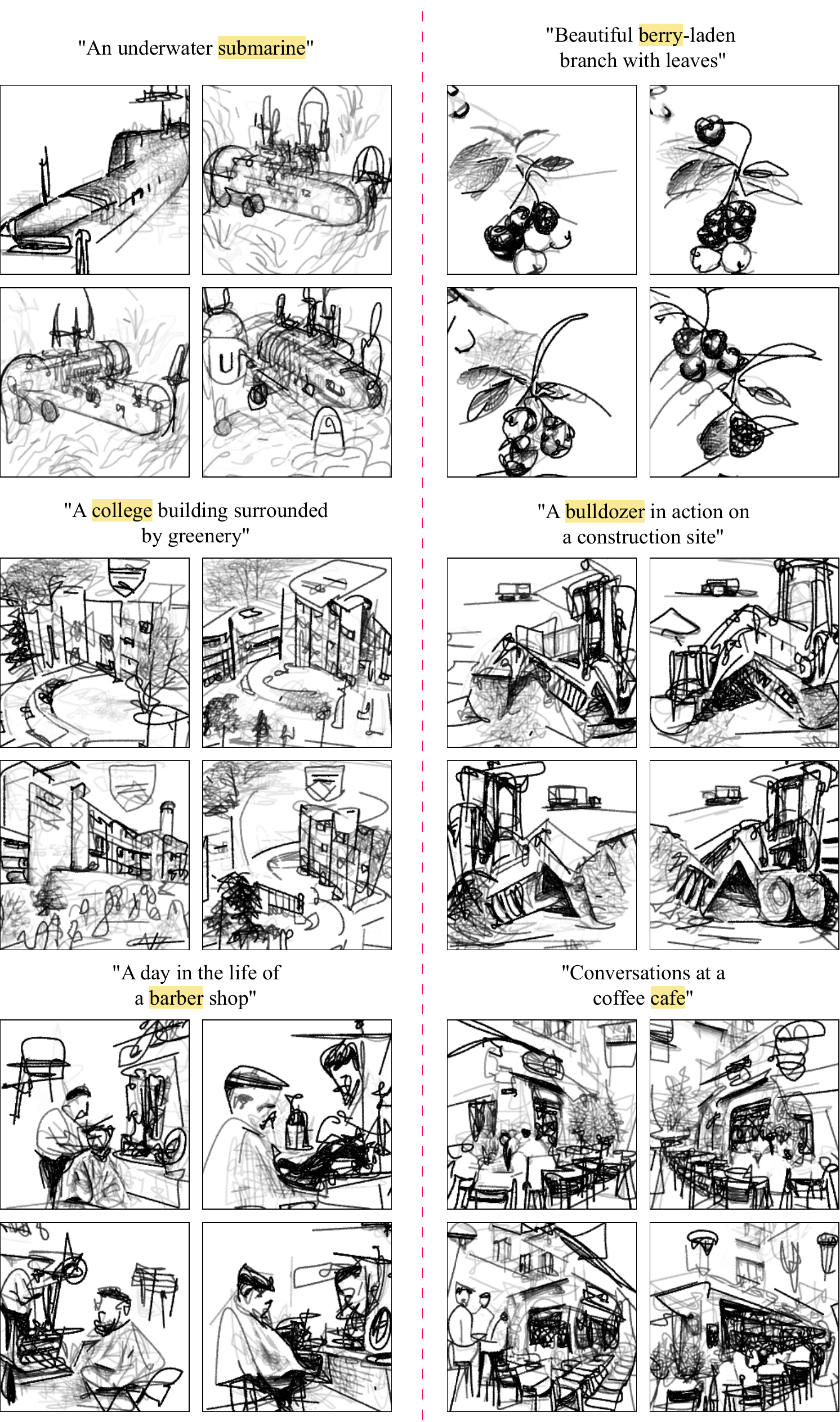}
    \caption{Our proposed CLIPDraw++ method offers the generation of diverse sketches while adhering to their corresponding textual semantics. In each example, given the same text prompt, four different sketches are synthesized using random seeds.}
    \label{fig:variations}
\end{figure*}

Being a generative sketch synthesis model, our CLIPDraw++ generates diverse sketches contextualized by the same text prompt. These enhanced variations are attributed to the utilization of robust latent diffusion models during the initialization stage. \cref{fig:variations} provides evidence for the aforementioned claims, displaying four different output sketches generated in response to the same input description.

% \begin{figure*}[!htbp]
%     \centering 
%     \includegraphics[width=\textwidth]{images/supp1.pdf}
%     \includegraphics[width=\textwidth]{images/supp2.pdf}
%     \caption{Qualitative comparison between CLIPDraw, BigGAN and our CLIPDraw++.}
%     \label{fig:comparative}
% \end{figure*}

% \begin{table*}[!htbp]
%     \centering
%     % \resizebox{\columnwidth}{!}{
%     \begin{tabular}{c|c|c|c|c}
%     \toprule
%         \textbf{Model} & CLIPDraw++ & CLIPDraw  & VectorFusion & CLIPasso \\\midrule
%         \textbf{CLIP-T} & 0.3365 & 0.3114 & 0.2949 & 0.2965 \\\bottomrule
%     \end{tabular}
%     % }
%     \caption{Quantitative comparisons using CLIP-T score. We have chosen to omit BigGAN from consideration, as it primarily focuses on image synthesis rather than sketch generation. Higher the cosine similarity, higher is the adherence of synthesized sketches to text prompts.}
% \end{table*}

% For each method, alongside the mean values, we also extract the standard deviations.
% \begin{figure*}[!ht]
%     \centering
%     \includegraphics[width=\textwidth]{images/clipasso_comp.pdf}
%     \caption{Qualitative comparison with CLIPasso and our CLIPDraw++.}
%     \label{fig:clipasso}
% \end{figure*}

\begin{figure*}[!h]
    \centering
    \includegraphics[width=\textwidth]{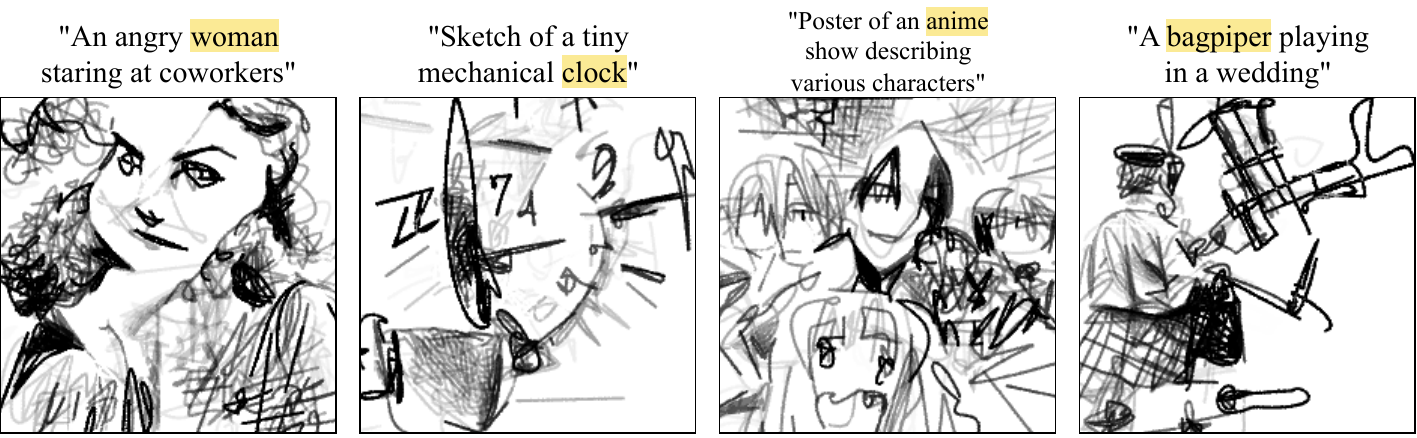}
    \caption{Some instances where the CLIPDraw++ model failed to synthesize sensible results.}
    \label{fig:failure}
\end{figure*}

\section{Limitations}
\label{sec:limitations}
In this section, we shed some light on the proposed method's limitations, showcasing failure cases and exploring possible reasons for such occurrences as follows:
\begin{itemize}
    \item \textbf{\textit{Interpretation of text prompts}}: LDM (used to compute DAAM maps) and CLIP are not trained on similar objectives, which could lead to discrepancies in the way they process and interpret the input signals. In ~\cref{fig:failure}, the sketch of \textit{``An angry woman staring at coworker''} might not depict the
    emotion of a woman as being angry or the context of her staring at a coworker.\\
    This issue can be somehow tackled by generating a variety of cross-attention maps through the alteration of the optimizer’s \textit{SEED}.

    \item \textbf{\textit{Lack of details}}: The sketches might lack the necessary detail to fully convey all the contexts within a prompt. For instance, the \textit{``Sketch of a tiny mechanical clock''} (\cref{fig:failure}) might not clearly show the mechanical aspects of the clock. Similarly for \textit{``Poster of an anime show describing various characters''} does not describe all the characters precisely.\\
    The complexity of the sketches generated can vary based on the primitive count. Increasing the number of primitives can generate more complex visual representations and textures for the input text prompt.
\end{itemize}

% \input{tex/supp}

\end{document}